%% file: main.tex
\documentclass[11pt]{article}

\input{preamble/packages}
\input{preamble/macros}

\input{preamble/environments}
\input{preamble/figures}

\title{Does More Compute Make Every Adversarial Attack Transferable?}

\author{
Greg Gluch\\UC Berkeley, RESI
\and
Shafi Goldwasser\\MIT, RESI
\and
Anastasiia Struss\\MIT
}
\date{}

\begin{document}

\maketitle

\begin{abstract}
\input{sections/abstract}
\end{abstract}

\newpage
\setcounter{tocdepth}{3}
\tableofcontents
\newpage

\section{Introduction}
\label{sec:introduction}
\input{sections/introduction}

\section{Technical Overview}
\label{sec:overview}
\input{sections/overview}

\section{Preliminaries and Definitions}
\label{sec:model}
\input{sections/preliminaries}

\section{Classification: Transferable Attacks or Adversarial Defenses at a Quadratic Gap}
\label{sec:classification}
\input{sections/classification}

\section{Generation: Active Adversarial Defenses Are Strictly Stronger}
\label{sec:separation}
\input{sections/generation}

\section{Generation: Neither Transferable Attacks nor Adversarial Defenses Below a Quadratic Gap}
\label{sec:neither}
\input{sections/neither}

\newpage
\section*{AI Disclosure}
\input{sections/ai_use_statement}

\renewcommand{\bibsection}{\section*{\refname}\phantomsection\addcontentsline{toc}{section}{\refname}}
\bibliography{bib/references}
\bibliographystyle{plainnat}

\newpage
\appendix

\addcontentsline{toc}{section}{Appendix}

\addtocontents{toc}{\protect\setcounter{tocdepth}{0}}

\section{Comparison with the Trichotomy of Gluch et al.}
\label{app:trichotomy}
\input{appendix/trichotomy}

\section{Two-Player Zero-Sum Games}
\label{app:games}
\input{appendix/games}

\section{Cryptographic Tools}
\label{app:crypto}
\input{appendix/crypto}

\end{document}

%% file: preamble/packages.tex
\usepackage[margin=1in]{geometry}
\usepackage{natbib}
\setcitestyle{authoryear,round,citesep={;},aysep={,},yysep={;}}

\usepackage{amsmath,amssymb,amsfonts,amsthm}
\allowdisplaybreaks
\usepackage{graphicx}
\usepackage{microtype}
\usepackage{xcolor}
\usepackage{enumitem}
\usepackage{todonotes}
\usepackage{caption}
\usepackage{booktabs}
\usepackage{dirtytalk}

\usepackage{tikz}
\usetikzlibrary{shapes.geometric,arrows.meta,positioning,decorations.pathreplacing,calc,intersections}

\usepackage{hyperref}
\usepackage{url}
\usepackage{xcolor}

\usepackage{booktabs,siunitx,threeparttable}
\usepackage{comment}

\usepackage{float}

%% file: preamble/macros.tex
\newcommand{\N}{\mathbb{N}}
\newcommand{\R}{\mathbb{R}}
\newcommand{\negl}{\mathsf{negl}}
\newcommand{\poly}{\mathsf{poly}}

\newcommand{\prm}{n}
\newcommand{\X}{\mathcal{X}}
\newcommand{\Y}{\mathcal{Y}}
\newcommand{\inspace}{\X_\prm}
\newcommand{\outspace}{\Y_\prm}
\newcommand{\dist}{\mathcal{D}}
\newcommand{\task}{\mathbb{L}}
\newcommand{\taskcon}{\mathbb{L}^\mathsf{data}}
\newcommand{\tasktime}{\mathbb{L}^\mathsf{time}}
\newcommand{\fclass}{\mathcal{F}}

\newcommand{\ver}{A}
\newcommand{\prov}{B}
\newcommand{\provtrain}{\prov_{\mathsf{Train}}}
\newcommand{\provfix}{\prov_{\mathsf{Inference}}}
\newcommand{\ttrain}{S_{\mathsf{Train}}}
\newcommand{\tfix}{S_{\mathsf{Inference}}}
\newcommand{\ktrain}{K_{\mathsf{Train}}}
\newcommand{\kfix}{K_{\mathsf{Inference}}}

\newcommand{\TA}{\mathsf{TA}}
\newcommand{\AD}{\mathsf{AD}}
\newcommand{\AAD}{\mathsf{AAD}}

\newcommand{\xs}{\mathbf{x}}
\newcommand{\ys}{\mathbf{y}}
\newcommand{\rs}{\mathbf{r}}
\newcommand{\priv}{\mathsf{priv}}
\newcommand{\err}{\operatorname{err}}
\newcommand{\errf}{\operatorname{err}_{\dist, h}(f)}
\newcommand{\fl}{\operatorname{fl}}
\newcommand{\ex}{\operatorname{Ex}(\dist)}
\newcommand{\exh}{\operatorname{Ex}(\dist, h)}
\newcommand{\exfor}[2]{\operatorname{Ex}(#1, #2)}

\newcommand{\pk}{\mathsf{pk}}
\newcommand{\sk}{\mathsf{sk}}
\newcommand{\idt}{\mathsf{id}}
\newcommand{\Gen}{\mathsf{Gen}}
\newcommand{\Enc}{\mathsf{Enc}}
\newcommand{\Dec}{\mathsf{Dec}}
\newcommand{\Eval}{\mathsf{Eval}}
\newcommand{\gensnark}{\mathsf{G}}
\newcommand{\provsnark}{\mathsf{P}}
\newcommand{\versnark}{\mathsf{V}}

\newcommand{\GivenExperiment}{\;\middle\vert\;}
\newcommand{\StateExperiment}[1]{
\!\begin{array}{l}
#1
\end{array}\!}

\newlength{\flushrowoffset}
\newcommand{\flushrow}[2][]{%
  \settowidth{\flushrowoffset}{$\displaystyle #1$}%
  \makebox[\dimexpr\linewidth-\flushrowoffset\relax][r]{$\displaystyle #2$}}

\newcounter{mycomment}

%% file: preamble/environments.tex
\theoremstyle{plain}
\newtheorem{theorem}{Theorem}
\newtheorem{lemma}[theorem]{Lemma}
\newtheorem{corollary}[theorem]{Corollary}

\newtheorem{remark}[theorem]{Remark}

\theoremstyle{definition}
\newtheorem{definition}[theorem]{Definition}

%% file: preamble/figures.tex
\definecolor{figRed}{HTML}{E63946}
\definecolor{figBlue}{HTML}{457B9D}
\definecolor{figGray}{HTML}{87919D}
\usepackage{fix-cm}
\newcommand{\figscale}{1}
\newcommand{\figsize}[1]{%
  \fontsize{\fpeval{#1/\figscale}}{\fpeval{1.2*#1/\figscale}}\selectfont}
\newcommand{\figheadingfont}{\figsize{9}\bfseries}
\newcommand{\figtextfont}{\figsize{9}}
\newcommand{\fignotefont}{\figsize{9}}
\tikzset{
  figthin/.style={line width=0.5pt/\figscale},
  figbox/.style={line width=0.8pt/\figscale},
  figarrow/.style={line width=0.9pt/\figscale},
  figfixed/.style={line width=1.9pt/\figscale},
  figaccent/.style={line width=1.4pt/\figscale,line cap=round},
}
\tikzset{
  pics/figpanel/.style 2 args={code={
    \node[font=\figheadingfont,text=black,anchor=base,inner sep=0pt]
      at (0,0) {#2};
    \draw[figaccent,draw=#1]
      (-0.6cm/\figscale,-5.8pt/\figscale) --
      (0.6cm/\figscale,-5.8pt/\figscale);
  }},
}

%% file: sections/abstract.tex
Transferable attacks, or transferable jailbreaks, are a particularly strong version of adversarial examples that fool all defenders in a class.
In this work, we study when more compute enables transferable attacks.
We show that, when there is a quadratic gap in computational resources between attacker and defender, every learnable \emph{classification} task admits either a transferable attack or an adversarial defense.
For \emph{generative} tasks, this dichotomy fails below a quadratic gap, i.e., more compute for the attacker does not lead to a transferable attack.

We prove a stronger result that implies this failure: a defense that spends additional computation after it sees the input can be provably strictly stronger than every defense that publishes a fixed model.
This result provides formal evidence that increasing inference-time compute can improve robustness, as empirically observed by \citet{boaz2025trading}. We assume the existence of an array of cryptographic primitives: identity-based fully homomorphic encryption, publicly verifiable zero-knowledge succinct non-interactive arguments of knowledge, non-parallelizing languages, and incrementally verifiable computation.

%% file: sections/introduction.tex
AI agents carry out user requests by processing documents, browsing webpages, and interacting with external tools. This access to information exposes them to adversarial content: instructions embedded in a document or tool response can redirect an agent away from its intended task. Such indirect prompt injection attacks have been demonstrated in applications that integrate large language models (LLMs) and studied systematically in tool-using AI agents \citep{greshake2023indirect,debenedetti2024agentdojo}. These attacks bring renewed urgency to a fundamental theoretical question in adversarial robustness: can an AI model detect adversarial inputs on which it would potentially make a mistake? Detection has been studied for bounded perturbations \citep{pmlr-v162-tramer22a}, for arbitrary test examples with abstention \citep{arbitraryexamples}, and for prompt filters of language models \citep{ball2025impossibilityseparatingintelligencejudgment}.

Transferability is one of the most surprising features of adversarial attacks. Different models could have entirely different weaknesses, yet an input constructed to fool one model can also fool others. This phenomenon was observed in image classification and later in jailbreaks against aligned language models \citep{intriguingprop,Papernot2016TransferabilityIM,liu2016delving,Zou2023UniversalAT}. \citet{bose2020adversarial} cast the search for attacks that transfer to an entire hypothesis class as a min-max game. The observed transfer between particular models invites the following question:

\vspace{\baselineskip}
\emph{Question 1.} \emph{When does the existence of an attack against each individual model imply the existence of a single attack that transfers to all models from a class?}

\vspace{\baselineskip}
In this work we ask how the answer depends on the computational resources available to the attacker and defender. 
A Transferable Attack (TA) is a single attack procedure that succeeds against every defender within a specified resource bound. 
An Adversarial Defense (AD) publishes a model that remains useful on natural inputs and withstands every attacker within the corresponding resource bound.\footnote{By a published model we mean the trained model made available (with white-box access) to the attacker.}

For classification, we prove that every learnable task admits a TA or an AD, with a roughly quadratic resource tradeoff. 
In this sense, this resolves \emph{Question 1 affirmatively for classification tasks} at this gap: if every model can be attacked, then a single attack transfers to every defender with quadratically fewer resources.
We build on the framework of \citet{gluch2026good}, who established a trichotomy between TAs, ADs, and Watermarks (WMs) depending on computational power of the parties.
Our result shows that the WM alternative is not needed at the quadratic gap.
We note \citet{bubeck2019adversarial, degwekar2019computational, gargcompadvexp} also considered computational constraints in the context of adversarial robustness.

We emphasize that the dichotomy between TAs and ADs, where it holds, is good news either way.
Either a defense exists, and we can use it.
Or a single attack wins against every defender within our resource bounds: no defender we could build at these resources would withstand it, so there is no point in trying harder.

We next consider generative tasks, where an input can have many valid answers.
We construct a learning task that, under cryptographic assumptions, admits
neither a TA nor an AD just below a quadratic resource gap.
This gives a \emph{negative answer to Question 1} for generative tasks
in this resource regime.
To establish this result, we prove the following statement first: additional inference-time computation can improve robustness.

This statement is motivated by modern AI systems, which often reason,
search, or call tools after receiving a query, devoting additional computation
to that query.
Experiments with reasoning models show that such computation can improve robustness \citep{boaz2025trading};
see also \citet{wu2025does} and \citet{mcdonald2026rich}.





We formalize defenses using inference-time resources as an Active Adversarial Defense (AAD): after the attacker sees the published model and chooses an attack, the defender uses additional computational resources to produce its final response. 
Our constructed learning task, under cryptographic assumptions, admits an AAD but no AD within the specified resource bounds.
Thus it constitutes a formal evidence that increasing inference-time resources can improve robustness.

We also show that the inference phase of the AAD is, in particular, a defender against TAs
within the relevant resource bounds.
It follows that our task admits AAD but neither a TA nor an AD, and this proves that the answer to \emph{Question 1} for generative tasks is negative as advertised. 





\subsection{Model and Definitions}\label{sec:intro-model}
\begin{figure}[t]
\centering
\begingroup
\renewcommand{\figscale}{1.0619}
\colorlet{protoBlue}{figBlue}
\colorlet{protoRed}{figRed}
\colorlet{protoGray}{figGray}
\colorlet{protoLine}{figGray!45}
\colorlet{protoInk}{black}
\colorlet{protoPanel}{figGray!6}

\resizebox{\textwidth}{!}{%
\begin{tikzpicture}[
  x=1cm,y=.80cm,
  font=\figtextfont,
  text=protoInk,
  >=Latex,
  detail/.style={
    font=\fignotefont,
    text=black
  },
  compute/.style={
    draw=protoBlue,
    fill=protoBlue!18,
    rounded corners=2pt,
    figbox,
    text width=1.85cm,
    minimum height=.52cm,
    align=center,
    font=\figtextfont,
    inner sep=2.5pt
  },
  attackstep/.style={
    compute,
    draw=protoRed,
    fill=protoRed!18,
    text width=1.70cm,
    minimum height=.54cm
  },
  wire/.style={
    draw=protoLine,
    figthin
  },
  message/.style={
    ->,
    draw=protoBlue,
    figarrow
  },
  challenge/.style={
    ->,
    draw=protoRed,
    figarrow
  },
  private/.style={
    ->,
    draw=protoBlue,
    figbox,
    dash pattern=on 2.3pt off 1.8pt,
    rounded corners=3pt
  },
  msglabel/.style={
    above,
    inner sep=2pt
  },
  partylabel/.style={
    align=center,
    inner sep=0pt,
    anchor=base,
    font=\figtextfont
  },
  fixedparty/.style={
    figfixed
  },
  pics/protoSampleCloud/.style={code={
    \begin{scope}[x=.85cm,y=.85cm]
      \path[
        draw=protoGray!70,
        fill=white,
        figthin,
        rounded corners=0pt
      ]
        (-.23,-.12)
        .. controls (-.35,-.13) and (-.40,.03) .. (-.28,.10)
        .. controls (-.30,.22) and (-.12,.26) .. (-.05,.17)
        .. controls (.02,.28) and (.20,.24) .. (.21,.13)
        .. controls (.35,.16) and (.41,-.01) .. (.29,-.09)
        .. controls (.27,-.21) and (.10,-.23) .. (.03,-.15)
        .. controls (-.05,-.24) and (-.23,-.24) .. (-.23,-.12)
        -- cycle;
      \node[
        shape=rectangle,
        draw=none,
        fill=none,
        minimum width=0pt,
        minimum height=0pt,
        inner sep=0pt,
        outer sep=0pt,
        font=\fignotefont,
        text=black
      ] at (0,0) {$\dist$};
    \end{scope}
  }}
]

\foreach \origin/\kind in {0/1,4.97/2,9.94/3}{
\begin{scope}[xshift=\origin cm]

  \path[
    rounded corners=4pt,
    fill=protoPanel,
    draw=protoLine,
    figthin
  ] (0,0) rectangle (4.70,7.00);

  \ifnum\kind=1
    \pic at (2.35,6.42) {figpanel={protoBlue}{AD}};
  \fi
  \ifnum\kind=2
    \pic at (2.35,6.42) {figpanel={protoBlue}{AAD}};
  \fi
  \ifnum\kind=3
    \pic at (2.35,6.42) {figpanel={protoRed}{TA}};
  \fi

  \ifnum\kind=3
    \node[partylabel,anchor=south,text=black] at (1.20,4.98)
      {\textbf{Fixed}\\attacker $\ver$};
    \node[partylabel,anchor=south,text=black] at (3.14,4.98)
      {Any\\defender $\prov$};
  \else
    \node[partylabel,anchor=south,text=black] at (1.20,4.98)
      {Any\\attacker $\ver$};
    \node[partylabel,anchor=south,text=black] at (3.14,4.98)
      {\textbf{Fixed}\\defender $\prov$};
  \fi

  \draw[wire] (1.20,4.88) -- (1.20,.42);
  \draw[wire] (3.14,4.88) -- (3.14,.42);

  \ifnum\kind<3
    \node[compute,fixedparty] (training) at (3.14,4.36) {Train};
    \pic at ([xshift=1pt,yshift=-1pt]training.north west)
      {protoSampleCloud};

    \draw[message] (3.14,3.53) --
      node[msglabel] {$f,R$} (1.20,3.53);
  \fi

  \ifnum\kind=3
    \node[attackstep,fixedparty] (attack) at (1.20,2.84)
      {Generate $\xs$};
  \else
    \node[attackstep] (attack) at (1.20,2.84)
      {Generate $\xs$};
  \fi
  \pic at ([xshift=-1pt,yshift=-1pt]attack.north east)
    {protoSampleCloud};

  \draw[challenge] (1.20,2.16) --
    node[msglabel] {$\xs$} (3.14,2.16);

  \ifnum\kind=1
    \node[compute,fixedparty] (answer) at (3.14,1.53)
      {Evaluate $R$};
  \fi
  \ifnum\kind=2
    \node[compute,fixedparty] (answer) at (3.14,1.53)
      {Inference};
  \fi
  \ifnum\kind=3
    \node[compute] (answer) at (3.14,1.53)
      {Respond};
  \fi

  \ifnum\kind>1
    \pic at ([xshift=1pt,yshift=-1pt]answer.north west)
      {protoSampleCloud};
  \fi

  \ifnum\kind<3
    \draw[private]
      (training.east) --
      (4.43,4.36) --
      (4.43,1.53) --
      (answer.east);

    \node[
      text=black,
      anchor=east,
      fill=protoPanel,
      inner sep=1.5pt
    ] at (4.35,2.99) {$\priv$};
  \fi

  \ifnum\kind=1
    \draw[message] (3.14,.54) --
      node[msglabel]
        {$\ys=f(\xs)$,\ $\rs=R(\priv,\xs)$}
      (1.20,.54);
  \fi
  \ifnum\kind=2
    \draw[message] (3.14,.54) --
      node[msglabel]
        {$\ys,\ \rs=R(\priv,\xs)$}
      (1.20,.54);
  \fi
  \ifnum\kind=3
    \draw[message] (3.14,.54) --
      node[msglabel] {$\ys,\rs$}
      (1.20,.54);
  \fi

\end{scope}
}

\draw[
  -{Latex[length=5pt,width=3.6pt]},
  draw=protoGray,
  figarrow
] (-.34,4.87) -- (-.34,.40);

\node[detail,rotate=90] at (-.67,2.63) {Time};

\pic at (2.03,-.45) {protoSampleCloud};
\node[detail,anchor=west] at (2.49,-.45)
  {Sample access};

\draw[private] (6.18,-.45) -- (6.94,-.45);
\node[detail,anchor=west] at (7.06,-.45)
  {Private state};

\node[detail,anchor=west] at (10.12,-.45)
  {$\ys$: answers, $\rs$: flags};

\end{tikzpicture}%
}
\endgroup

\caption{\textbf{ Protocols between the attacker \(\ver\) and the defender \(\prov\).} The fixed party must succeed against any opponent. In an AD, \(\prov\) trains with sample access, sends a model \(f\) and a distinguishing rule \(R\), and keeps a private string \(\priv\). Then \(\ver\) sends a batch \(\xs\), and \(\prov\) answers \(\ys=f(\xs)\) and flags \(\rs=R(\priv,\xs)\). An AAD has the same training phase, but its inference phase computes the answers \(\ys\) after seeing \(\xs\), again with sample access. In a TA, \(\ver\) sends \(\xs\) without seeing a model, and every defender answers and flags it.}
\label{fig:definition-protocols}
\end{figure}
An error oracle is a function \(h : \X \times \Y \to \{0,1\}\),
that takes a pair \((x,y)\), and outputs \(0\) if \(y\) is a valid answer for \(x\), and \(1\) otherwise.
A model is a potentially randomized function \(f : \X \to \Y\).
A distinguishing rule is a potentially randomized function \(R: \X \to \{0,1\}\), and we call its output \(r = R(x)\) a flag.
For a distribution \(\dist \in \Delta(\X \times \Y)\) we denote its marginal distribution over \(\X\) by \(\dist_{\X}\).
We define the error of a model \(f\) as \(\err(f) := \Pr_{x \gets \dist_{\X}} \big[ h(x,f(x)) = 1 \big]\).
For questions \(\xs \in \X^q\), answers \(\ys \in \Y^q\) and flags \(\rs \in \{0,1\}^q\) we also define the empirical error as \(\err_{\xs,h}(\ys) := \frac1q \sum_{i \in [q]} h(x_i,y_i)\)
and the flag rate as \(\fl(\rs) := \frac1q |\{ i \in [q] : r_i = 1 \}|.\)
\begin{definition}[Learning Task, informal version of Definition~\ref{def:task}]\label{def:task-informal}
    A learning task \(\task\) is a family \(\{\task_\prm\}_{\prm \in \N}\), where for every \(\prm\), \(\task_\prm\) is an element of \(\Delta\big(\Delta(\inspace \times \outspace) \times \{0,1\}^{\inspace \times \outspace}\big)\).
\end{definition}

We model efficient computation by non-uniform families of Boolean circuits of a certain size. Our circuits can read uniformly random bits, and they query the example oracle \(\ex\) that samples \((x,y) \gets \dist\) and returns \((x,y)\).
Before the interaction between a defender and an attacker starts, \((\dist,h)\) is sampled.
The interaction is modeled by two- and three-round interactive protocols \citep{ip_goldwasser} between Alice and Bob. Alice is the attacker \(\ver=\{\ver_\prm\}_\prm\) and Bob is the defender \(\prov=\{\prov_\prm\}_\prm\) of sizes \(S_\ver\) and \(S_\prov\).

Now recall the definition of probably approximately correct (PAC) learnability \citep{valiant1984, kearns1994introduction}. Taking the computational aspect of learning algorithms into account, \citet{gluch2026good} defined computationally bounded learnability.
\begin{definition}[Computationally Bounded Learnability, informal version of Definition~\ref{def:learnability}]\label{def:learnability-informal}
   We say that a learning task is learnable by circuits of size \(S\), if there exists a circuit family of size \(S\), which outputs \(f\),\footnote{Formally, strings \(z_f\) represent \(f\). Evaluation circuits on input \(z_f\) and \(x\) output \(f(x)\) (see Section~\ref{sec:circuits}). There is also an upper bound on the size of the evaluation circuit.} such that for every \(\prm\) large enough, with high probability, \(\err(f)\) is small.
\end{definition}
Now consider Figure~\ref{fig:definition-protocols}. A TA produces a batch \(\xs\) that fools \emph{all} computationally bounded defenders, none of which can distinguish it from a natural batch.
\begin{definition}[Transferable Attack, informal version of Definition~\ref{def:transferableattack}]
    A family of circuits \(\ver_{\TA}\)
    of size \(S_\ver\) implements a TA against size \(S_\prov\) defenders for a learning task \(\task\),
    if an interactive protocol in which
   \(\ver_{\TA}\) computes \(\xs\) and an arbitrary \(\prov\)
    based on \(\xs\) outputs \((\ys, \rs)\) satisfies both of the following: For circuits \(\prov\) of size \(S_\prov\)
    \begin{itemize}[topsep=0pt, itemsep=0pt, parsep=0pt]
        \item \textbf{Transferability.} \(\prov\) answers the batch that \(\ver\) sends mostly incorrectly: the expected error \(\err_{\xs,h}(\ys)\) is large.
        \item \textbf{Indistinguishability.} \(\prov\) flags the batch that \(\ver\) sends about as often as a natural batch \(\xs\gets (\dist_{\X})^q\): the expected flag rates \(\fl(\rs)\) on the two batches are almost the same.
    \end{itemize}
\end{definition}
Notice that Transferability alone is typically easy: for a task whose ground truth is nontrivial outside the support of the natural distribution, Alice can send a batch from outside the support that no small circuit answers correctly. Indistinguishability alone is also easy, since Alice can forward a natural batch.

An AD publishes a low-error model \(f\) and a distinguishing rule \(R\), such that \emph{every} computationally bounded adversary either fails to fool \(f\) or is distinguished by \(R\). 
Notice that the two notions are not negations of one another. The absence of a TA says only that each adversary is withstood by \emph{some} defender, chosen once that adversary is fixed and possibly a different one for each, whereas an AD asks for \emph{one} defender that withstands all attacks.

\begin{definition}[Adversarial Defense, informal version of Definition~\ref{def:detectiondefense}]
    A family of circuits \(\prov_{\AD}\)
    of size \(S_\prov\) implements an AD against size \(S_\ver\) adversaries for a learning task \(\task\),
    if an interactive protocol in which \(\prov_{\AD}\) computes a model \(f\), a distinguishing rule \(R\) and a private string \(\priv\) that instantiates \(R\), sends \(f\) and \(R\) to \(\ver_\prm\), which replies with \(\xs\), satisfies:
    \begin{itemize}[topsep=0pt, itemsep=0pt, parsep=0pt]
        \item \textbf{Correctness.} With high probability, the model has low error: \(\err(f)\) is small.
    \end{itemize}
    Moreover, for every circuit \(\ver\) of size \(S_\ver\) at least one of the following holds.
    \begin{itemize}[topsep=0pt, itemsep=0pt, parsep=0pt]
        \item \textbf{Robustness.} The model answers the questions of \(\ver\) mostly correctly: the expected error \(\err_{\xs,h}(f(\xs))\) is small.
        \item \textbf{Distinguishability.} The rule flags the questions of \(\ver\) at a different rate than natural questions: the expected flag rates \(\fl(R(\priv,\xs))\) on the two batches differ noticeably.
    \end{itemize}
\end{definition}

\begin{figure}[H]
\centering
\begingroup
\renewcommand{\figscale}{1.0970}
\colorlet{protoBlue}{figBlue}
\colorlet{protoRed}{figRed}
\colorlet{protoGray}{figGray}
\colorlet{protoLine}{figGray!45}
\resizebox{\textwidth}{!}{%
\begin{tikzpicture}[
  x=1cm,y=1.08cm,
  font=\figtextfont,
  text=black,
  >=Latex,
  line cap=round,line join=round,
  detail/.style={
    font=\fontsize{10}{11.5}\selectfont,
    text=black
  },
  track/.style={
    fill=protoGray!6,draw=protoLine,figthin,
    rounded corners=2pt
  },
  exists/.style={
    draw=protoGray,figbox,rounded corners=2pt
  },
  neither/.style={
    fill=protoGray!18,draw=protoGray,figbox,
    rounded corners=2pt
  },
  guide/.style={
    figbox,dash pattern=on 2.3pt off 1.8pt
  }
]
\path[use as bounding box] (0,.08) rectangle (15.05,2.80);

\node[detail] at (7.52,2.55)
  {\textbf{Fixed party:} size $S$
   \qquad
   \textbf{Every opponent:} size $S^{1/\rho}$};

\node[anchor=west,align=left,inner sep=0pt] at (.06,1.86)
  {{\fontsize{10.5}{12}\selectfont\bfseries Classification tasks}\\[1pt]
   {\fontsize{9}{10.5}\selectfont General guarantee}};

\node[anchor=west,align=left,inner sep=0pt] at (.06,1.10)
  {{\fontsize{10.5}{12}\selectfont\bfseries Generative tasks}\\[1pt]
   {\fontsize{9}{10.5}\selectfont Cryptographic construction}};

\begin{scope}[xshift=.90cm]

\path[track] (3.65,1.57) rectangle (13.57,2.15);
\begin{scope}
  \clip[rounded corners=2pt]
    (8.25,1.57) rectangle (13.57,2.15);
  \fill[protoBlue!30]
    (8.25,1.57) rectangle (13.57,2.15);
  \foreach \k in {0,...,10}{
    \fill[protoRed!30]
      ({7.55+\k*.60},1.57)
      -- ++(.30,0)
      -- ++(.626,.58)
      -- ++(-.30,0)
      -- cycle;
  }
\end{scope}
\path[exists] (8.25,1.57) rectangle (13.57,2.15);
\node at (10.85,1.86) {TA or AD exists};
\draw[->,draw=protoGray,figarrow]
  (13.05,1.86)--(13.43,1.86);
\node at (5.95,1.86) {?};

\path[track] (3.65,.81) rectangle (13.57,1.39);
\path[neither] (3.65,.81) rectangle (7.79,1.39);
\node[align=center] at (5.72,1.10)
  {Neither TA nor AD\\
   AAD exists};
\node at (10.68,1.10) {?};

\draw[guide,draw=protoGray] (8.25,.52)--(8.25,2.23);
\draw[guide,draw=protoGray] (7.79,.52)--(7.79,1.47);

\draw[->,draw=protoGray,figarrow]
  (3.65,.52)--(13.76,.52);
\node[detail,anchor=west] at (13.80,.52) {$\rho$};

\foreach \x/\label in {3.65/1,7.79/\beta,8.25/2,12.85/3}{
  \draw[draw=protoGray,figthin]
    (\x,.52)--(\x,.42);
  \node[anchor=north,inner sep=2pt] at (\x,.39)
    {$\label$};
}

\end{scope}
\end{tikzpicture}%
}
\endgroup
\caption{\textbf{Scaling for adversarial robustness.}
A fixed party of size \(S\) initiates the protocol and wins against every opposing party of size \(S^{1/\rho}\): if the fixed party is an attacker, its opponents are all defenders, and vice versa. Here \(\rho \geq 1\) is the relative size exponent.
In the classification setting, for \(\rho > 2\) every task admits a TA or an AD.\protect\footnotemark{} In the generative case, for \(\rho < \beta\) there is a task admitting neither a TA nor an AD, while an AAD exists.\protect\footnotemark}
\label{fig:robustness-scaling-laws}
\end{figure}
\addtocounter{footnote}{-1}
\footnotetext{Here we assume that the task is learnable by circuits
of size \(\sqrt{S}/(4\log S)\).}
\stepcounter{footnote}
\footnotetext{Here \(\beta<2\) is given by
\(S^{1/\beta}=\poly(\prm)\cdot S^{1/(2\eta)}\cdot S_E\),
with \(\eta\) and \(S_E\) as in
Theorem~\ref{thm:mitdetgentime_informal}.}

An AAD has an additional inference-time phase, in which Bob uses additional resources to answer questions instead of just using \(f\).

\begin{definition}[Active Adversarial Defense, informal version of Definition~\ref{def:mitigationdefense}]
    A pair of circuit families \(\prov_{\AAD}= (\provtrain, \provfix)\)
    of sizes \(\ttrain\) and \(\tfix\) implements an AAD against size \(S_\ver\) adversaries for a learning task \(\task\),
    if an interactive protocol in which \(\provtrain\) computes a model \(f\), a distinguishing rule \(R\) and a private string \(\priv\) that instantiates \(R\), sends \(f\) and \(R\) to \(\ver\), which replies with \(\xs\), and \(\provfix\) on input \(\priv\) and \(\xs\) outputs answers \(\ys\) and the flags \(\rs = R(\priv, \xs)\), satisfies:
    \begin{itemize}[topsep=0pt, itemsep=0pt, parsep=0pt]
        \item \textbf{Correctness.} With high probability, the model has low error: \(\err(f)\) is small.
        \item \textbf{Completeness.} With high probability, the inference phase answers a natural batch \(\xs\gets (\dist_{\X})^q\) mostly correctly: \(\err_{\xs,h}(\ys)\) is small.
    \end{itemize}
    Moreover, for every circuit \(\ver\) of size \(S_\ver\) at least one of the following holds.
    \begin{itemize}[topsep=0pt, itemsep=0pt, parsep=0pt]
        \item \textbf{Robustness.} The inference phase answers the batch of \(\ver\) mostly correctly: the expected error \(\err_{\xs,h}(\ys)\) is small.
        \item \textbf{Distinguishability.} The rule flags the batch of \(\ver\) at a different rate than a natural batch: the expected flag rates \(\fl(R(\priv,\xs))\) on the two batches differ noticeably.
    \end{itemize}
\end{definition}

Notice that the definition of AAD \say{interpolates} between AD and TA in the following sense.
At the one extreme if $\tfix = 0$, then the definition of AAD becomes equivalent\footnote{Where in the definition of Robustness in AAD we replace $\ys$ with $f(\xs)$.} to that of AD.
At the other extreme if $\ttrain = 0$ then the existence of AAD implies\footnote{Formally AAD with $\ttrain = 0$ can't exists because the correctness property requires $f$ produced by $\provtrain$ to have low-error but if its size is $0$ it's impossible.} that TA doesn't exist.
In between the two extremes there's a more nuanced relationship, namely AAD implies that TA doesn't exist, where the defender's size is $\ttrain + \tfix$ and the attacker size is $S_\mathsf{attack}$.

Note that AAD's inference procedure ($\provfix$) could in principle be sent (maybe composed with the $f$ produced by $\provtrain$) as an $f$ for a construction of a succesful AD.
If it was possible AAD would be equivalent to AD.
However, it is only possible if the size of $\provfix$ fits inside the bound on the size of $f$ (see Section~\ref{sec:circuits} for details on what models $f$ are allowed to be sent and how they are represented).


\subsection{Our Results}\label{sec:results}

Figure~\ref{fig:robustness-scaling-laws} summarizes our results
as a function of the computational resources available to the
competing parties.
We fix one party's size at $S$ and bound its
opponents' size by $S^{1/\rho}$. 

\paragraph{Classification: A Dichotomy at a Quadratic Gap.} We establish a dichotomy answering \emph{Question 1} in the classification setting.
We prove that a learnable classification task either has an attack that fools every sufficiently small defender, or there is a defense that holds up against every sufficiently small attacker. This is a win-win statement similar in spirit to \citet{degwekar2019computational}.

\begin{lemma}[Dichotomy, informal version of Lemma~\ref{lem:dichotomy}]\label{lem:dichotomy_informal}
For every size function \(S\) and every classification task that is learnable by circuits of size \(\sqrt{S}/(4\log S)\), with models that have evaluation circuits of size \(\leq S\), at least one of the following exists:
\begin{enumerate}[label=(\roman*), leftmargin=2em, topsep=0pt, itemsep=0pt, parsep=0pt]
    \item a TA of size \(O(S)\) against defenders of size \(o(\sqrt{S})\), or
    \item an AD of size \(O(S)\) against attackers of size \(o(\sqrt{S})\).
\end{enumerate}
\end{lemma}

As mentioned, \citet{gluch2026good} showed that every classification task admits at least one of three schemes: a TA, an AD, or a WM. The WM is a claim of ownership: it lets whoever trained the model prove that the model is theirs, so that nobody who copies it can pass it off as their own.
We note that \citet{gluch2026good} have a slightly different notion of an AD: it must rarely flag natural inputs and often flag attacks. Ours asks that the two flag rates differ. 
We observed that there is a way to turn their AD into our AD and their WM into a TA against the same defenders. 
This means that their trichotomy also gives an ``attack or defense'' dichotomy. However, this dichotomy is \textit{at much worse or even vacuous parameters}.

Their result considers constant robustness and transferability parameters. A close analysis shows that it is possible to attain one degree of freedom in their parameters. Our result has four, the constants \(c, s, \alpha\) and \(\kappa\) of Lemma~\ref{lem:dichotomy}, which move the two alternatives independently. 
Running their argument with a free threshold \(1-c\) yields a TA with transferability \(c\) and indistinguishability \(1-2c\), or an AD with robustness \(4c+4\delta\), where \(\delta\) is the confidence parameter. Notice that Indistinguishability of an attack and Robustness of a defense sum to \(1+2c+4\delta>1\), so whichever threshold is chosen, one of the two options gives weak guarantees. A small \(c\) gives an attack whose inputs may be flagged at a rate far from that of natural inputs, and a \(c\) close to \(1/4\) gives a defense with no robustness guarantee. 
We compare their results to our results in Table~\ref{tab:trichotomy}. The derivation of the values is in Appendix~\ref{app:trichotomy}.
\begin{table}[t]
  \centering
  \small
  \setlength{\tabcolsep}{7pt}
  \renewcommand{\arraystretch}{1.1}
  \begin{threeparttable}
    \begin{tabular}{@{}l
      S[table-format=1.2]
      S[table-format=1.2]
      S[table-format=1.2]
      S[table-format=1.5]@{}}
      \toprule
      & \multicolumn{2}{c}{Attack}
      & \multicolumn{2}{c}{Defense} \\
      \cmidrule(lr){2-3}\cmidrule(lr){4-5}
      Result & {Transferability~\(\uparrow\)} & {Indistinguishability~\(\downarrow\)}
             & {Robustness~\(\downarrow\)} & {Distinguishability~\(\uparrow\)} \\
      \midrule
      \citet{gluch2026good}
        & 0.21 & 0.58 & 0.92 & {\(5 \times 10^{-4}\)} \\
      Lemma~\ref{lem:dichotomy} 
        & 0.21 & {\boldmath\(0.02\)} & {\boldmath\(0.54\)} & {\boldmath\(0.001\)} \\
      \addlinespace[7pt]
      \citet{gluch2026good}
        & 0.10 & 0.80 & 0.44 & {\(5 \times 10^{-6}\)} \\
      Lemma~\ref{lem:dichotomy} 
        & 0.10 & {\boldmath\(0.01\)} & {\boldmath\(0.25\)} & {\boldmath\(0.00005\)} \\
      \bottomrule
    \end{tabular}
  \end{threeparttable}
\caption{\textbf{Comparing dichotomy results.}
Each pair of rows compares \citet{gluch2026good} with our dichotomy result.\protect\footnotemark{}
Arrows show whether larger (\(\uparrow\)) or smaller (\(\downarrow\)) values are better, and bold marks the better value in each pair of rows.}
\label{tab:trichotomy}
\end{table}
\footnotetext{The fourth row is Lemma~\ref{lem:dichotomy} at the parameters of Corollary~\ref{cor:dichotomy}.}
\vspace{-2mm}

\paragraph{Classification: Transferable Attacks from Weaker Assumptions.} We give an explicit construction of a classification task admitting a TA. Our construction requires a secure public-key encryption (PKE) \citep{Goldwasser1984ProbabilisticE}, weakening the cryptographic assumption used by \citet{gluch2026good}, whose construction relied on fully homomorphic encryption (FHE). We then show how to lift \emph{any} classification task to a task that, under the same cryptographic assumption, admits a TA (see Section~\ref{sec:constructions}).

\paragraph{Generation: Active Adversarial Defenses Are Strictly Stronger.} In the generative setting, the following two theorems say that AADs are strictly stronger than ADs. On each task, an AAD exists while no AD with the same training budget does. 

\begin{theorem}[Running-Time Separation, informal version of Theorem~\ref{thm:mitdetgentime}]\label{thm:mitdetgentime_informal}
For every \(S\) there exists a generative task such that, under cryptographic assumptions, the following holds for every evaluation size \(S_E \ll \sqrt{S}\) of the published model:
\begin{enumerate}[label=(\roman*), leftmargin=2em, topsep=0pt, itemsep=0pt, parsep=0pt]
    \item There is an AAD with \(\ttrain = O(S)\) and \(\tfix = \sqrt{S}\cdot\poly(\prm)\), against attackers of size \(O(S)\).
    \item There is no AD with size \(O(S)\), even against attackers of size \(\sqrt{S} \cdot S_E\cdot\poly(\prm)\).\footnote{The formal exponent of \(S\) is \(1/(2\eta)\), where \(\eta<1\) is the hardness gap of the underlying non-parallelizing language.}
\end{enumerate}
\end{theorem}

\begin{theorem}[Sample-Complexity Separation, informal version of Theorem~\ref{thm:mitdetgen}]\label{thm:mitdetgen_informal}
There exists
a generative task such that, under cryptographic assumptions, 
for every 
sample complexity \(K\):
\begin{enumerate}[label=(\roman*), leftmargin=2em, topsep=0pt, itemsep=0pt, parsep=0pt]
    \item There is an AAD 
    with sample complexity \(\ktrain = O(K)\) and \(\kfix = O(\sqrt{K})\),
    against attackers that are restricted to sample
    complexity \(O(K)\).
    \item There is no AD with sample
    complexity \(O(K)\) against
    attackers with sample complexity
    \(O(\sqrt{K}\,\log^2\prm)\).
\end{enumerate}
\end{theorem}

Our constructions make use of several cryptographic primitives: identity-based fully homomorphic encryption (IB-FHE) \citep{gentryIB-FHE, clear2014bootstrappable}, zero-knowledge succinct non-interactive arguments of knowledge (zk-SNARKs) \citep{Alefullysuccinctsnarks}, digital signatures \citep{bonehsignatures}, non-parallelizing languages (NPLs) \citep{tlp}, and incrementally verifiable computation (IVC) \citep{valiant2008incrementally, bitansky2013recursive}.\footnote{Our focus is on demonstrating this limitation, rather than minimizing the cryptographic assumptions.} NPLs together with IVC play the role of a continuous verifiable delay function \citep{ephraim2020continuous}.

\paragraph{Generation: No Dichotomy Below a Quadratic Gap.}
It turns out that the answer to \emph{Question 1} in the generative setting below a quadratic gap in computational resources is different from that given by Lemma~\ref{lem:dichotomy} in the classification setting at a quadratic gap. This answer follows from Theorem~\ref{thm:mitdetgentime_informal}. The inference phase of its AAD, run on its own, is a defender that no attack of size \(O(S)\) fools, so the task admits no TA, and by item (ii) of Theorem~\ref{thm:mitdetgentime_informal} it admits no AD.

\begin{corollary}[Neither a Transferable Attack nor an Adversarial Defense, informal version of Corollary~\ref{cor:neithertime}]\label{cor:neithertime_informal}
For every \(S\) and every evaluation size \(S_E \ll \sqrt{S}\) of the published model, the task of Theorem~\ref{thm:mitdetgentime_informal} satisfies:
\begin{enumerate}[label=(\roman*), leftmargin=2em, topsep=0pt, itemsep=0pt, parsep=0pt]
    \item There is no TA of size \(O(S)\) against defenders of size \(\sqrt{S}\cdot\poly(\prm)\).
    \item There is no AD with size \(O(S)\), even against attackers of size \(\sqrt{S} \cdot S_E\cdot\poly(\prm)\).
\end{enumerate}
\end{corollary}

Figure~\ref{fig:robustness-scaling-laws} shows whether a TA or an AD exists depending on the resources available to the parties.
Our results suggest that the \emph{relative size exponent} \(\rho\), with a fixed party of size \(S\) and opponents of size \(S^{1/\rho}\), is a key parameter for this classification.

\subsection{Discussion and Open Questions}\label{sec:discussion}
We showed that every learnable classification task admits a TA or an AD once the stronger party has a quadratic resource advantage over its opponent. For generative tasks the picture changes. We constructed a task that admits neither at a similar resource gap, while there is a successful AAD that spends the square root of its training budget during inference.
Inference-time computation can thus improve robustness that no fixed model can provide, which is in line with \citet{boaz2025trading}.
These results taken together strongly suggest that the relative size exponent is a key parameter for determining whether a TA or an AD exists for a given task. 
Two main questions whose answers would fill the empty parts of Figure~\ref{fig:robustness-scaling-laws} remain open.

\vspace{2mm}

\emph{Open Question 1.} \emph{Does every learnable generative task admit a TA or an AD at a quadratic, or at least some polynomial, resource advantage?}

\vspace{2mm}

\emph{Open Question 2.} \emph{Is there a classification task that admits neither a TA nor an AD when the fixed party has less than a quadratic resource advantage?}

\vspace{2mm}

In Section \ref{sec:overview} we give a justification on why these questions cannot be answered by extending results in the classification setting to the generative case and vice versa. 
\subsection{Related Work}\label{sec:related-work}
\paragraph{Robustness and Adversarial Examples.}
The field of adversarial robustness has a rich literature
\citep{intriguingprop, adversarialspheres, certifieddefense, wongprovable, Engstrom2017ARA}.
For classification tasks, adversarial training \citep{adversarialtraining}, which trains the model on adversarial examples, has shown empirical promise, and certified defenses \citep{certifieddefense, wongprovable} provide provable guarantees within a specified perturbation bound.
Both address bounded perturbations of natural inputs, whereas the attacker in our model may submit arbitrary inputs.

\paragraph{A Computational Lens on Adversarial Examples.}
The view that adversarial vulnerability can be a computational rather than a statistical phenomenon goes back to \citet{bubeck2019adversarial}, who exhibit learning tasks on which a robust classifier exists but cannot be found by efficient learners \citep[see also][]{bubeck2018adversarial}, and to \citet{degwekar2019computational}, who prove such separations under cryptographic assumptions and complement them with ``win-win'' results, in which the failure of efficient robust learning yields a cryptographic primitive.
\citet{mahloujifar2019can} identify settings in which polynomial-time attackers are as strong as unbounded ones, so that computational hardness cannot help the defender, whereas \citet{gargcompadvexp} use digital signatures to build classifiers that are robust against polynomial-time attackers although no classifier is robust against unbounded ones.
These works consider bounded perturbations of natural inputs and ask whether a robust classifier exists or can be found efficiently.
We consider arbitrary inputs, let the defender flag inputs, ask whether a single attack works against a whole class of bounded defenders, and measure the answer in explicit circuit-size exponents.
The dichotomy of Lemma~\ref{lem:dichotomy} is a win-win statement in the sense of \citet{degwekar2019computational}, with a transferable attack in place of a cryptographic primitive.
The resource-bounded protocol view that we adopt is due to \citet{gluch2026good}, who proved the trichotomy between TAs, ADs and WMs that Lemma~\ref{lem:dichotomy} refines, constructed a TA from FHE, and showed that the TA alternative cannot be removed in general.

\paragraph{Transferability of Adversarial Examples.}
One of the most intriguing facts about adversarial examples is their \emph{transferability}: adversarial examples crafted for one model often transfer to other architectures or training sets.
This phenomenon was discovered early and studied extensively \citep{goodfellow2014explaining, Papernot2016TransferabilityIM, liu2016delving} and more recently was observed in the context of LLMs \citep{Zou2023UniversalAT}.
\citet{NEURIPS2019_e2c420d9} hypothesize that adversarial examples exist because of highly predictive but non-robust features in the data, which would explain transferability.
Universal perturbations \citep{moosavi2017universal} fool one model on most inputs. A transferable attack in our sense is the dual object, one input distribution that fools every model in a class.
\citet{ball2025impossibilityseparatingintelligencejudgment} show, for concrete language models, that adversarial prompts can be computationally indistinguishable from benign ones for every efficient filter, which is the indistinguishability requirement of a transferable attack against filters.
\citet{gluch2026good} showed that the existence of ADs and TAs depends critically on the computational resources of the defender and the attacker, treated in aggregate.
We separate training-time from inference-time resources.
For classification tasks, an attack with a quadratic resource advantage transfers to all bounded models unless a defense exists (Lemma~\ref{lem:dichotomy}).
On our generative task, every published model with bounded evaluation cost can be attacked, yet no attack transfers to an active defender that spends about the square root of the training budget at inference (Corollary~\ref{cor:neithertime}).

\paragraph{Game-Theoretic Formulations.}
Zero-sum games between attackers and classifiers have been used to characterize the minimax adversarial risk \citep{pydi2020adversarial}, to show that pure equilibria need not exist and that randomized classifiers help \citep{pinot2020randomization, meunier2021mixed}, and to identify particular attack--defense pairs as equilibria \citep{pal2020game}.
Closest to our transferable attacks is the adversarial example game of \citet{bose2020adversarial}, in which a generator is trained to fool every classifier in a hypothesis class and the optimal generator is characterized by a min-max equilibrium.
In those games the strategies range over perturbation sets and parametric model classes with a fixed loss.
In ours the pure strategies are all circuits of a given size, the payoff combines correctness with a distinguishing term, and the sparse approximate equilibria of \citet{lipton1994simple} and \citet{althofer1994sparse} are what turn an equilibrium into a circuit of bounded size.
The uniform mixture of \(k\) attackers produced by Lemma~\ref{lem:dichotomy} is the theoretical counterpart of the ensemble-based transfer attacks of \citet{liu2016delving}.

\paragraph{Backdoors, Watermarks and Obfuscation.}
\citet{plantingbackdoors} plant backdoors in models that no efficient party can detect, and \citet{goldwasseroblivious} show how to \emph{mitigate} even such a backdoor by exploiting local self-correction properties of the ground truth for linear and low-degree polynomial regression tasks.\footnote{It is important to note that \citet{goldwasseroblivious} and our work consider different models.
In our work, the party that returns the answer is the same as the one that produced the model, and aims to compute a model without introducing backdoors.
In contrast, in \citet{goldwasseroblivious}, the model may be produced by a different party, one that is actively attempting to insert backdoors.
This key difference explains why the conclusions of the two works are compatible.}
\citet{christiano2024backdoor} define defendability against backdoors as a game between an attacker who plants a backdoor and a defender who must detect it at evaluation time, and relate it to PAC learnability and obfuscation.
\citet{gluch2026good} treat backdoor-based watermarks as a third alternative to ADs and TAs. Lemma~\ref{lem:dichotomy} shows that at a quadratic resource gap this alternative is not needed.

\paragraph{Detection, Abstention and Defenses That Act at Test Time.}
Selective classification lets a classifier abstain on some inputs \citep{el2010foundations}. Our flag plays that role, except that a flagged entry still counts as an error when it is answered incorrectly.
\citet{arbitraryexamples} study the transductive setting in which the learner sees the unlabeled test batch, may abstain, and must have low error on the points it labels and a low abstention rate on natural data, against an unbounded adversary and for classes of bounded Vapnik--Chervonenkis (VC) dimension.
\citet{montasser2022transductive} give transductive robust learning guarantees with robust error linear in the VC dimension, and \citet{chen2021towards} study transductive defenses empirically.
Our AD commits its model and its rule before seeing the input, is scored entry by entry, and bounds all parties computationally (see Section~\ref{sec:definitions}).
\citet{pmlr-v162-tramer22a} shows that in the bounded-perturbation model detecting adversarial examples is as hard as classifying them robustly, if computation is disregarded.
Our Robustness-or-Distinguishability alternative concerns arbitrary inputs, computationally bounded parties and a distributional flag rate rather than per-input detection, so that reduction does not apply. The coupling of the two alternatives in Lemma~\ref{lem:dichotomy} is its computational counterpart.
Randomized smoothing \citep{kolterrandomizedsmoothing} certifies \(\ell_2\) robustness by averaging a base classifier over Gaussian noise at prediction time and, in our language, is (partially) active.\footnote{What we mean by partially active is that there are two ways to look at smoothing.
Smoothing can be seen as performed during inference time, thus making the defense active, but it can also be ``hard-coded'' into \(f\) and then the defense would be considered as static according to our definitions.}
Adaptive test-time defenses such as adversarial purification \citep{nie2022diffusion}, test-time training \citep{sun2020test} and runtime masking \citep{pmlr-v119-wu20f} spend computation on the received input.
\citet{croce2022evaluating} found that, under an adaptive evaluation \citep{tramer2020adaptive}, none of the defenses they tested improved on static defenses, while all of them raised the cost of inference.
An AAD is the formal counterpart of these defenses.
Theorem~\ref{thm:mitdetgentime} exhibits a task on which computation after the input provably helps, and Section~\ref{sec:overview} explains the mechanism.

\paragraph{Inference-Time Computation and Robustness.}
\citet{boaz2025trading} report that giving reasoning models more inference-time computation improves their robustness across many attack families, with exceptions when the attacker can make the model reason less, divert its reasoning, or exploit ambiguous policies.
\citet{wu2025does} find that these gains depend on the reasoning trace being hidden from the attacker and can reverse when it is exposed, and \citet{mcdonald2026rich} find that inference-time computation helps mainly when training-time robustness already covers the attacked inputs.
Our notion of an AAD captures the abstract mechanism, computation spent after the input is known, and Theorem~\ref{thm:mitdetgentime} shows that it can provably help.
The defenses we construct do not build on the published model at inference (Section~\ref{sec:definitions}), so they model a separate certified computation rather than a model that reasons longer. Whether some task forces robust inference to reuse the trained model is an open question.

\paragraph{Cryptographic Tools.}
Time-lock puzzles \citep{rivest1996timelock, tlp} formalize computations that take \(t\) sequential steps, and verifiable delay functions \citep{boneh2018vdf} and continuous verifiable delay functions \citep{ephraim2020continuous} add that the result can be verified quickly.
A continuous verifiable delay function is iteratively sequential and iteratively verifiable, which is exactly properties 1 and 2 of Section~\ref{sec:overview}. The clear part of \(\tasktime\) is such a function, assembled from a non-parallelizing language \citep{tlp} and incrementally verifiable computation \citep{valiant2008incrementally, bitansky2013recursive}.
We keep the two primitives separate because our proofs need average-case hardness against non-uniform circuits with sample access, and evaluation of the certificates under identity-based FHE \citep{gentryIB-FHE}.
Knowledge soundness of SNARKs with respect to arbitrary auxiliary input is implausible under indistinguishability obfuscation \citep{bitansky2014extractable, boyle2015limits}. Lemma~\ref{lem:pool}, and hence Theorem~\ref{thm:mitdetgen}, uses extraction only with respect to the benign distribution of fresh signatures under an independently generated key and uniform coins.
Fully homomorphic encryption implies collision-resistant hashing \citep{ishai2005sufficient}, and there is no black-box construction of collision-resistant hashing from identity-based encryption \citep{mahmoody2016hierarchical}, and so none from public-key encryption, which has a black-box construction from identity-based encryption. Hence there is no black-box construction of FHE from public-key encryption. In this sense the assumption of Section~\ref{sec:constructions} is weaker than that of \citet{gluch2026good}.

\paragraph{Alignment and Jailbreaking.}
In the context of LLMs, \emph{alignment} aims at making sure that the models act consistently with human goals, e.g., by not answering dangerous prompts or not using offensive language, etc.
There is a fast-growing literature about trying to achieve alignment \citep{amodei2016concrete, greenblatt2024alignment, Ji2023AIAA}.
In consequence, there is a lot of research on trying to avoid alignment by designing \emph{jailbreaks} \citep{andriushchenko2024jailbreaking, chao2023jailbreaking, mehrotra2024tree, Wei2023JailbrokenHD}, which involves crafting prompts that bypass filters and protections embedded in the models.
There is also some interest in adversarial examples for LLMs \citep{Zou2023UniversalAT, Carlini2023AreAN, NEURIPS2023_a0054803}.

%% file: sections/overview.tex
\subsection{Dichotomy for Classification Tasks via a Zero-Sum Game}\label{sec:overview-dichotomy}
The proof of the dichotomy follows a game-theoretic approach. We consider a finite zero-sum game \citep{v1928theorie} between attackers and defenders and use sparse approximate optimal strategies \citep{lipton1994simple, althofer1994sparse} to obtain computationally bounded implementations.

\begin{proof}[Proof sketch of Lemma~\ref{lem:dichotomy_informal}] We view the interaction as a zero-sum game. Rows are indexed by size \(o(\sqrt{S})\) attackers \(\ver\), each producing an input \(x\), and columns by size \(o(\sqrt{S})\) defenders \(\prov\), each answering with a label \(y\) and a flag \(r\). The entry \(M_{\ver,\prov}\) rewards the defender, being the sum of the probability that \(\prov\) labels \(x \gets \ver\) correctly and of the distinguishability advantage of \(\prov\) against \(\ver\). The attacker plays \(\mathsf{Min}\) and wants a small sum, while the defender plays \(\mathsf{Max}\) and wants it large. By the Min-Max Theorem \citep{v1928theorie} the game has a value, and by a result of \citet{lipton1994simple} each player has a nearly optimal strategy that mixes only a few circuits and can therefore be implemented by a circuit of size \(O(S)\). Which player wins determines which notion we obtain.

If the value is small, the attacker wins. The nearly optimal mixture \(\ver_N\) keeps both terms small against every small defender, which is exactly a TA.

If the value is large, the defender wins, and we turn the nearly optimal mixture \(\prov_N\) into a defense \(\prov_{\AD}\), of size \(O(S+S_E)\) since it also evaluates the learned model. Bob outputs the classifier \(f\) produced by the learner. Rule \(R\) tosses a fair coin, which he keeps private, to decide between two checks: flag \(x\) if \(\prov_N\) flags it, or flag \(x\) if the label of \(\prov_N\) disagrees with \(f(x)\). A probabilistic analysis shows that \(\prov_{\AD}\) satisfies Distinguishability or Robustness.
\end{proof}

A careful analysis of the proof shows that uniqueness of labels is crucial for it to work. The defense flags a disagreement between \(\prov_N\) and \(f\), which signals an error only because two correct labels must agree. In a generative task they may give different valid answers, and the defender cannot check validity itself since it only sees samples. It is therefore not obvious how to extend the dichotomy to generative tasks, and Corollary~\ref{cor:neithertime} rules it out only below a quadratic advantage.

\subsection{Transferable Attacks from Public-Key Encryption}\label{sec:overview-pke}
The construction of a TA in \citet{gluch2026good} relies on FHE. In this work, ordinary secure PKE suffices, which is believed to be a strictly weaker assumption (see Section~\ref{sec:related-work}).

We construct a TA for a learning task \(\task_{\mathsf{PKE}}\) that is based on a PKE scheme \(\mathsf{PKE}=(\Gen, \Enc, \Dec)\) over the message space \(\mathcal{M}= \{0,1\}\).
\(\mathsf{PKE}\) is secure \citep{Goldwasser1984ProbabilisticE}, if any efficient adversary \(\mathsf{Adv}\) knowing only the public key \(\pk\) cannot decrypt \(c=\Enc(\pk, y)\): its guess is wrong with probability close to \(1/2\).

If we think of \((\pk, \Enc(\pk, y))\) as an adversarially chosen input \(x\) and of \(y\) as its true label, then security corresponds to the transferability of an attack producing such \(x\). Remember that we would like to construct a TA for a task that can be learned.
An easy solution is to add the plain message \(y\) to the description of \(x\), that is, to set \(x:=(\pk, y,\Enc(\pk, y))\) and keep \(y\) as the true label. Now the task is learnable: the bit next to the public key is the true label of \(x\). An adversary instead of appending \(y\), outputs \(x=(\pk, y',\Enc(\pk, y))\), where \(y\) and \(y'\) are independent uniformly random bits.
The modified attack is transferable, since the additional information \(y'\) does not help a defender guess the label \(y\) of \(x\). Moreover, an adversarially generated \(x\) looks like it came from the natural distribution by the security of \(\mathsf{PKE}\). See Section~\ref{sec:constructions} for a formal description of the task and the attack.

Now let \(\task\) be \emph{any} learning task. Building upon the above construction, we provide a way to append additional dimensions to \(x\in \{0,1\}^\lambda\) without changing the labels of natural inputs, such that the resulting learning task \(\task'\) admits a TA.

We continue working with a secure \(\mathsf{PKE}\) as before. Now let natural inputs be of the form \(x':=(\pk, x,\Enc(\pk, 0))\), where \(x \gets \dist_\lambda\). An arbitrary \(x'=(\pk', x, c) \in \{0,1\}^{\poly(\lambda)}\) receives the label \(h_\lambda(x)\oplus \Dec(\sk, c)\) if \(\pk'=\pk\), and zero otherwise.
The TA samples one natural input \((\pk, x,c)\) and outputs \(x':=(\pk, x, \Enc(\pk, b))\), where \(b\) is a uniformly random bit. Transferability and Indistinguishability again follow from the security of the \(\mathsf{PKE}\).

In Section~\ref{sec:constructions}, we also provide a construction of a learning task that admits an AD.

\subsection{Inference-Time Resources Let Active Adversarial Defenses Beat Adversarial Defenses}

\subsubsection{A Generative Task from Sequential Work Hidden by Encryption}\label{sec:overview-generative}
We now turn to generative tasks. We construct tasks that separate ADs from AADs: every AD within the stated training budget can be attacked, while an AAD using additional inference time can respond robustly. One of the tasks falls outside the dichotomy, admitting neither TA nor AD.

\begin{proof}[Proof sketch of Theorem~\ref{thm:mitdetgentime_informal}]

\begin{figure}[t]
\centering
\resizebox{\linewidth}{!}{%
\begingroup
\renewcommand{\figscale}{0.8767}
\colorlet{ltAttack}{figRed}
\colorlet{ltDefense}{figBlue}
\colorlet{ltGray}{figGray}
\colorlet{ltTask}{figGray!70!black}
\colorlet{ltSafe}{figBlue!15}
\begin{tikzpicture}[
  x=1cm,y=1.18cm,font=\normalfont\rmfamily\figtextfont,text=black,
  >={Latex[length=1.8mm,width=1.2mm]},line cap=round,line join=round,
  every node/.append style={text=black},
  flow/.style={->,figarrow},
  dot/.style={circle,inner sep=0pt,minimum size=4.5pt},
  enc/.style={rectangle,inner sep=0pt,minimum size=5pt,
              fill=white,figbox},
  note/.style={font=\normalfont\rmfamily\fignotefont,align=center},
  lab/.style={font=\normalfont\rmfamily\figtextfont,align=center},
  callout/.style={lab,rounded corners=3pt,figbox,text=black,
    inner xsep=5pt,inner ysep=5pt},
  tag/.style={fill=white,inner sep=2pt}
]
\pic at (2.4,5.52) {figpanel={ltGray}{Questions and answers}};
\pic at (8.20,5.52) {figpanel={ltAttack}{Attack on AD}};
\pic at (14.00,5.52) {figpanel={ltDefense}{AAD inference step}};
\node at (2.4,4.67) {\(\X=\Y\)};
\node[callout,fill=ltDefense!18,draw=ltDefense] at (6.55,4.67)
  {\(S\)\\training budget};
\node[callout,fill=ltAttack!18,draw=ltAttack] at (9.75,4.67)
  {\(O(\sqrt S)\)\\calls to \(f\) in total};
\node[callout,fill=ltDefense!18,draw=ltDefense] at (14.00,4.67)
  {\(O(\sqrt S)\)\\steps after the challenge};

\fill[ltGray!6] (0,0) rectangle (2.4,4.02);
\fill[ltGray!13] (2.4,0) rectangle (4.8,4.02);
\fill[ltAttack!10] (5.80,0) rectangle (10.60,4.02);
\fill[ltSafe] (11.60,0) rectangle (16.40,4.02);
\foreach \yy in {.65,1.30,1.95,2.60,3.25}{
  \draw[ltGray!30,line width=.35pt] (0,\yy) -- (2.4,\yy);
  \draw[ltGray!35,dashed,line width=.35pt] (2.4,\yy) -- (4.8,\yy);
  \draw[ltGray!30,line width=.35pt] (5.80,\yy) -- (8.20,\yy);
  \draw[ltGray!35,dashed,line width=.35pt] (8.20,\yy) -- (10.60,\yy);
  \draw[ltGray!30,line width=.35pt] (11.60,\yy) -- (14.00,\yy);
  \draw[ltGray!35,dashed,line width=.35pt] (14.00,\yy) -- (16.40,\yy);
}
\foreach \ll/\rr in {0/4.8,5.80/10.60,11.60/16.40}{
  \draw[ltGray!45,figthin,rounded corners=3pt] (\ll,0) rectangle (\rr,4.02);
}
\draw[ltGray!60,dashed,line width=.55pt] (2.4,0) -- (2.4,4.02);
\node[lab,text=black] at (1.2,3.77) {Clear};
\node[lab,text=black] at (3.6,3.77) {Encrypted};

\draw[->,ltGray,figarrow] (-1.40,.1) -- (-1.40,3.65);
\node[note,rotate=90,text=black] at (-1.68,1.9) {underlying level};
\draw[ltGray!70,densely dotted,line width=.7pt] (-0.28,1.53) -- (4.8,1.53);
\draw[ltGray!70,densely dotted,line width=.7pt] (-0.28,2.20) -- (4.8,2.20);
\node[lab,anchor=east,text=black,fill=white,inner sep=1pt] at (-0.08,1.53) {\(\ell\)};
\node[lab,anchor=east,text=black,fill=white,inner sep=1pt] at (-0.08,2.20) {\(\ell+\sqrt{\ell}\)};
\foreach \xx/\yy in {.35/.24,.95/.40,1.55/.23,2.05/.45,.85/.92,1.80/1.05}{
  \fill[ltGray!60] (\xx,\yy) circle (1pt);
  \draw[ltGray!60,fill=white] ({\xx+2.4-.03},{\yy-.03})
    rectangle ({\xx+2.4+.03},{\yy+.03});
}
\node[dot,fill=ltTask,label={[text=black]below left:\(x\)}] (x) at (0.6,1.53) {};
\node[dot,fill=ltTask,label=above right:\(y_1\)] (y) at (1.85,2.20) {};
\node[dot,fill=ltTask,label=right:\(y_2\)] (z) at (1.3,3.03) {};
\draw[flow,ltTask] (x) -- (y);
\draw[flow,ltTask] (x) to[out=90,in=230] (z);
\node[enc,draw=ltTask] (ex) at (3,1.53) {};
\node[enc,draw=ltTask] (ey) at (4.25,2.20) {};
\node[enc,draw=ltTask] (ez) at (3.7,3.03) {};
\draw[flow,ltTask] (ex) -- (ey);
\draw[flow,ltTask] (ex) to[out=90,in=230] (ez);
\node[lab,text=black,anchor=east] at (2.88,1.35) {\(\bar x\)};
\node[lab,text=black,anchor=south,inner sep=2pt] at (4.25,2.26) {\(\bar y_1\)};
\node[lab,text=black,anchor=west] at (3.79,3.03) {\(\bar y_2\)};

\node[note,anchor=north] at (2.4,-.22) {Each question at level \(\ell\) has many\\ answers,  all at level \(\geq\ell+\sqrt{\ell}\).};

\draw[ltGray!60,dashed,line width=.55pt] (8.20,0) -- (8.20,4.02);
\node[lab,text=black] at (7.00,3.77) {Clear};
\node[lab,text=black] at (9.40,3.77) {Encrypted};
\draw[ltAttack,figarrow] (5.80,2.35) -- (10.60,2.35);
\draw[ltDefense,figarrow] (11.60,2.35) -- (16.40,2.35);
\node[lab,anchor=east,text=black,fill=white,inner sep=1pt] at (5.72,2.35) {\(S\)};
\node[lab,anchor=east,text=black,fill=white,inner sep=1pt] at (11.52,2.35) {\(S\)};

\node[dot,fill=ltAttack] (u1) at (6.75,.25) {};
\node[enc,draw=ltAttack] (e1) at (8.85,.25) {};
\node[enc,draw=ltAttack] (v1) at (10.13,.96) {};
\node[dot,fill=ltAttack] (u2) at (6.75,.96) {};
\node[enc,draw=ltAttack] (e2) at (8.85,.96) {};
\node[enc,draw=ltAttack] (v2) at (10.13,1.74) {};
\node[dot,fill=ltAttack] (u3) at (6.75,1.74) {};
\node[enc,draw=ltAttack] (e3) at (8.85,1.74) {};
\node[enc,draw=ltAttack] (d) at (10.18,2.77) {};
\foreach \from/\to in {u1/e1,u2/e2,u3/e3}{
  \draw[flow,ltAttack] (\from) -- (\to)
    node[midway,lab,text=black,fill=ltAttack!10,inner sep=1.5pt] {Enc};
}
\draw[flow,ltAttack,name path=decOne] (v1) to[out=155,in=25]
  node[midway,lab,text=black,fill=ltAttack!10,inner sep=1.5pt] {Dec} (u2);
\draw[flow,ltAttack,name path=decTwo] (v2) to[out=155,in=25]
  node[midway,lab,text=black,fill=ltAttack!10,inner sep=1.5pt] {Dec} (u3);
\path[name path=evalTwo] (e2) -- (v2);
\path[name path=evalThree] (e3) -- (d);
\path[name intersections={of=decOne and evalTwo,by=crossOne}];
\path[name intersections={of=decTwo and evalThree,by=crossTwo}];
\fill[ltAttack!10] (crossOne) circle (2.7pt);
\fill[ltAttack!10] (crossTwo) circle (2.7pt);
\foreach \from/\to in {e1/v1,e2/v2,e3/d}{
  \draw[flow,ltAttack] (\from) -- (\to)
    node[midway,right=3pt,text=black] {\(f\)};
}
\foreach \yy/\index in {.25/1,.96/2,1.74/3}{
  \node[lab,text=black,anchor=east] at (6.59,\yy) {\(x_\index\)};
}
\draw[flow,ltAttack,dashed] (d) -- (8.62,3.40);
\node[circle,fill=ltAttack!18,draw=ltAttack,figbox,text=black,inner sep=0pt,minimum size=10pt,
      font=\normalfont\rmfamily\fignotefont]
  at (8.62,3.40) {\(\times\)};
\node[lab,text=black] at (10.18,2.56) {\(\bar x^\star\)};
\node[note,anchor=north] at (8.20,-.22) {The attacker iterates \(f\) under\\encryption until \(f\) fails on \(\bar x^\star\).};

\draw[ltGray!60,dashed,line width=.55pt] (14.00,0) -- (14.00,4.02);
\node[lab,text=black] at (12.80,3.77) {Clear};
\node[lab,text=black] at (15.20,3.77) {Encrypted};
\draw[ltDefense,line width=1.8pt] (12.60,.30) -- (12.60,2.35);
\node[dot,fill=ltDefense] at (12.60,.30) {};
\foreach \yy in {.85,1.35,1.85,2.35}{
  \node[circle,inner sep=0pt,minimum size=4.5pt,fill=white,draw=ltDefense,figbox] at (12.60,\yy) {};
}
\node[note,rotate=90,anchor=south,text=black] at (12.44,1.30) {run \(M\)};
\node[note,rotate=90,anchor=north,text=black] at (12.76,1.30) {checkpoints};
\node[enc,draw=ltAttack] (given) at (14.65,2.77) {};
\node[lab,text=black,anchor=east,inner sep=1pt] at (14.53,2.77) {\(\bar x^\star\)};
\node[circle,fill=ltAttack!18,draw=ltAttack,figbox,text=black,inner sep=0pt,minimum size=10pt,
      font=\normalfont\rmfamily\fignotefont] (fail) at (14.30,3.50) {\(\times\)};
\draw[flow,ltAttack,dashed] (given) -- (fail) node[midway,left=2pt,lab,text=black] {\(f\)};
\node[enc,draw=ltDefense] (hop1) at (15.08,2.98) {};
\node[enc,draw=ltDefense] (hop2) at (15.52,3.19) {};
\node[enc,draw=ltDefense] (answer) at (15.95,3.40) {};
\draw[flow,ltDefense] (given) -- (hop1);
\draw[flow,ltDefense] (hop1) -- (hop2) node[midway,sloped,above=1pt,note,text=black] {Update};
\draw[flow,ltDefense] (hop2) -- (answer);
\node[lab,text=black,anchor=west,inner sep=2pt] at (16.02,3.40) {\(\bar y^\star\)};
\node[note,anchor=north] at (14.00,-.22)
  {Store checkpoints up to \(S\). Inference\\extends \(\bar x^\star\) under encryption.};
\end{tikzpicture}
\endgroup
}
\caption{\textbf{Schematic of the construction of Theorem~\ref{thm:mitdetgentime_informal}.}
Questions and answers share the same space.
Horizontal lines separate difficulty levels.
Valid answers for questions on level \(\ell\) are at level at least \(\ell+\sqrt \ell\).
The attacker evaluates \(f\) on encrypted queries, decrypts, checks if the outputs are correct, and reencrypts them as new questions.
This iteration eventually yields an \(f\)'s mistake.
Active defense answers by extending the configuration in \(\bar x^\star\) through homomorphic computation for about \(\sqrt{S}\) more steps.
}
\label{fig:l-time-schematic}
\end{figure}
A task \(\tasktime\) will satisfy the following four properties.
\begin{enumerate}[topsep=0pt, itemsep=0pt, parsep=0pt]
\item \textbf{Correct Answers Are Higher-Level Inputs.}
For an input \(x\) of level \(\ell\), every valid output \(y\) is itself an
input and has level \(\geq \ell+\sqrt \ell\).
Thus, every correct answer advances an underlying sequential computation by
\(\sqrt \ell\) steps.

\item \textbf{Levels Certify Sequential Work.}
A valid level-\(\ell\) input can be produced in \(O(\ell)\) sequential time, but no circuit of size \(\ell^{\eta}\) produces a valid input at the terminal level \(\ell\) of the underlying computation, where \(\eta<1\) is the hardness gap of the NPL.
Thus, correctness at the terminal level certifies an expended computational effort (is a ``proof of work''),
but the certificate itself does not provide a shortcut for reaching
substantially later configurations.\footnote{Properties 1 and 2 together are the defining properties of a continuous verifiable delay function \citep{ephraim2020continuous}. See also verifiable delay functions \citep{boneh2018vdf}.}
\item \textbf{Level Hiding.}
The marginal on inputs is an equal mixture of \emph{clear} and \emph{opaque}
inputs.
The level of a clear input is efficiently computable.
For opaque inputs, the level is strongly computationally hidden.
In particular, opaque encodings of inputs at different levels are
computationally indistinguishable to a party that lacks the relevant secret
key.

\item \textbf{Asymmetric Verifiability.}
The attacker can generate a subset of opaque inputs for which it privately
knows a decryption key.  Consequently, it can apply the model to an opaque
input, decrypt the answer, and check whether the answer is valid.
\end{enumerate}

We explain why these properties yield a separation, and then give the cryptographic instantiation.
\paragraph{How to Attack Any Adversarial Defense.}
Let \(f\) be the model produced by a training algorithm of size \(S\).
Starting from the input \(x_1\) of a clear sample, the attacker can call \(f\) on itself multiple times, that is,
\begin{equation*}
x_i = f(f(\cdots f(x_1)\cdots)).
\end{equation*}
By property 1, \(O(\sqrt S)\) iterations suffice to reach level \(\Theta(S)\).
By property 2, a model produced by training of size \(S\) cannot continue supplying valid
configurations once this iteration passes the amount of sequential work
available during training.  Thus there is a first index \(i=O(\sqrt S)\) for
which \(f(x_i)\) is invalid.

The key challenge for the attacker is to compute an adversarial example \(x\), which is \emph{indistinguishable}, for the defender, from a random sample from \(\dist_{\inspace}\).
If \(x\) is a clear input then it is obviously distinguishable by inspecting its level.
Thus, to construct an indistinguishable \(x\), the attacker uses property 4 to choose an opaque input \(x\) for which it knows its level as well as the levels of \(f(x), f(f(x)), \dots\).
In contrast, by property 3, the defender (model \(f\)) cannot compute the level of \(x\), and so an AD does not exist.

\paragraph{How to Defend Using Inference Time.}
Training uses a circuit of size \(S\) to compute the underlying sequential trajectory and
stores a few certified configurations above level \(2\prm\), each of which is a valid answer to every input of level at most \(\prm\). Since the honest level has a
geometrically decreasing tail, it is at most \(\prm\) except with negligible probability, and this is sufficient for Correctness.

After the attacker chooses \(x\), the inference procedure continues from the configuration in \(x\) for \(O(\sqrt{S})\) steps.
An attacker of size \(O(S)\), even when combined with the information in
the trained model, cannot produce a valid input of level more than \(O(S)\) by
property 2.
This is the
source of the separation: an AD must commit to its model
before seeing the adversarial input, whereas an AAD may spend
\(O(\sqrt S)\) work extending precisely the configuration that the attacker
supplies.

\paragraph{How to Instantiate \(\tasktime\) Satisfying Properties 1--4.}
We use an average-case NPL (Definition~\ref{def:npl}), an IVC scheme
(Definition~\ref{def:ivc}), and IB-FHE (Definition~\ref{def:ibfhe}).
Their roles are distinct.
NPL supplies a computation with no substantially faster parallel shortcut.
IVC attaches a short, publicly verifiable proof to each configuration and permits the proof to be updated one step at a time.
IB-FHE hides the level while allowing both the learned model and the inference procedure to operate on encrypted inputs.

A first attempt, containing most of the ideas of the proof, at putting all the pieces together and defining the learning task \(\tasktime\) concretely is as follows.
We let the distribution \(\dist\) be composed of two equally probable parts
\(
\dist = \frac12 \dist_\mathsf{Clear} + \frac12 \dist_\mathsf{Enc}.
\)
For every \((x,y) \in \mathsf{supp}(\dist_\mathsf{Clear})\),
\[
x = c \| \ell \| \pi, \  y = c' \| (\ell+\sqrt \ell) \| \pi'.
\]
Let \(\mathsf{Samp}\) be the sampler and \(M\) the decision algorithm of an average-case NPL, and let \(\chi \gets \mathsf{Samp}(1^\prm, \prm \cdot S^{1/\eta})\) be a single instance, at a difficulty just beyond the combined budget of training and attack.
Pair \((x,y)\) satisfies: (i) \(c\) and \(c'\) are the configurations of \(M(\prm \cdot S^{1/\eta}, \cdot)\) on \(\chi\) after \(\ell\) and \(\ell+\sqrt \ell\) steps, (ii) \(\pi\) and \(\pi'\) are IVC proofs of these two statements, computed step by step with the IVC algorithm \(\mathsf{Update}\), which turns a proof for \(\ell\) steps into a proof for \(\ell+1\) steps at the cost of one step.
The level \(\ell\) is distributed as \(\mathsf{Geom}(\frac12)\).
Every \(y\) is correct on an \(x\) with an invalid proof, and otherwise \(y\) is incorrect on \(x\) if it contains an invalid proof or its level is smaller than \(\ell+\sqrt \ell\).
\(\dist_\mathsf{Enc}\) is defined as \(\dist_\mathsf{Clear}\) under FHE encryption.

\paragraph{How to Design an Active Adversarial Defense.}
To show that an AAD exists, we define \(\provtrain\) that computes a model \(f\) that answers correctly for levels up to \(S\).
Next, upon receiving an input \(x\), \(\provfix\) \emph{computes} a correct output ``from scratch'', that is, if \(x\) is unencrypted it parses \(x = c \ \| \ \ell \ \| \ \pi\) and runs \(M\) for \(\sqrt{\ell}\) steps with \(c\) as a starting point.
Additionally, it computes an IVC proof that its computation was correct.
If \(x\) is encrypted it runs the same procedure under the FHE.

The intuitive reason why \(\provfix\) is a valid AAD is that the combined running time of \(\ver\) and \(\provtrain\) is \(O(S) + S = O(S)\) and so, by the sequential hardness of the NPL corresponding to \(M\), they can compute the correct configurations of \(M\) only for \(O(S)\) steps, which implies that the highest level of \(x\) that \(\ver\) can create is \(O(S)\).
This implies that \(O(\sqrt{S})\) computation of \(\provfix\) is enough.

\paragraph{An Update to the Attack on Adversarial Defenses.}
To attack, \(\ver\) upon receiving \(f\) computes a valid input \(x_1\) on level \(1\) and iterates \(x_i = f(x_{i-1})\) until \(f\) no longer produces a valid output for \(x_i\).
Then she sends as adversarial input \(x = \Enc(x_i)\) with probability \(\frac12\) or a fresh sample from the marginal of \(\dist_\mathsf{Clear}\) otherwise.
Notice that this input is indistinguishable from a sample from \(\dist_{\inspace}\) by the properties of FHE.
Observe that such \(\ver\) makes \(O(\sqrt{S})\) calls to \(f\), because by the properties of the NPL, \(f\) can be correct only up to level \(O(S)\). Hence its size is \(\sqrt{S}\cdot S_E\cdot\poly(\prm)\).

Unfortunately, we cannot quite yet guarantee that \(f\) would make an error on \(x\).
It is possible that \(\prov\) could have provided \(f\) that is weaker on \(\dist_\mathsf{Clear}\) than it is on \(\dist_\mathsf{Enc}\)!
After all, this is ``how'' an AAD works, that is, \(\provfix\) ``hides'' from \(\ver\) a stronger model than \(f\), with which he answers.
Maybe, \(\prov\) did ``hide'', e.g., by obfuscating, a stronger model on the \(\dist_\mathsf{Enc}\) part.
To solve this issue we use a special type of FHE, called an identity-based FHE that will make \(\tasktime\) satisfy property 4.
Modifying \(\tasktime\) slightly will allow the parties partial decryption access and the evaluation of \(h(x,f(x))\).
With this final change we get a proof. See Section~\ref{sec:separation} for the full proof.
\end{proof}

\subsubsection{A Generative Task from Unforgeable Tokens Hidden by Encryption}
We now sketch the proof of Theorem~\ref{thm:mitdetgen_informal} for a task \(\taskcon\). It follows the pattern of \(\tasktime\), with signatures in place of computation steps: every AD with sample complexity \(K\) can be attacked, while an AAD that draws fresh samples at inference responds robustly.

The schematic of Figure~\ref{fig:l-time-schematic} applies to \(\taskcon\) as well, with signatures in place of computation steps, samples in place of time, and fresh signatures in place of the additional computation of the inference phase. 

\begin{proof}[Proof sketch of Theorem~\ref{thm:mitdetgen_informal}]
 The task \(\taskcon\) satisfies analogues of the four properties of \(\tasktime\), which we describe briefly.

\begin{enumerate}[topsep=0pt, itemsep=0pt, parsep=0pt]
\item \textbf{Level Amplification.} For an input \(x\) of level \(\ell\), every valid output \(y\) is itself an
input and has level \(\geq \ell+\sqrt \ell\).
\item \textbf{Levels Certify Samples.}
It now says that observing \(s\) inputs up to level \(\ell\) does not help to produce proofs above level \(\max\{s,\ell\}\).
\item \textbf{Level Hiding.} For opaque inputs, the level is hidden.
\item \textbf{Asymmetric Verifiability.} The attacker can create opaque inputs whose answers it can decrypt and check.
\end{enumerate}

As before, \(x_1\) will be easy to construct from the samples and \(x_i\) is the result of iteratively applying a model \(f\). The attacker again will be able to reach a level of order \(K\) on which \(f\) will fail.

An AAD will not fail by using fresh samples to modify \(f\) by making it work on inputs of higher levels.

\paragraph{How to Instantiate \(\taskcon\) Satisfying Properties 1--4.}
We use strongly unforgeable signatures (Definition~\ref{def:signatures}), a zk-SNARK
(Definition~\ref{def:zk-snark}), and IB-FHE (Definition~\ref{def:ibfhe}).
Signatures supply information that a party can only obtain from samples.
The zk-SNARK attaches a short, publicly verifiable proof of knowledge of many signatures to each input without revealing them.
IB-FHE hides the level (number of signatures) while allowing both the learned model and the inference procedure to operate on encrypted inputs.

As before, we let the distribution \(\dist\) be composed of two equally probable parts.
For every \((x,y) \in \mathsf{supp}(\dist_\mathsf{Clear})\),
\[
x = \sigma \| \ell \| \pi, \  y = \sigma \| (\ell+\sqrt \ell) \| \pi'.
\]
Let \((\mathsf{vk},\sk)\gets\Gen(1^\prm)\) be a key pair of a strongly unforgeable signature scheme.
Pair \((x,y)\) satisfies: (i) \(\sigma\) is a signature of \(0\) under \(\mathsf{vk}\), (ii) \(\pi\) and \(\pi'\) are zk-SNARK proofs of knowledge of \(\ell\) and \(\ell+\sqrt \ell\) distinct signatures of \(0\), each computed from fresh signatures.
The level \(\ell\) is distributed as \(\mathsf{Geom}(\frac12)\).
\(\dist_\mathsf{Enc}\) is defined as \(\dist_\mathsf{Clear}\) under the IB-FHE encryption.

\paragraph{How to Design an Active Adversarial Defense.}
We define \(\provtrain\) that computes a model \(f\) that answers correctly for levels up to \(K/8\).
Next, upon receiving an input \(x\), \(\provfix\) computes a correct output by first parsing \(x = \sigma \ \| \ \ell \ \| \ \pi\) and joining the signatures of \(O(\sqrt K)\) fresh samples with the \(K/4\) signatures kept from training.
It computes a zk-SNARK proof \(\pi'\) of knowledge of these \(K/4+\sqrt K/2\) signatures and returns \(\sigma\|(K/4+\sqrt K/2)\|\pi'\).
If \(x\) is encrypted it runs the same procedure under the FHE.

The intuitive reason why \(\provfix\) is a valid AAD is that the attacker sees at most \(K/4\) samples and the trained model reveals only zero-knowledge proofs up to level \(K/4\), and so, by Lemma~\ref{lem:pool}, they can produce valid proofs only up to level \(K/4\), which implies that the highest level of \(x\) that \(\ver\) can create is \(K/4\).
This implies that \(O(\sqrt K)\) fresh samples of \(\provfix\) are enough.

\paragraph{How to Attack Any Adversarial Defense.}
The attack is to send an adversarial input \(x = \Enc(x_i)\) for some \(i\leq O(\sqrt K)\) such that \(f\) fails on encryptions of \(x_i\) with probability \(\frac12\) or a fresh sample from the marginal of \(\dist_\mathsf{Clear}\) otherwise.
As before, such an input is indistinguishable from a natural sample by the properties of FHE.

By property 2, \(f\) can be correct only up to level \(O(K)\), while by property 1, \(O(\sqrt K)\) valid answers in a row starting from \(x_1\) reach a higher level. Hence such an \(i\) exists, the attacker finds it by property 4, and a defender producing such a model fails.
See Section~\ref{sec:separation-samples} for the full proof.
\end{proof}

\subsection{No Dichotomy for Generative Tasks}\label{sec:overview-neither}
\begin{proof}[Proof sketch of Corollary~\ref{cor:neithertime_informal}]
Item (ii) is item (ii) of Theorem~\ref{thm:mitdetgentime_informal}. For item (i), take the AAD of Theorem~\ref{thm:mitdetgentime_informal}. Its inference phase does not use \(\priv\) from the training phase, so it is a defender on its own, of size \(\sqrt{S}\cdot\poly(\prm)\). An attacker of size \(O(S)\) in a TA sees no published model, so it is an attacker of the AAD that ignores \(f\), and by item (i) of the theorem the inference phase answers its questions mostly correctly. Hence no attacker of size \(O(S)\) fools every defender of size \(\sqrt{S}\cdot\poly(\prm)\), and none is a TA.
\end{proof}

The proof crucially uses that the task is generative. An AD could run the inference step inside its distinguishing rule \(R\), but a disagreement between that answer and \(f(x)\) does not show that \(f\) errs. For classification tasks a disagreement does show an error. An AD can publish \(f\) and, according to a private fair coin drawn in training, flag an input if the AAD flags it, or flag it if the answer of the AAD disagrees with \(f\), which turns an AAD into an AD of size about \(\ttrain+\tfix\). If the AAD distinguishes an attack, the AD does so with half the gap. If the AAD answers it mostly correctly, the disagreement check flags it at a rate close to the error of \(f\) on it and flags natural inputs rarely, so \(f\) is robust or the AD distinguishes. Hence no separation as in Theorem~\ref{thm:mitdetgentime_informal} exists for classification tasks.

Similarly the proof of Corollary~\ref{cor:neithertime_informal} does not work for classification tasks. This is because the AAD would turn into an AD of size \(O(S)\) as above and contradict item (ii). Whether the dichotomy is tight for classification tasks therefore remains open.

%% file: sections/preliminaries.tex
\subsection{Notation}\label{sec:notation}
We write \(\Delta(\cdot)\) for the set of probability measures on a set, with respect to an implicitly fixed \(\sigma\)-algebra.
We write \(x \gets \dist\) for a draw from a distribution \(\dist\), and for a uniform draw when \(\dist\) is a finite set, and \(y \gets C(x)\) for the output of a randomized circuit or algorithm \(C\) on input \(x\). We write \(|\cdot|\) for the length of a string, and \(\ell^+ := \ell + \sqrt{\ell}\) for \(\ell \in \N\). Whenever a number is part of an input to a party or an algorithm, it is written in binary, except for the unary strings \(1^\prm\) and \(1^\lambda\).

A function \(\mu: \N \to \R\) is called negligible if for all \(c \in \R_{>0}\) there exists \(\prm_0 \in \N\) such that for all integers \(\prm \geq \prm_0\), we have \(|\mu(\prm)|<1/\prm^c\).

We write \(\negl(\prm)\) for an unspecified negligible function and \(\poly(\prm)\) for an unspecified polynomial. Parameters of our definitions, such as error bounds, size bounds and sample complexity bounds, are functions of \(\prm\). Cryptographic primitives are instantiated with security parameter \(\lambda = \prm\) unless stated otherwise.

Our schemes are not required to succeed at every input length, and we measure how often they do.
A boolean sequence \(a : \N \to \{0,1\}\) is true with frequency \(\varphi \in [0,1]\) if
\begin{align*}
    \limsup_{\prm \to \infty} \frac{\sum_{i \in [\prm]} a_i}{\prm} \geq \varphi.
\end{align*}
We write \(\ver \in_{\varphi} \mathsf{S}(\cdot)\) to say that \(\ver\) implements the scheme \(\mathsf{S}\) with frequency \(\varphi\), and \(\ver \notin_{\varphi} \mathsf{S}(\cdot)\) when this does not hold, and drop the subscript when the frequency is clear from the context.

\subsection{Learning Tasks}\label{sec:learning-tasks}
We denote the input space by \(\inspace\) and the output space by \(\outspace\).
An error oracle is a function \(h : \inspace \times \outspace \to \{0,1\}\),
i.e., it is a function that takes a pair \((x,y)\) of input \(x\) and output \(y\), and outputs \(0\) if \(y\) is a valid answer for \(x\), and \(1\) otherwise.
Note that this captures both the classification and generation settings. If for every \(x\) there is exactly one \(y\) such that \(h(x,y) = 0\), then \(h\) corresponds to a classification task, but if there are many such \(y\)'s we deal with generation.
In the classification case we also write \(h(x)\) for the unique valid answer, slightly overloading the notation. Thus, \(h(x,y) = \mathbf{1}_{y \neq h(x)}\).

For a distribution \(\dist \in \Delta(\inspace \times \outspace)\) we denote its marginal distribution over \(\inspace\) by \(\dist_{\inspace}\).
For a classification task we assume that every sample is of the form \((x,h(x))\), so \(\dist\) is determined by \(\dist_{\inspace}\) and \(h\). In this case we identify \(\dist\) with \(\dist_{\inspace}\) and write \(x \gets \dist\).

\begin{definition}[Learning Task]\label{def:task}
    A learning task \(\task\) is a family \(\{\task_\prm\}_{\prm \in \N}\), where for every \(\prm\), \(\task_\prm\) is an element of \(\Delta\big(\Delta(\inspace \times \outspace) \times \{0,1\}^{\inspace \times \outspace}\big)\), i.e., it is a distribution over pairs \((\dist,h)\), where \(\dist \in \Delta(\inspace \times \outspace)\) and \(h : \inspace \times \outspace \to \{0,1\}\).
\end{definition}
We note that Definition~\ref{def:task}, which is similar to the one used in \citet{gluch2026good}, is nonstandard. Tasks are indexed by the size parameter \(\prm\) so that the computational resources of the parties can be measured, and the outer distribution over \((\dist,h)\) pairs represents the prior knowledge of the learner. Since a non-uniform circuit may hard-code advice for each \(\prm\) \citep{karp1980some}, a task fixed to a single \((\dist,h)\) would be settled by a circuit that simply stores the answers. The outer distribution is therefore what leaves something to be learned, and what lets a task be built from a cryptographic primitive and its security be stated as a bound on every party of a given size.

A model is a potentially randomized function \(f : \inspace \to \outspace\). A classifier is a deterministic model. A distinguishing rule is a potentially randomized function \(R: \inspace \to \{0,1\}\). We call its output \(r = R(x)\) a flag. This rule will be used to distinguish natural and adversarially generated inputs.

For a distribution \(\dist \in \Delta(\inspace \times \outspace)\) and an error oracle \(h\) we define the error of a model \(f\) as
\begin{align*}
    \errf := \Pr_{x \gets \dist_{\inspace}} \big[ h(x,f(x)) = 1 \big],
\end{align*}
where the probability also includes the potential randomness of \(f\). We will omit the indices \(\dist\) and \(h\) if they are clear from the context.
For \(q \in \N\), a batch \(\xs \in \inspace^q\), answers \(\ys \in \outspace^q\) and flags \(\rs \in \{0,1\}^q\) we also define the empirical error and the flag rate as
\[\err_{\xs,h}(\ys) := \frac1q \sum_{i \in [q]} h(x_i,y_i)\quad \text{  and  }\quad\fl(\rs) := \frac1q |\{ i \in [q] : r_i = 1 \}|.\]

\subsection{Circuits, Sample Access, and Representations}\label{sec:circuits}
Throughout, efficient computation is modeled by non-uniform families of Boolean circuits \(C=\{C_\prm\}_\prm\). The size of a circuit is its number of gates including input gates. We write \(|C|\) for the size of a circuit \(C\). A family has size \(S: \N \to \N\) if \(|C_\prm| \leq S(\prm)\) for every \(\prm\).
When we describe a party as an algorithm, its size is the size of a circuit implementing the algorithm. Up to a polylogarithmic factor, the size of a circuit implementing an algorithm is its running time \citep{arora2009computational}, and we write \(\tilde O(\cdot)\) to hide polylogarithmic factors. Polynomial bounds are unaffected by this factor, so for them we use running time and size interchangeably.

Our circuits have access to an example oracle \(\ex\) \citep{valiant1984} that samples \((x,y) \gets \dist\) and returns \((x,y)\). For a classification task, where we identify \(\dist\) with \(\dist_{\inspace}\), we write \(\exh\) for this oracle: it returns \((x, h(x))\) with \(x \gets \dist\).

Besides its size, a circuit family has sample and randomness complexity. A family has sample complexity \(K: \N \to \N\) and randomness complexity \(J: \N \to \N\) if \(C_\prm\) makes at most \(K(\prm)\) queries to \(\ex\) and reads at most \(J(\prm)\) random bits, for every \(\prm\).
We sometimes explicitly write \(C_\prm(\cdot\,, Z)\), where, given \((\dist,h)\), \(Z\gets \{0,1\}^{J(\prm)} \times \ex^{K(\prm)}\).

Models and distinguishing rules are exchanged between parties as strings.
The functions \(f\) and \(R\) that a defender produces belong to classes \(\fclass_\prm\) and \(\mathcal{R}_\prm\), whose members are described by strings \(z_f\) and \(z_R\) of length \(m_{\fclass}(\prm)\) and \(m_{\mathcal{R}}(\prm)\).

A family of evaluation circuits \(E=\{E_\prm\}_\prm\) of size \(S_E\) for \(\fclass\) satisfies \(E_\prm(z_f,x)=f(x)\), so that a party holding \(z_f\) can compute \(f\). For a randomized \(f\), the circuit \(E_\prm\) also reads the random bits of \(f\). The cost of evaluating \(f\) is thus \(S_E\), while \(m_{\fclass}\) is only the cost of storing or transmitting \(z_f\). Evaluation circuits for the class \(\mathcal{R}\) of distinguishing rules are defined in the same way.

\subsection{Learnability}\label{sec:learnability}
Recall the definition of PAC learnability \citep{valiant1984, kearns1994introduction}. Taking the computational aspect of learning algorithms into account, \citet{gluch2026good} defined computationally bounded learnability.
\begin{definition}[Computationally Bounded Learnability]\label{def:learnability}
    For error and confidence parameters \(\varepsilon, \delta :\N \to (0,1)\), a size bound \(S_L:\N \to \N\) and a class of models \(\fclass=\{\fclass_\prm\}_\prm\) with description length \(m_{\fclass}\), we say that a learning task \(\task\) is \((\varepsilon, \delta, S_L, \fclass)\)-learnable, if there exists a circuit family \(\{L_\prm\}_\prm\) of size \(S_L\), which outputs descriptions \(z_f\) of models \(f \in \fclass_\prm\), such that for every \(\prm\) large enough
    \begin{align*}
        \Pr\left[\errf < \varepsilon(\prm) \GivenExperiment \StateExperiment{(\dist, h) \gets \task_\prm ,\; z_f \gets L_\prm^{\ex}}\right] \geq 1-\delta(\prm).
    \end{align*}
\end{definition}

\subsection{Interaction}\label{sec:interaction}
All attacks and defenses in this paper are with respect to some learning task \(\task\). Before the interaction between a defender and an attacker starts, \((\dist,h)\) is sampled from \(\task_\prm\). The task \(\task\) is known to all the parties, but the particular \((\dist,h)\) is unknown, i.e., it can be accessed only through the example oracle.

The notions studied in this paper are two- and three-round interactive protocols \citep{ip_goldwasser} between Alice and Bob. Alice is the attacker \(\ver=\{\ver_\prm\}_\prm\) and Bob is the defender \(\prov=\{\prov_\prm\}_\prm\). Both are circuit families with oracle access to \(\ex\), of sizes \(S_\ver\) and \(S_\prov\) and sample complexities \(K_\ver\) and \(K_\prov\). Alice sends a batch \(\xs \in \inspace^q\), where the batch size \(q\) is a polynomial known to all parties. Bob answers every entry \(x_i\) with some \(y_i \in \outspace\) and flags it with some \(r_i \in \{0,1\}\), either directly or through a published model and distinguishing rule. A defender may act in two phases \(\prov = (\provtrain, \provfix)\). The training phase \(\provtrain\) runs before Alice's message and the inference phase \(\provfix\) runs after it. Their sizes are \(\ttrain\) and \(\tfix\), and their sample complexities are \(\ktrain\) and \(\kfix\). We allow \(\provtrain\) to produce, during learning, a private string \(\priv\) that is passed to \(\provfix\) and is not sent to Alice. In Definitions~\ref{def:detectiondefense} and~\ref{def:mitigationdefense} the defender also publishes a distinguishing rule, and \(\priv\) instantiates it: all randomness of the rule is in \(\priv\), so \(R(\priv, \cdot)\) is a deterministic function. Evaluating the model and the rule, with the evaluation circuits of Section~\ref{sec:circuits}, is part of the defender, so its size bounds these evaluations.

\subsection{Transferable Attacks and (Active) Adversarial Defenses}\label{sec:definitions}

The definitions of this subsection apply to classification tasks and to generative tasks, where an input may have many valid answers. A defender may act in two phases, answering once more at inference time after the attacker has spoken, and the attacker sends a batch \(\xs \in \inspace^q\).
A defender answers every entry of the batch and, separately, flags it, so its output is a pair \((\ys, \rs) \in \outspace^q \times \{0,1\}^q\), scored by the empirical error and the flag rate.

Transferable Attacks are families of circuits that generate batches \(\xs \in \inspace^q\) that all computationally bounded defenders fail to answer correctly while also being unable to distinguish them from natural batches.
\begin{definition}[Transferable Attack]
\label{def:transferableattack}
Let \(\task\) be a learning task. For a batch size \(q : \N \to \N\), size functions \(S_\ver, S_\prov: \N \to \N\), transferability and indistinguishability bounds \(c_{\TA}, s_{\TA}: \N \to (0,1)\) and a frequency \(\varphi \in [0,1]\), we say that a family of circuits \(\ver_{\TA} = \{\ver_{\prm,\TA}\}_\prm\) of size \(S_\ver\) implements a \emph{Transferable Attack} for \(\task\) with frequency \(\varphi\), denoted by
\[
\ver_{\TA} \in_{\varphi} \TA(\task, q, S_\ver, S_\prov, c_{\TA}, s_{\TA}),
\]
if, with frequency \(\varphi\) over \(\prm\), an interactive protocol in which \((\dist,h) \gets \task_\prm\), \(\ver^{\ex}_{\prm,\TA}\) computes a batch \(\xs \in \inspace^q\) and an arbitrary \(\prov^{\ex}_\prm\) based on \(\xs\) outputs \((\ys, \rs) \in \outspace^q \times \{0,1\}^q\) satisfies:
\begin{itemize}[topsep=0pt, itemsep=0pt, parsep=0pt]
    \item \textbf{Transferability} (fast defenders answer the batch mostly incorrectly). For every circuit \(\prov_\prm\) of size \(S_\prov(\prm)\)
    \[
    \mathbb{E}\left[\err_{\xs,h}(\ys) \GivenExperiment \StateExperiment{(\dist,h) \gets \task_\prm ,\; \xs \gets \ver^{\ex}_{\prm,\TA} ,\; (\ys,\rs) \gets \prov^{\ex}_\prm(\xs)}\right] \geq c_{\TA}(\prm).
    \]
    \item \textbf{Indistinguishability} (fast defenders flag the batch about as often as a natural batch). For every circuit \(\prov_\prm\) of size \(S_\prov(\prm)\)
    \[
    \left|\,\mathbb{E}\left[\fl(\rs) \GivenExperiment \StateExperiment{(\dist,h) \gets \task_\prm \\ \xs \gets \ver^{\ex}_{\prm,\TA} \\ (\ys,\rs) \gets \prov^{\ex}_\prm(\xs)}\right]
        - \mathbb{E}\left[\fl(\rs) \GivenExperiment \StateExperiment{(\dist,h) \gets \task_\prm \\ \xs \gets (\dist_{\inspace})^{q} \\ (\ys,\rs) \gets \prov^{\ex}_\prm(\xs)}\right]\right| < s_{\TA}(\prm).
\]
\end{itemize}
\end{definition}

Adversarial Defenses are counterparts of Transferable Attacks. An Adversarial Defense fixes its model before the attacker moves. A defender that may answer again at inference time is called active.

\begin{remark}
An attack for a fixed batch size \(q\) enables attacks of the same strength for all \(q'\leq q\).
Indeed, given an attack successful for $q$, one can construct an attack for any $q' < q$ by choosing a random subset of size $q'$ from the batch of $q$ questions produced by the initial attack.
This new attack inherits at least as good $c_{\TA}, s_{\TA}$ as the initial one. 
\end{remark}


\begin{definition}[Adversarial Defense]
\label{def:detectiondefense}
Let \(\task = \{\task_\prm\}_\prm\) be a learning task. For a batch size \(q: \N \to \N\), size functions \(S_\ver, S_\prov : \N \to \N\), an error parameter \(\varepsilon: \N \to (0,1)\), a confidence parameter \(\delta: \N \to (0,1)\), distinguishability and robustness parameters \(c_{\AD}, s_{\AD}: \N \to (0,1)\) and a frequency \(\varphi \in [0,1]\), we say that a circuit family \(\prov = \{\prov_\prm\}_\prm\) of size \(S_\prov\) implements an \emph{Adversarial Defense} for \(\task\) with frequency \(\varphi\), denoted by
\[
\prov \in_{\varphi} \AD(\task, q, S_\ver, S_\prov, \varepsilon, \delta, c_{\AD}, s_{\AD}),
\]
if, with frequency \(\varphi\) over \(\prm\), an interactive protocol in which \((\dist,h) \gets \task_\prm\), \(\prov^{\ex}\) computes descriptions \(z_f\) and \(z_R\) of a model \(f \in \fclass_\prm\) and of a distinguishing rule \(R \in \mathcal{R}_\prm\), together with a private string \(\priv\) that instantiates \(R\), sends \(z_f\) and \(z_R\) to \(\ver_\prm^{\ex}\), which replies with a batch \(\xs \in \inspace^q\), and the defender outputs the answers \(\ys := f(\xs)\) and the flags \(\rs := R(\priv, \xs)\), both computed entry by entry, satisfies the conditions below.
\begin{itemize}[topsep=0pt, itemsep=0pt, parsep=0pt]
    \item \textbf{Correctness} (\(f\) has low error).
    \[
    \Pr\left[\errf < \varepsilon(\prm) \GivenExperiment \StateExperiment{(\dist,h) \gets \task_\prm ,\; z_f \gets \prov^{\ex}}\right] \geq 1 - \delta(\prm).
    \]
\end{itemize}
Moreover, for every circuit \(\ver_\prm\) of size \(S_\ver(\prm)\) at least one of the two conditions holds.
\begin{itemize}[topsep=0pt, itemsep=0pt, parsep=0pt]
    \item \textbf{Robustness} (quickly generated adversarial batches are mostly answered correctly).
    \[
    \mathbb{E}\left[\err_{\xs,h}(f(\xs)) \GivenExperiment \StateExperiment{(\dist,h) \gets \task_\prm ,\; (z_f,z_R,\priv) \gets \prov^{\ex} ,\; \xs \gets \ver_\prm^{\ex}(z_f,z_R)}\right] < s_{\AD}(\prm).
    \]
    \item \textbf{Distinguishability} (the published rule \(R\), instantiated by the private string \(\priv\), flags quickly generated adversarial batches at a different rate than natural batches).
    \begin{align*}
        &\mathbb{E}\left[\left|\mathbb{E}\left[\fl(\rs) \GivenExperiment \StateExperiment{\xs \gets \ver_\prm^{\ex}(z_f,z_R) ,\; \rs \gets R(\priv,\xs)}\right]-\right.\right.\\
        &\flushrow{\left.\left.\mathbb{E}\left[\fl(\rs) \GivenExperiment \StateExperiment{\xs \gets (\dist_{\inspace})^{q} ,\; \rs \gets R(\priv,\xs)}\right]\right| \GivenExperiment \StateExperiment{(\dist,h) \gets \task_\prm ,\; (z_f,z_R,\priv)\gets \prov^{\ex}}\right] \geq c_{\AD}(\prm).}
    \end{align*}
\end{itemize}
\end{definition}

\begin{remark}
Our definition of Adversarial Defense is related to the guarantee for arbitrary test-time examples of \citet{arbitraryexamples}, where abstaining plays the role of flagging, but the two notions differ in three respects. First, their learner is transductive and sees the unlabeled test batch before it answers, whereas ours commits \(f\) and \(R\) before the batch is chosen. Second, their guarantee is conjunctive, low error on the points it labels together with a low abstention rate on natural inputs, whereas ours asks, for each attacker, for Robustness or Distinguishability and does not bound the flag rate on natural inputs. Third, their adversary is computationally unbounded and their classes have bounded VC dimension, whereas we bound all parties by circuit size.
\end{remark}

\begin{definition}[Active Adversarial Defense]
\label{def:mitigationdefense}
Let \(\task = \{\task_\prm\}_\prm\) be a learning task. For a batch size \(q: \N \to \N\), size functions \(S_\ver, \ttrain, \tfix : \N \to \N\), an error parameter \(\varepsilon: \N \to (0,1)\), a confidence parameter \(\delta: \N \to (0,1)\), distinguishability and robustness parameters \(c_{\AAD}, s_{\AAD}: \N \to (0,1)\) and a frequency \(\varphi \in [0,1]\), we say that a pair of circuit families \((\provtrain,\provfix)\) of sizes \(\ttrain\) and \(\tfix\) implements an \emph{Active Adversarial Defense} for \(\task\) with frequency \(\varphi\), denoted by
\[
(\provtrain,\provfix) \in_{\varphi} \AAD(\task, q, S_\ver, \ttrain, \tfix, \varepsilon, \delta, c_{\AAD}, s_{\AAD}),
\]
if, with frequency \(\varphi\) over \(\prm\), an interactive protocol in which \((\dist,h) \gets \task_\prm\), \(\provtrain^{\ex}\) computes descriptions \(z_f\) and \(z_R\) of a model \(f \in \fclass_\prm\) and of a distinguishing rule \(R \in \mathcal{R}_\prm\), together with a private string \(\priv\) that instantiates \(R\) and is passed to the inference phase, sends \(z_f\) and \(z_R\) to \(\ver_\prm^{\ex}\), which replies with a batch \(\xs \in \inspace^q\), and \(\provfix^{\ex}(\priv, \xs)\) outputs answers \(\ys \in \outspace^q\) together with the flags \(\rs := R(\priv, \xs)\), computed entry by entry, satisfies the conditions below.
\begin{itemize}[topsep=0pt, itemsep=0pt, parsep=0pt]
    \item \textbf{Correctness} (the published model \(f\) has low error).
    \[
    \Pr\left[\errf < \varepsilon(\prm) \GivenExperiment \StateExperiment{(\dist,h) \gets \task_\prm ,\; z_f \gets \provtrain^{\ex}}\right] \geq 1 - \delta(\prm).
    \]
    \item \textbf{Completeness} (the inference phase has low error on natural batches).
    \[
    \Pr\left[
        \err_{\xs,h}(\ys) < \varepsilon(\prm)
        \GivenExperiment
        \StateExperiment{
            (\dist,h) \gets \task_\prm \\
            (z_f,z_R,\priv) \gets \provtrain^{\ex} \\
            \xs \gets (\dist_{\inspace})^q \\
            (\ys,\rs) \gets \provfix^{\ex}(\priv,\xs)
        }
    \right] \geq 1-\delta(\prm).
    \]
\end{itemize}
Moreover, for every circuit \(\ver_\prm\) of size \(S_\ver(\prm)\) at least one of the two conditions holds.
\begin{itemize}[topsep=0pt, itemsep=0pt, parsep=0pt]
    \item \textbf{Robustness} (inference time lets the defender answer quickly generated adversarial batches mostly correctly).
    \[
    \mathbb{E}\left[\err_{\xs,h}(\ys) \GivenExperiment \StateExperiment{(\dist,h) \gets \task_\prm \\ (z_f,z_R,\priv) \gets \provtrain^{\ex} \\ \xs \gets \ver_\prm^{\ex}(z_f,z_R) \\ (\ys,\rs) \gets \provfix^{\ex}(\priv,\xs)}\right] < s_{\AAD}(\prm).
    \]
    \item \textbf{Distinguishability} (the published rule \(R\), instantiated by the private string \(\priv\), flags quickly generated adversarial batches at a different rate than natural batches).
    \begin{align*}
        &\mathbb{E}\left[\left|\mathbb{E}\left[\fl(\rs) \GivenExperiment \StateExperiment{\xs \gets \ver_\prm^{\ex}(z_f,z_R) ,\; \rs \gets R(\priv,\xs)}\right]-\right.\right.\\
        &\flushrow{\left.\left.\mathbb{E}\left[\fl(\rs) \GivenExperiment \StateExperiment{\xs \gets (\dist_{\inspace})^{q} ,\; \rs \gets R(\priv,\xs)}\right]\right| \GivenExperiment \StateExperiment{(\dist,h) \gets \task_\prm ,\; (z_f,z_R,\priv)\gets \provtrain^{\ex}}\right] \geq c_{\AAD}(\prm).}
    \end{align*}
\end{itemize}
The two defenses differ only in the answers. Both publish \(z_f\) and \(z_R\) and flag the batch with \(R(\priv, \xs)\), and an Adversarial Defense \(\prov\) is the Active Adversarial Defense whose training phase is \(\prov\) and whose inference phase answers with \(\ys := f(\xs)\), where for batch size \(1\) Completeness holds with confidence \(1-\varepsilon-\delta\) in place of \(1-\delta\).
\end{definition}
\begin{remark}
We also make some comments on the batch size $q$.
Our definitions are general in that they allow arbitrary batch sizes $q\geq 1$.
All our results will be stated for $q = 1$ for simplicity.
However, most of the main theorems extend to all $q \geq 1$, namely our constructions of TAs (Section~\ref{sec:construction_PKE}) and the separation results for generative tasks (Theorems~\ref{thm:mitdetgen} and~\ref{thm:mitdetgentime}) extend to all $q \geq 1$.
We note that in full generality it is necessary to consider definitions that allow larger batch sizes in the sense that the existence of a primitive might depend on the value of $q$.
For example, \cite{arbitraryexamples} shows that larger batches are \emph{necessary} to design defenses against (arbitrary) test-time adversarial examples.\footnote{This is proved for a slightly different definition of an adversarial defense as mentioned.}
\end{remark}

When a size bound in Definitions~\ref{def:transferableattack}--\ref{def:mitigationdefense} is written \(\poly(\prm)\), it ranges over every polynomial if it bounds the quantified party, that is, the defenders in Definition~\ref{def:transferableattack} and the attackers in Definitions~\ref{def:detectiondefense} and~\ref{def:mitigationdefense}, and it stands for some polynomial if it bounds the party that is claimed to exist. The negligible functions may then depend on the universally quantified polynomial.

%% file: sections/classification.tex
This section studies binary classification tasks with \(\inspace = \{0,1\}^\prm\) and \(\outspace= \{0,1\}\). As in Section~\ref{sec:learning-tasks}, we identify \(\dist\) with \(\dist_{\inspace} \in \Delta(\{0,1\}^\prm)\) and write \(h(x)\) for the unique valid answer for \(x\), so that \(h(x,y) = \mathbf{1}_{y \neq h(x)}\) and the example oracle \(\exh\) returns \((x, h(x))\) with \(x \gets \dist\). Each party is bounded only by its size. For simplicity we consider \(q=1\) here. With \(q=1\) a batch is a single input \(x\), the defender outputs \((y,r) \in \{0,1\}^2\), the empirical error is \(\mathbf{1}_{y \neq h(x)}\) and the flag rate is \(r\), so the expectations in Definitions~\ref{def:transferableattack} and~\ref{def:detectiondefense} are the probabilities that \(y \neq h(x)\) and that \(r = 1\).

\subsection{The Dichotomy Lemma and Its Proof}\label{sec:dichotomy}
The following lemma loosely says that every learnable task either has a Transferable Attack or an Adversarial Defense.
\begin{lemma}[Dichotomy]\label{lem:dichotomy}
Fix a size function \(S: \N \to \N\) and constants \(0<c,s,\alpha, \varepsilon,\delta,\kappa<1 \) such that\footnote{This condition makes sure that the robustness parameter of the adversarial defense below is smaller than \(1\), so that the defense it produces is not vacuous.}
\begin{align*}
    \kappa((2-\alpha)(1-c)-s\alpha)>1+\varepsilon +\delta.
\end{align*}
Assume that a learning task \(\task\) is \((\varepsilon, \delta, S_L=\sqrt{S}/(4 \log S), \fclass)\)-learnable, where \(\fclass\) is a class of classifiers. Also assume existence of an evaluation circuit family \(E=\{E_\prm\}_\prm\) of size \(S_E\) for \(\fclass\). Finally, assume that for all sufficiently large \(\prm\) we have \(m_{\fclass}(\prm) \leq 3S_L(\prm)\) and \(\prm + 1 \leq 4S_L(\prm)\).
Then, at least one of the following exists:
\[\ver_{\TA} \in_{1/2} \TA\left(\task, 1, O(S), 4S_L, c, s\right),\]
or
\[\prov_{\AD} \in_{1/2} \AD\left(\task,1,3S_L,O(S+S_E),
\varepsilon,\delta, (s\alpha \kappa)/2, \varepsilon+\delta+2
-\kappa\left((2-\alpha)(1-c)-s\alpha \right)\right).\]
\end{lemma}
\begin{proof}
\textbf{Setup.} Let \(\theta:=(1-\kappa)/(1+\kappa)\), so that \(0<\theta<1\) and \(\kappa=(1-\theta)/(1+\theta)\). Let \(\{L_\prm\}_\prm\) be a circuit family that can learn \(\task\). To simplify notation, we omit subscript \(\prm\). Let \(Z_L,Z_\ver,Z_\prov\) be the strings of random bits and labeled samples consumed by \(L,\ver\) and \(\prov\), as in Section~\ref{sec:circuits}: \(Z_L \gets \{0,1\}^{J_L} \times \exh^{K_L}\), where \(J_L\) and \(K_L\) are the randomness and sample complexity of \(L\), and likewise for \(\ver\) and \(\prov\), each drawn independently. The same applies to every other circuit below, such as \(\ver_i\), \(\ver_N\) and \(\prov_N\). An experiment or a subscript lists the strings it averages over.
Throughout the proof, \((\dist,h) \gets \task_\prm\) is drawn once before every other draw. Following Section~\ref{sec:interaction}, every probability and expectation below is taken over this draw as well, and we omit it from the experiments to lighten the notation.

Now let \(\mathfrak{A}\) be the set of circuits \(\ver\) of size \(4S_L\) that output \(x \in \{0,1\}^\prm\). Also, let \(\mathfrak{B}\) be the set of pairs \((\prov, \sigma)\), where \(\sigma \in \{-1,1\}\) and \(\prov\) is a circuit of size \(4S_L\).

For any \(\ver \in \mathfrak{A}\) and \((\prov, \sigma) \in \mathfrak{B}\) we use the following shortcuts
\begin{align*}
    z_{f_{Z_L}}:= L(Z_L), \quad x_{Z_\ver}:=\ver(Z_\ver),\quad(y_{Z_\prov}(x), r_{Z_\prov}(x)):= \prov(x, Z_\prov)
\end{align*}
and define a matrix value
\begin{multline*}
    M_{\ver, (\prov,\sigma)}:= \frac{1+\theta}{1-c}\Pr\left[h(x_{Z_\ver})=y_{Z_\prov}(x_{Z_\ver}) \GivenExperiment \StateExperiment{Z_\ver, Z_\prov}\right]\\
    +\frac{1+\theta}{s}\big(\mathbb{E}\left[\sigma r_{Z_\prov}(x_{Z_\ver}) \GivenExperiment \StateExperiment{Z_\ver, Z_\prov}\right]-\mathbb{E}\left[\sigma r_{Z_\prov}(x) \GivenExperiment \StateExperiment{x\gets \dist ,\; Z_\prov}\right]\big).
\end{multline*}

\paragraph{Game Theory Argument.} Let \(\gamma:= \theta/(\max_{\ver,(\prov,\sigma)}M_{\ver,(\prov,\sigma)} - \min_{\ver,(\prov,\sigma)}M_{\ver,(\prov,\sigma)})\). By the results of \citet{v1928theorie} and \citet{lipton1994simple} (see Appendix~\ref{app:games}, which also defines the value \(\operatorname{val}(M)\) of the game) there exist \(k=O(S_L\log(S_L))\) circuits \(\ver_1, \dots, \ver_k\) in \(\mathfrak{A}\) such that the uniform distribution over them yields an approximately optimal strategy. This means that for all \((\prov,\sigma)\in \mathfrak{B}\)
\begin{align}\label{eq:nash-approx-1}
    \sum_{i=1}^k \frac{1}{k} M_{\ver_i,(\prov,\sigma)} \leq \operatorname{val}(M)  + \theta.
\end{align}
Let \(\ver_N\) be a circuit corresponding to the uniform strategy. Then its size is
\begin{align*}
    4kS_LO(\log k)=O(S_L^2\log^2S_L)=O(S)
\end{align*}
and
\begin{align*}
    \sum_{i=1}^k \frac{1}{k} M_{\ver_i,(\prov,\sigma)} &=\frac{1+\theta}{1-c}\sum_{i=1}^k \frac{1}{k}\Pr\left[h(x_{Z_{\ver_i}})=y_{Z_\prov}(x_{Z_{\ver_i}}) \GivenExperiment \StateExperiment{Z_{\ver_i}, Z_\prov}\right]\\
    &\flushrow[\sum_{i=1}^k \frac{1}{k} M_{\ver_i,(\prov,\sigma)}]{+\frac{1+\theta}{s}\Big(\sum_{i=1}^k \frac{1}{k}\mathbb{E}\left[\sigma r_{Z_\prov}(x_{Z_{\ver_i}}) \GivenExperiment \StateExperiment{Z_{\ver_i}, Z_\prov}\right]-\sum_{i=1}^k \frac{1}{k}\mathbb{E}\left[\sigma r_{Z_\prov}(x) \GivenExperiment \StateExperiment{x\gets \dist ,\; Z_\prov}\right]\Big)}\\
    &=\frac{1+\theta}{1-c}\Pr\left[h(x_{Z_{\ver_N}})=y_{Z_\prov}(x_{Z_{\ver_N}}) \GivenExperiment \StateExperiment{Z_{\ver_N}, Z_\prov}\right]\\
    &\flushrow[\sum_{i=1}^k \frac{1}{k} M_{\ver_i,(\prov,\sigma)}]{+\frac{1+\theta}{s}\Big(\mathbb{E}\left[\sigma r_{Z_\prov}(x_{Z_{\ver_N}}) \GivenExperiment \StateExperiment{Z_{\ver_N}, Z_\prov}\right]-\mathbb{E}\left[\sigma r_{Z_\prov}(x) \GivenExperiment \StateExperiment{x\gets \dist ,\; Z_\prov}\right]\Big)=:M_{\ver_N,(\prov,\sigma)}.}
\end{align*}
Analogously, there exists a circuit \(\prov_N\) of size \(O(S)\) and randomized sign \(\sigma_{Z_{\prov_N}}\), determined by the randomness \(Z_{\prov_N}\) of \(\prov_N\), such that for all circuits \(\ver \in \mathfrak{A}\)
\begin{align}\label{eq:nash-approx}
    M_{\ver, (\prov_N,\sigma_{Z_{\prov_N}})} \geq \operatorname{val}(M)
-\theta.
\end{align}

\paragraph{Construction of a Transferable Attack.} If \(\operatorname{val}(M) < 1\), we construct a Transferable Attack by setting \(\ver_{\TA}:=\ver_N\). Indeed, take an arbitrary circuit \(\prov\) of size \(4S_L\) that on input \(x\) outputs \((y,r) \in \{0,1\}^2\). Let \(\sigma\) be the sign of
\begin{align*}
    \Pr\left[r_{Z_\prov}(x_{Z_{\ver_{\TA}}})=1 \GivenExperiment \StateExperiment{Z_{\ver_{\TA}}, Z_\prov}\right]-\Pr\left[r_{Z_\prov}(x)=1 \GivenExperiment \StateExperiment{x\gets \dist ,\; Z_\prov}\right].
\end{align*}
Then \((\prov, \sigma) \in \mathfrak{B}\) and (\ref{eq:nash-approx-1}) implies
\begin{multline*}
    \frac{1}{1-c}\Pr\left[h(x_{Z_{\ver_N}})=y_{Z_\prov}(x_{Z_{\ver_N}}) \GivenExperiment \StateExperiment{Z_{\ver_N}, Z_\prov}\right]\\
    +\frac{1}{s}\left|\Pr\left[r_{Z_\prov}(x_{Z_{\ver_N}})=1 \GivenExperiment \StateExperiment{Z_{\ver_N}, Z_\prov}\right]-\Pr\left[r_{Z_\prov}(x)=1 \GivenExperiment \StateExperiment{x\gets \dist ,\; Z_\prov}\right]\right|< 1.
\end{multline*}
Transferability and Indistinguishability follow immediately.

\paragraph{Construction of an Adversarial Defense.} If \(\operatorname{val}(M) \geq 1\), we construct an Adversarial Defense.
Notice that by (\ref{eq:nash-approx}), taking the circuit \(\ver\) that outputs a random \(x \gets \dist\), which reads one labeled pair and so has size \(\prm+1 \leq 4S_L\) and lies in \(\mathfrak{A}\), we immediately have
 \begin{align*}
     \Pr\left[h(x)\neq y_{Z_{\prov_N}}(x) \GivenExperiment \StateExperiment{x \gets \dist ,\; Z_{\prov_N}}\right] \leq 1- \kappa(1-c).
 \end{align*}
 Combining this with learnability yields,
 \begin{align}\label{eq:game-notice}
      &\Pr\left[y_{Z_{\prov_N}}(x)\neq f_{Z_L}(x) \GivenExperiment \StateExperiment{x \gets \dist ,\; Z_L, Z_{\prov_N}}\right] \notag\\
      \quad\leq  &\Pr\left[h(x)\neq y_{Z_{\prov_N}}(x) \GivenExperiment \StateExperiment{x \gets \dist ,\; Z_{\prov_N}}\right] +  \Pr\left[h(x)\neq f_{Z_L}(x) \GivenExperiment \StateExperiment{x \gets \dist ,\; Z_L}\right] \notag\\
      \quad\leq &1- \kappa(1-c) + \varepsilon + \delta.
 \end{align}
We will use this fact a bit later.

Now let \(\prov_{\AD}\) be a circuit that follows the algorithm:
\begin{enumerate}[topsep=0pt, itemsep=0pt, parsep=0pt]
    \item Sample \(Z_L, Z_{\prov_N}\) and a uniformly random bit \(\beta\), and let \(\priv:=(Z_L, Z_{\prov_N}, \beta)\).
    \item Set \(z_{f_{Z_L}} :=L(Z_L)\) and
    \begin{align*}
    R(\priv,x):= \begin{cases}
        \mathbf{1}_{r_{Z_{\prov_N}}(x)=(1+\sigma_{Z_{\prov_N}})/2}, \quad &\text{ if } \beta=0,\\
        \mathbf{1}_{y_{Z_{\prov_N}}(x) \neq f_{Z_L}(x)},\quad &\text{ if } \beta=1.
    \end{cases}
\end{align*}
    \item \textbf{Output:} the representation \(z_{f_{Z_L}}\) of \(f_{Z_L}\) together with \(z_R\).
\end{enumerate}
By learnability, \(\err(f_{Z_L}) < \varepsilon\) with probability at least \(1-\delta\), which is Correctness.
Here \(z_R\) describes the general form of the rule above\footnote{That is, \(z_R\) describes the map \((\priv, x) \mapsto R(\priv,x)\). The private string \(\priv\) is the one drawn in step 1, with the same \(Z_L\) that produced \(z_{f_{Z_L}}\). It is never redrawn, and it is not part of \(z_R\).}, and depends on \(\prm\) alone, so it is a fixed string. The rule class \(\mathcal{R}_\prm\) consists of this single rule, so \(m_{\mathcal{R}} = O(1)\).
The circuit \(\prov_{\AD}\) runs \(L\), runs \(\prov_N\) on \(x\), computes \(f_{Z_L}(x)\) as \(E_\prm(z_{f_{Z_L}},x)\), reads \(x\) and \(\beta\), and emits \(z_{f_{Z_L}}\) and the hard-wired \(z_R\), so \(|\prov_{\AD}| \leq S_L + O(S) + S_E + \prm + O(1) = O(S+S_E)\) by \(\prm + 1 \leq 4S_L\) and \(S_L = O(S)\).

Now let \(\ver\) be any circuit of size \(3S_L\) that on input \((z_f, z_R)\) samples \(Z_\ver\) and outputs \(x \in \{0,1\}^\prm\). We construct a circuit \(\tilde \ver\) from \(\mathfrak{A}\).
\begin{enumerate}[topsep=0pt, itemsep=0pt, parsep=0pt]
    \item Sample \(Z_L, Z_\ver\).
    \item Set \(z_{f_{Z_L}} :=L(Z_L)\).
    \item \textbf{Output:} \(x_{Z_L,Z_\ver}:=\tilde \ver(Z_L,Z_\ver):=\ver(z_{f_{Z_L}}, z_R, Z_\ver)\).
\end{enumerate}
Since \(z_R\) is a fixed string for each \(\prm\), the circuit \(\tilde \ver\) hard-wires it into the corresponding inputs of \(\ver\), which costs no gates. Hence \(|\tilde \ver| \leq |\ver| + |L| \leq 3S_L + S_L = 4S_L\), so \(\tilde \ver \in \mathfrak{A}\) exactly as before. Then applying (\ref{eq:nash-approx}) on \(\tilde \ver\) yields
\begin{multline*}
    \frac{1}{1-c}\Pr\left[h(x_{Z_L,Z_\ver})=y_{Z_{\prov_N}}(x_{Z_L,Z_\ver}) \GivenExperiment \StateExperiment{Z_L, Z_\ver, Z_{\prov_N}}\right]\\
    +\frac{1}{s}\left(\mathbb{E}\left[\sigma_{Z_{\prov_N}}r_{Z_{\prov_N}}(x_{Z_L,Z_\ver}) \GivenExperiment \StateExperiment{Z_L, Z_\ver, Z_{\prov_N}}\right]-\mathbb{E}\left[\sigma_{Z_{\prov_N}}r_{Z_{\prov_N}}(x) \GivenExperiment \StateExperiment{x\gets \dist ,\; Z_{\prov_N}}\right]\right)\geq \kappa.
\end{multline*}
There are two cases. Either
\begin{align*}
    \mathbb{E}\left[\sigma_{Z_{\prov_N}}r_{Z_{\prov_N}}(x_{Z_L,Z_\ver}) \GivenExperiment \StateExperiment{Z_L, Z_\ver, Z_{\prov_N}}\right]-\mathbb{E}\left[\sigma_{Z_{\prov_N}}r_{Z_{\prov_N}}(x) \GivenExperiment \StateExperiment{x\gets \dist ,\; Z_{\prov_N}}\right]\geq \alpha s\kappa,
\end{align*}
which implies Distinguishability
\begin{align*}
    &\mathbb{E}\left[\left|\Pr_{Z_\ver}[R(\priv,x_{Z_L,Z_\ver})=1]-\Pr_{x\gets \dist}[R(\priv,x)=1]\right| \GivenExperiment \StateExperiment{Z_L, Z_{\prov_N} ,\; \beta}\right]\\
    \geq \frac{1}{2}&\mathbb{E}\left[\left|\Pr_{Z_\ver}[ r_{Z_{\prov_N}}(x_{Z_L,Z_\ver})=(1+\sigma_{Z_{\prov_N}})/2]-\Pr_{x\gets \dist}[r_{Z_{\prov_N}}(x)=(1+\sigma_{Z_{\prov_N}})/2]\right| \GivenExperiment \StateExperiment{Z_L, Z_{\prov_N}}\right]\\
    = \frac{1}{2}&\mathbb{E}\left[\left|\sigma_{Z_{\prov_N}}\Pr_{Z_\ver}[ r_{Z_{\prov_N}}(x_{Z_L,Z_\ver})=1]-\sigma_{Z_{\prov_N}}\Pr_{x\gets \dist}[r_{Z_{\prov_N}}(x)=1]\right| \GivenExperiment \StateExperiment{Z_L, Z_{\prov_N}}\right]\\
    = \frac{1}{2}&\mathbb{E}\left[\left|\mathbb{E}_{Z_\ver}[ \sigma_{Z_{\prov_N}}r_{Z_{\prov_N}}(x_{Z_L,Z_\ver})]-\mathbb{E}_{x\gets \dist}[\sigma_{Z_{\prov_N}}r_{Z_{\prov_N}}(x)]\right| \GivenExperiment \StateExperiment{Z_L, Z_{\prov_N}}\right]\\
    \geq \frac{1}{2}&\alpha s\kappa.
\end{align*}
Otherwise,
\begin{align*}
    \Pr\left[h(x_{Z_L,Z_\ver})\neq y_{Z_{\prov_N}}(x_{Z_L,Z_\ver}) \GivenExperiment \StateExperiment{Z_L, Z_\ver, Z_{\prov_N}}\right] \leq 1-(1-c)(1-\alpha)\kappa,
\end{align*}
and hence
 \begin{align*}
&\Pr\left[h(x_{Z_L,Z_\ver})\neq f_{Z_L}(x_{Z_L,Z_\ver}) \GivenExperiment \StateExperiment{Z_L, Z_\ver}\right]\\
  \quad\leq &\Pr\left[y_{Z_{\prov_N}}(x_{Z_L,Z_\ver})\neq f_{Z_L}(x_{Z_L,Z_\ver}) \GivenExperiment \StateExperiment{Z_L, Z_\ver, Z_{\prov_N}}\right]\\
  &\flushrow[\quad\leq]{+ \Pr\left[h(x_{Z_L,Z_\ver})\neq y_{Z_{\prov_N}}(x_{Z_L,Z_\ver}) \GivenExperiment \StateExperiment{Z_L, Z_\ver, Z_{\prov_N}}\right]}\\
  \quad\leq&\Pr\left[y_{Z_{\prov_N}}(x_{Z_L,Z_\ver})\neq f_{Z_L}(x_{Z_L,Z_\ver}) \GivenExperiment \StateExperiment{Z_L, Z_\ver, Z_{\prov_N}}\right]+1-(1-c)(1-\alpha)\kappa.
 \end{align*}
Assume additionally that Distinguishability does not hold:
\begin{align*}
    &\frac{1}{2}\alpha s\kappa>\mathbb{E}\left[\left|\Pr_{Z_\ver}[R(\priv,x_{Z_L,Z_\ver})=1]-\Pr_{x\gets \dist}[R(\priv,x)=1]\right| \GivenExperiment \StateExperiment{Z_L, Z_{\prov_N} ,\; \beta}\right].
\end{align*}
 Together with (\ref{eq:game-notice}), we obtain
 \begin{align*}
    \frac{1}{2}\alpha s\kappa>&\frac{1}{2}\mathbb{E}\left[\left|\Pr_{Z_\ver}[ y_{Z_{\prov_N}}(x_{Z_L,Z_\ver})\neq f_{Z_L}(x_{Z_L,Z_\ver})]-\Pr_{x\gets \dist}[y_{Z_{\prov_N}}(x)\neq f_{Z_L}(x)]\right| \GivenExperiment \StateExperiment{Z_L, Z_{\prov_N}}\right]\\
    \quad\geq&\frac{1}{2}\Big(\Pr\left[y_{Z_{\prov_N}}(x_{Z_L,Z_\ver})\neq f_{Z_L}(x_{Z_L,Z_\ver}) \GivenExperiment \StateExperiment{Z_\ver, Z_L, Z_{\prov_N}}\right]\\
    &\flushrow[\frac{1}{2}\alpha s\kappa>]{-\Pr\left[y_{Z_{\prov_N}}(x)\neq f_{Z_L}(x) \GivenExperiment \StateExperiment{x\gets \dist ,\; Z_L, Z_{\prov_N}}\right]\Big)} \\
    \quad\geq&\frac{1}{2}\Big(\Pr\left[y_{Z_{\prov_N}}(x_{Z_L,Z_\ver})\neq f_{Z_L}(x_{Z_L,Z_\ver}) \GivenExperiment \StateExperiment{Z_\ver, Z_L, Z_{\prov_N}}\right]-1+ \kappa(1-c) - \varepsilon - \delta\Big).
\end{align*}
Rearranging, we conclude
\begin{align*}
    \Pr\left[y_{Z_{\prov_N}}(x_{Z_L,Z_\ver})\neq f_{Z_L}(x_{Z_L,Z_\ver}) \GivenExperiment \StateExperiment{Z_\ver, Z_L, Z_{\prov_N}}\right] <\alpha s\kappa + 1- \kappa(1-c) + \varepsilon + \delta.
\end{align*}
It follows that
\begin{align*}
    s_{\AD}=\varepsilon+\delta+2
-\kappa\left((2-\alpha)(1-c)-s\alpha \right).
\end{align*}

\paragraph{Frequency.} The argument above fixes \(\prm\) and produces, for that \(\prm\), either the attack or the defense. Let \(a_\prm\) and \(b_\prm\) record whether the conditions of the Transferable Attack and of the Adversarial Defense hold at \(\prm\). Then \(a_i + b_i \geq 1\) for all \(i \geq \prm_0\), where \(\prm_0\) is the threshold of learnability and of the size hypotheses, and hence \(\sum_{i \in [\prm]} a_i + \sum_{i \in [\prm]} b_i \geq \prm - \prm_0\). Dividing by \(\prm\) and using subadditivity of the \(\limsup\), the two frequencies sum to at least \(1\), so at least one of them is at least \(1/2\).
\end{proof}
For \(\varepsilon<1/25, \kappa=1-\varepsilon, \alpha=\varepsilon/\kappa, c=10\varepsilon, s=\delta=\varepsilon\) we obtain the following corollary.
\begin{corollary}[Dichotomy with a Single Error Parameter]\label{cor:dichotomy}
Fix a size function \(S\) and a constant \(0<\varepsilon<1/25\). Assume that a learning task \(\task\) is \((\varepsilon, \varepsilon, S_L=\sqrt{S}/(4 \log S), \fclass)\)-learnable, where \(\fclass\) is a class of classifiers. Also assume existence of an evaluation circuit family \(E=\{E_\prm\}_\prm\) of size \(S_E\) for \(\fclass\). Assume that for all sufficiently large \(\prm\) we have \(m_{\fclass}(\prm) \leq 3S_L(\prm)\) and \(\prm + 1 \leq 4S_L(\prm)\).
Then, at least one of the following exists:
\[\ver_{\TA} \in_{1/2} \TA\left(\task, 1, O(S), \sqrt{S}/(\log S), 10\varepsilon, \varepsilon\right),\]
or
\[\prov_{\AD} \in_{1/2} \AD\left(\task,1,3\sqrt{S}/(4\log S),O(S+S_E),
\varepsilon,\varepsilon, \varepsilon^2/2, 25\varepsilon\right).\]
\end{corollary}

\subsection{Explicit Classification Tasks with a Transferable Attack or an Adversarial Defense}
\label{sec:constructions}
\subsubsection{A Transferable Attack from Public-Key Encryption}\label{sec:construction_PKE}
Here we construct a learning task \(\task_{\mathsf{PKE}}\) using a public-key encryption scheme \(\mathsf{PKE}=(\Gen, \Enc, \Dec)\) that is indistinguishable under chosen-plaintext attacks (IND-CPA secure, Definition~\ref{def:pke} in Appendix~\ref{app:crypto}) with message space \(\{0,1\}\). As a concrete example we can take Regev's PKE \citep{regev2009lattices} with errors truncated so that decryption never fails.  We will also prove that \(\task_{\mathsf{PKE}}\) admits a Transferable Attack. In the proofs of this subsection a defender \(\prov_\prm^{\exh}\) outputs on an input \(x\) a pair \((y,r) \in \{0,1\}^2\), as in the rest of Section~\ref{sec:classification}. We write \(\prov_\prm(x)\) for its answer \(y\) where it is compared with a label, and for its flag \(r\) where it is compared with \(1\).

For each \(\lambda\) let \(\prm=\prm(\lambda)\) be the sum of the public key and ciphertext lengths, and one. For such \(\prm=\prm(\lambda)\) and \((\pk, \sk) \gets \Gen(1^\lambda)\) define the distribution
\begin{align*}
    \dist_\pk = \left\{(\pk, y, c) \GivenExperiment \StateExperiment{y \gets \{0,1\} ,\; c \gets \Enc(\pk, y)}\right\}.
\end{align*}
Each \(x=(\pk, y, c) \gets \dist_\pk\) is labeled by \(y\). More generally, for any \(x\in \{0,1\}^\prm\), parse \(x=(\pk', y, c)\) and set \(h_{(\pk, \sk)}(x):= \Dec(\sk, c)\) if \(\pk= \pk'\) and \(h_{(\pk, \sk)}(x):=0\), otherwise.

If \(\prm=\prm(\lambda)\) for some security parameter \(\lambda\), then
\begin{align*}
    \task_{\mathsf{PKE},\prm} = \left\{\big(\dist_\pk, h_{(\pk, \sk)}\big) \GivenExperiment \StateExperiment{(\pk, \sk) \gets \Gen(1^\lambda)}\right\}.
\end{align*}
If there is no such \(\lambda\), take the largest \(\lambda\) with \(\prm(\lambda) \leq \prm\) and let \(\task_{\mathsf{PKE},\prm}\) be the task above with every input padded to length \(\prm\) by trailing zeros, which \(h_{(\pk, \sk)}\) ignores. For the finitely many \(\prm\) below \(\prm(1)\), let \(\task_{\mathsf{PKE},\prm}\) be the distribution that always outputs the pair of the uniform distribution over \(\{0,1\}^\prm\) and \(h_\prm\equiv 0\).

For \(\prm \geq \prm(1)\) let \(\fclass_\prm\) be the singleton containing the function that maps \(x=(\pk', y, c)\) to \(y\), reading \(x\) up to the padding. Otherwise, set \(\fclass_\prm=\{f\equiv0\}\).
\paragraph{Learnability of \(\task_{\mathsf{PKE}}\)}
We show that for any \(\varepsilon, \delta >0 \), \(\task_{\mathsf{PKE}}\) is an \((\varepsilon, \delta, O(1), \fclass)\)-learnable task.

    For any \(\prm\), let the learner \(L_\prm\) be the circuit that outputs the description \(z_f\) of the unique function \(f \in \fclass_\prm\).
    For any \(\prm \leq \prm(1)\), the unique \(f\in \fclass_\prm\) clearly has zero error with probability one.

    Otherwise, for any \((\pk, \sk) \gets \Gen(1^\lambda)\) and any \(x=(\pk, y, c) \gets \dist_\pk\) we have \(c=\Enc(\pk, y)\) and hence
    \begin{align*}
        f(x) =y= \Dec(\sk, c) = h_{(\pk, \sk)}(x).
    \end{align*}
    Again, \(f\) has error zero with probability one.

Now we construct a Transferable Attack for \(\task_{\mathsf{PKE}}\). For each \(\prm\geq \prm(1)\), let \(\ver_{\prm,\TA}\) be a circuit that follows the algorithm:
\begin{enumerate}[topsep=0pt, itemsep=0pt, parsep=0pt]
    \item Sample \(x\gets \dist_\pk\) to extract a public key \(\pk\).
    \item Sample \((y,y') \gets \{0,1\}^2\).
    \item \textbf{Output:} \((\pk, y', \Enc(\pk, y))\).
\end{enumerate}
For the finitely many \(\prm < \prm(1)\) it outputs \(0^\prm\).
\paragraph{Transferable Attack for \(\task_{\mathsf{PKE}}\)} We show that
    \[\ver_{\TA} \in_{1} \TA(\task_{\mathsf{PKE}}, 1, \poly(\prm), \poly(\prm), 1/2 - \negl(\prm), \negl(\prm)).\]

    Fix \(\prm \geq \prm(1)\) and let \(\lambda\) be the security parameter attached to \(\prm\), so that \(\lambda = \prm^{\Omega(1)}\), \(\negl(\lambda) = \negl(\prm)\), and circuits of size \(\poly(\prm)\) are of size \(\poly(\lambda)\).
    To prove Transferability we need to show that for any polynomial-size \(\prov_\prm\)
    \[
    \Pr\left[h_{(\pk, \sk)}(\pk, y', \Enc(\pk, y)) \neq \prov_\prm(\pk, y', \Enc(\pk, y)) \GivenExperiment \StateExperiment{(\pk, \sk)\gets \Gen(1^\lambda) \\ (y, y') \gets \{0,1\}^2}\right] \geq 1/2 - \negl(\lambda),
\]
    which is equivalent to
    \begin{align*}
        \Pr\left[\prov_\prm(\pk, y', \Enc(\pk, y)) = y \GivenExperiment \StateExperiment{(\pk, \sk)\gets \Gen(1^\lambda) ,\; (y, y') \gets \{0,1\}^2}\right] \leq 1/2 + \negl(\lambda)
    \end{align*}
    and directly follows from the IND-CPA security of \(\mathsf{PKE}\) as in (\ref{eq:ind-cpa-guess}).

    Indistinguishability says that for any polynomial-size \(\prov_\prm\)
    \begin{multline*}
        \left|\Pr\left[\prov_\prm(\pk, y', \Enc(\pk, y))=1 \GivenExperiment \StateExperiment{(\pk, \sk)\gets \Gen(1^\lambda) ,\; (y, y') \gets \{0,1\}^2}\right]\right. \\
        \left.- \Pr\left[\prov_\prm(\pk, y, \Enc(\pk, y))=1 \GivenExperiment \StateExperiment{(\pk, \sk)\gets \Gen(1^\lambda) ,\; y \gets \{0,1\}}\right]\right| < \negl(\lambda).
    \end{multline*}
    Assume the opposite. Then there is a non-negligible function \(e\) such that
    \begin{multline*}
        \Big|\Pr_{(\pk, \sk)\gets \Gen(1^\lambda)}[\prov_\prm(\pk, 1, \Enc(\pk, 0))=1] + \Pr_{(\pk, \sk)\gets \Gen(1^\lambda)}[\prov_\prm(\pk, 0, \Enc(\pk, 1))=1] \\
        - \Pr_{(\pk, \sk)\gets \Gen(1^\lambda)}[\prov_\prm(\pk, 0, \Enc(\pk, 0))=1] - \Pr_{(\pk, \sk)\gets \Gen(1^\lambda)}[\prov_\prm(\pk, 1, \Enc(\pk, 1))=1] \Big| \geq e(\lambda).
    \end{multline*}
    Without loss of generality assume that
    \begin{align*}
        &\Big|\Pr_{(\pk, \sk)\gets \Gen(1^\lambda)}[\prov_\prm(\pk, 1, \Enc(\pk, 0))=1] - \Pr_{(\pk, \sk)\gets \Gen(1^\lambda)}[\prov_\prm(\pk, 1, \Enc(\pk, 1))=1] \Big| \geq \frac{e(\lambda)}{2}.
    \end{align*}
    This is a contradiction to the IND-CPA security of \(\mathsf{PKE}\) as in (\ref{eq:ind-cpa}).

    In both parts, the IND-CPA adversary answers the oracle queries of \(\prov_\prm\) itself, by sampling \(y_i \gets \{0,1\}\) and encrypting it under \(\pk\).
\subsubsection{A Transferable Attack for a Lifting of Any Task}
Now let \(\task=\{\task_\lambda\}_\lambda\) be any binary classification task. We will provide a way to append additional dimensions to \(x\in \{0,1\}^\lambda\) not changing labels of natural inputs such that the resulting learning task \(\task'\) admits a Transferable Attack.

Let \(\mathsf{PKE}\) be an IND-CPA secure public-key encryption scheme as before. For each \(\lambda\), let \(\prm=\prm(\lambda)\) be the sum of the public key and ciphertext lengths, and \(\lambda\).\footnote{We will consider such \(\prm\)'s only and take care of the remaining \(\prm\)'s as before.} For each \(\lambda\), we modify distribution \(\dist_\lambda\) and \(h_\lambda\) where \((\dist_\lambda, h_\lambda) \gets \task_\lambda\). For each \(\prm \geq \prm(1)\) and \((\pk, \sk)\gets \Gen(1^\lambda)\) let
\begin{align*}
    \dist'_{\dist_\lambda,\pk} = \left\{(\pk, x, c) \GivenExperiment \StateExperiment{x \gets \dist_\lambda ,\; c \gets \Enc(\pk, 0)}\right\},
\end{align*}
and write \(x'\) for a draw from it.
Correctness of the \(\mathsf{PKE}\) says that if we decrypt \(c\) from \(x'=(\pk, x,c)\gets \dist'_{\dist_\lambda,\pk}\), then we get zero. Thus, adding \(\Dec(\sk, c)\) to the true label \(h_\lambda(x)\) of \(x\) does not change anything. We extend this function to all \(x'\in \{0,1\}^\prm\). Set
\begin{align*}
    h'_{h_\lambda, (\pk, \sk)}(x') = \begin{cases}
        h_\lambda(x)\oplus \Dec(\sk, c), & \text{ if } x'=(\pk, x,c)\\
        0, & \text{ otherwise.}
    \end{cases}
\end{align*}
Similarly to the previous construction,
\begin{align*}
    \task'_{\prm} = \left\{\big(\dist'_{\dist_\lambda,\pk}, h'_{h_\lambda,(\pk, \sk)}\big) \GivenExperiment \StateExperiment{(\pk, \sk) \gets \Gen(1^\lambda) ,\; (\dist_\lambda, h_\lambda) \gets \task_\lambda}\right\}.
\end{align*}
It is easy to see that the modified learning task \(\task'\) is learnable if and only if \(\task\) is learnable.

For each \(\prm\geq \prm(1)\), let \(\ver'_{\prm,\TA}\) be a circuit that follows the algorithm:
\begin{enumerate}[topsep=0pt, itemsep=0pt, parsep=0pt]
    \item Sample \(x'\gets \dist'_{\dist_\lambda,\pk}\) to extract a public key \(\pk\) and a natural input \(x\).
    \item Sample a bit \(b \gets \{0,1\}\).
    \item \textbf{Output:} \((\pk, x, \Enc(\pk, b))\).
\end{enumerate}
For the finitely many \(\prm < \prm(1)\) it outputs \(0^\prm\).
\paragraph{Transferable Attack for the Lifted Task \(\task'\)} We show that
    \[\ver'_{\TA} \in_{1} \TA(\task', 1, \poly(\prm), \poly(\prm), 1/2 - \negl(\prm), \negl(\prm)).\]
    Fix \(\prm \geq \prm(1)\) and let \(\lambda\) be the security parameter attached to \(\prm\), so that \(\lambda = \prm^{\Omega(1)}\), \(\negl(\lambda) = \negl(\prm)\), and circuits of size \(\poly(\prm)\) are of size \(\poly(\lambda)\).
    To prove Transferability we need to show that for any polynomial-size \(\prov_\prm\)
    \[
    \Pr\left[h'_{h_\lambda,(\pk, \sk)}(\pk, x, \Enc(\pk, b)) \neq \prov_\prm(\pk, x, \Enc(\pk, b)) \GivenExperiment \StateExperiment{(\pk, \sk)\gets \Gen(1^\lambda) \\ (\dist_\lambda, h_\lambda) \gets \task_\lambda \\ x\gets \dist_\lambda \\ b \gets \{0,1\}}\right]
    \geq 1/2 - \negl(\lambda),
\]
    which is equivalent to
    \[
    \mathbb{E}\left[\Pr\left[\prov_\prm(\pk, x, \Enc(\pk, b)) = h_\lambda(x)\oplus b \GivenExperiment \StateExperiment{(\pk, \sk)\gets \Gen(1^\lambda) \\ b \gets \{0,1\}}\right] \GivenExperiment \StateExperiment{(\dist_\lambda, h_\lambda) \gets \task_\lambda \\ x\gets \dist_\lambda}\right]
    \leq 1/2 + \negl(\lambda).
\]
    Assume the opposite. Then there is a non-negligible function \(e\) such that, for infinitely many \(\lambda\), the expectation above is at least \(1/2 + e(\lambda)\). By the averaging argument, there exists a pair \((\dist^*_\lambda, h^*_\lambda)\) and an input \(x^* \in \operatorname{supp}(\dist^*_\lambda)\) such that
    \begin{align*}
        \Pr\left[\prov_\prm(\pk, x^*, \Enc(\pk, b)) = h^*_\lambda(x^*)\oplus b \GivenExperiment \StateExperiment{(\pk, \sk)\gets \Gen(1^\lambda) ,\; b \gets \{0,1\}}\right] \geq 1/2 + e(\lambda).
    \end{align*}
    The adversary that hard-codes \(x^*\) and \(h^*_\lambda(x^*)\) and answers \(\prov_\prm(\pk, x^*, c) \oplus h^*_\lambda(x^*)\) on a challenge \(c\) then guesses the encrypted bit with probability at least \(1/2 + e(\lambda)\). This is a contradiction to the IND-CPA security of \(\mathsf{PKE}\) as in (\ref{eq:ind-cpa-guess}).

    Indistinguishability says that for any polynomial-size \(\prov_\prm\)
    \begin{multline*}
        \left|\Pr\left[\prov_\prm(\pk, x, \Enc(\pk, b))=1 \GivenExperiment \StateExperiment{(\pk, \sk)\gets \Gen(1^\lambda) ,\; (\dist_\lambda, h_\lambda) \gets \task_\lambda ,\; x\gets \dist_\lambda ,\; b \gets \{0,1\}}\right]\right.\\
        \left.- \Pr\left[\prov_\prm(\pk, x, \Enc(\pk, 0))=1 \GivenExperiment \StateExperiment{(\pk, \sk)\gets \Gen(1^\lambda) ,\; (\dist_\lambda, h_\lambda) \gets \task_\lambda ,\; x\gets \dist_\lambda}\right]\right| < \negl(\lambda).
    \end{multline*}
    Assume the opposite. Then there is a non-negligible function \(e\) such that, for infinitely many \(\lambda\), the difference above is at least \(e(\lambda)\) in absolute value. Write \(X_{x}\) for this difference conditioned on \((\dist_\lambda, h_\lambda)\) and \(x\). Then \(|\mathbb{E}[X_x]| \geq e(\lambda)\), hence \(\mathbb{E}|X_x| \geq e(\lambda)\), and by the averaging argument there exists \((\dist^*_\lambda, h^*_\lambda)\) and an input \(x^* \in \operatorname{supp}(\dist^*_\lambda)\) such that \(|X_{x^*}| \geq e(\lambda)\). Since \(b\) is uniform,
    \[
    X_{x^*} = \frac{1}{2}\Big(\Pr_{(\pk, \sk)\gets \Gen(1^\lambda)}[\prov_\prm(\pk, x^*, \Enc(\pk, 1))=1]
    - \Pr_{(\pk, \sk)\gets \Gen(1^\lambda)}[\prov_\prm(\pk, x^*, \Enc(\pk, 0))=1]\Big),
\]
    so the adversary \(\prov_\prm(\pk, x^*, \cdot)\), with \(x^*\) hard-coded, distinguishes encryptions of \(0\) and \(1\) with advantage at least \(2e(\lambda)\). This is a contradiction to the IND-CPA security of \(\mathsf{PKE}\) as in (\ref{eq:ind-cpa}).

    In both parts, the IND-CPA adversary also answers the oracle queries of \(\prov_\prm\). The natural inputs \(x_i\) and their labels \(h^*_\lambda(x_i)\) are independent of \(\pk\), so the averaging argument can fix them together with \(x^*\), and the adversary hard-codes them and encrypts \(0\) under \(\pk\) itself.
\subsubsection{An Adversarial Defense for a Simple Hypothesis Class}
For each \(\prm\), let \(\mathbb{D}_\prm\) be a distribution over \(\Delta(\{0,1\}^\prm)\) and \(\mathcal{H}_\prm\) be a fixed finite PAC learnable hypothesis class of Boolean functions with domain \(\{0,1\}^\prm\).

Let \(\dist_\prm \gets \mathbb{D}_\prm\) be a distribution over \(\{0,1\}^\prm\) and \(h_\prm \gets \mathcal{H}_\prm\) a hypothesis. Fix \(\varepsilon,\delta>0\). Then there exists \(\ell=\ell(\delta, \varepsilon)\) such that with probability \(\geq 1-\delta\), an empirical risk minimization (ERM) algorithm \citep{vc_dim1971} on \(\ell\) samples of the form \((x_i, h_\prm(x_i)), x_i \gets \dist_\prm\) returns a hypothesis \(f \in \mathcal{H}_\prm\) with error at most \(\varepsilon/2\), in particular less than \(\varepsilon\).

Let
\begin{align*}
    \task_\prm = \left\{(\dist_\prm, h_\prm) \GivenExperiment \StateExperiment{\dist_\prm \gets \mathbb{D}_\prm ,\; h_\prm \gets \mathcal{H}_\prm}\right\}.
\end{align*}
We construct an Adversarial Defense \(\prov\) based on an ERM algorithm for \(\mathcal{H}_\prm\) and selective classification \citep{el2010foundations}. The following algorithm defines \(\prov_\prm\).
\begin{enumerate}[topsep=0pt, itemsep=0pt, parsep=0pt]
    \item Sample \(\mathcal{S}:=\{(x_i, h_\prm(x_i))\}_{i \in [\ell]}\), where each \((x_i, h_\prm(x_i)) \gets \exfor{\dist_\prm}{h_\prm}\).
    \item Run an ERM algorithm to find \(f \in \mathcal{H}_\prm\) with zero empirical error.
    \item The distinguishing rule \(R\) works as follows: For any \(x \in \{0,1\}^\prm\), run ERM on \(\mathcal{S}\cup \{(x,0)\}\) and \(\mathcal{S}\cup \{(x,1)\}\). If both runs are successful, output \(R(\priv,x)=1\). Otherwise, output \(R(\priv,x)=0\).
    \item \textbf{Output:} the representations \(z_f\) of \(f\) and \(z_R\) of \(R\).\footnote{Here \(z_R\) describes the rule of step 3 without \(\mathcal{S}\). The rule is instantiated with the sample \(\mathcal{S}\) drawn in step 1, which \(\prov\) keeps as its private string \(\priv{}=\mathcal{S}\) and reuses whenever it evaluates \(R\).}
\end{enumerate}
\paragraph{Adversarial Defense from ERM}
    Let \(T\) be the runtime of an ERM algorithm and let
    \begin{align*}
    \mu:=\mathbb{E}
    \left[\Pr_{x \gets \dist_\prm}[R(\priv,x)=1] \GivenExperiment \StateExperiment{(\dist_\prm, h_\prm) \gets \task_\prm ,\; \mathcal{S} \gets \exfor{\dist_\prm}{h_\prm}^\ell}\right]
    \end{align*}
    be the expected value of the probability that there are two functions in the hypothesis class \(\mathcal{H}_\prm\) that are consistent with \(\mathcal{S}\) but disagree on \(x\). Then for all \(\tau > \mu\)\footnote{In the display below, \(S_\ver= \infty\) indicates that the defense works against attackers of any size.}
    \[\prov \in_{1} \AD(\task, 1, \infty, \tilde O(T), \varepsilon, \delta, \tau-\mu, \tau).\]
    Correctness follows from the PAC learnability of \(\mathcal{H}_\prm\). Let \(\ver_\prm\) be any attacker. If
    \[
        \Pr\left[h_\prm(x) \neq f(x) \GivenExperiment \StateExperiment{(\dist_\prm, h_\prm) \gets \task_\prm ,\; \mathcal{S} \gets \exfor{\dist_\prm}{h_\prm}^\ell ,\; f \gets \mathsf{ERM}(\mathcal{S}, \mathcal{H}_\prm) ,\; x \gets \ver_\prm(z_f, z_R)}\right]< \tau
    \]
    then Robustness holds and we are done. Otherwise, notice that inequality \(h_\prm(x) \neq f(x)\) implies that ERM on both \(\mathcal{S}\cup \{(x,0)\}\) and \(\mathcal{S}\cup \{(x,1)\}\) will be successful. In other words,
    \[
    \Pr\left[R(\priv,x)=1 \GivenExperiment \StateExperiment{(\dist_\prm, h_\prm) \gets \task_\prm \\ \mathcal{S} \gets \exfor{\dist_\prm}{h_\prm}^\ell \\ f \gets \mathsf{ERM}(\mathcal{S}, \mathcal{H}_\prm) \\ x \gets \ver_\prm(z_f, z_R)}\right] \geq \Pr\left[h_\prm(x) \neq f(x) \GivenExperiment \StateExperiment{(\dist_\prm, h_\prm) \gets \task_\prm \\ \mathcal{S} \gets \exfor{\dist_\prm}{h_\prm}^\ell \\ f \gets \mathsf{ERM}(\mathcal{S}, \mathcal{H}_\prm) \\ x \gets \ver_\prm(z_f, z_R)}\right]\geq \tau.
\]
    On the other hand, \(\mu\) is exactly the expected value of the probability that \(R(\priv,x)=1\) for natural \(x \gets \dist_\prm\). Hence,
    \begin{multline*}
    \mathbb{E}\left[\Pr_{x \gets \ver_\prm(z_f, z_R)}[R(\priv,x)=1] \GivenExperiment \StateExperiment{(\dist_\prm, h_\prm) \gets \task_\prm ,\; \mathcal{S} \gets \exfor{\dist_\prm}{h_\prm}^\ell ,\; f \gets \mathsf{ERM}(\mathcal{S}, \mathcal{H}_\prm)}\right]\\
    -\mathbb{E}\left[\Pr_{x \gets \dist_\prm}[R(\priv,x)=1] \GivenExperiment \StateExperiment{(\dist_\prm, h_\prm) \gets \task_\prm ,\; \mathcal{S} \gets \exfor{\dist_\prm}{h_\prm}^\ell ,\; f \gets \mathsf{ERM}(\mathcal{S}, \mathcal{H}_\prm)}\right] \geq \tau-\mu.
    \end{multline*}
    This implies Distinguishability.
\paragraph{Instantiation}    
To give a concrete example, consider a simple threshold learning task, where \(\{0,1\}^\prm\) is identified with \(\{0,1, \dots, 2^\prm-1\}\), \(\dist_\prm\) is the uniform distribution over \(\{0,1\}^\prm\) and \(\mathcal{H}_\prm\) is the set of threshold functions \(h_k=\mathbf{1}_{x\geq k}\). The VC dimension of \(\mathcal{H}_\prm\) is equal to one. Therefore, the hypothesis class is PAC learnable with sample complexity \(\ell = O((\log(1/\varepsilon) + \log(1/\delta))/\varepsilon)\). For this \(\ell\), with probability at least \(1-\delta\) over \(\mathcal{S}\), every hypothesis in \(\mathcal{H}_\prm\) consistent with \(\mathcal{S}\) has error at most \(\varepsilon\), which is the uniform bound used below. Let \(\task_\prm\) be a uniform distribution over pairs \(\{(\dist_\prm, h_k): h_k \in \mathcal{H}_\prm\}\). Let \((\dist_\prm, h_k)\) be the pair drawn from \(\task_\prm\). Notice that
\begin{multline*}
    \mu=\mathbb{E}\Bigg[\Pr_{x \gets \dist_\prm}[\exists f_1,f_2 \in \mathcal{H}_\prm : f_1(x)\neq f_2(x),\\
    f_1(x_i)=f_2(x_i)=h_k(x_i)\ \text{for all}\ i \in [\ell]] \Biggm| \StateExperiment{(\dist_\prm, h_k) \gets \task_\prm ,\; \mathcal{S}:=\{(x_i,h_k(x_i))\}_{i \in [\ell]} \gets \exfor{\dist_\prm}{h_k}^\ell}\Bigg]
\end{multline*}
and for all \(h_j\in \mathcal{H}_\prm\)
\begin{align*}
    \err_{\dist_\prm,h_k}(h_j)=\frac{|j-k|}{2^\prm}.
\end{align*}
Hence,
\begin{align*}
    \mu&\leq\Pr_{x \gets \dist_\prm}[\exists h_i, h_j \in \mathcal{H}_\prm : \err_{\dist_\prm,h_k}(h_i),\err_{\dist_\prm,h_k}(h_j) \leq \varepsilon, h_i(x) \neq h_j(x)]+\delta\\&\leq
    \Pr_{x \gets \dist_\prm}[\exists i<j : |j-k|, |i-k|\leq 2^\prm\varepsilon, i\leq x < j]+ \delta\\
    &\leq \Pr_{x \gets \dist_\prm}[k-2^\prm\varepsilon \leq x < k+2^\prm\varepsilon]+\delta\\
    &\leq 2\varepsilon+ \delta.
\end{align*}
The interval in the third line contains at most \(2 \cdot 2^\prm\varepsilon\) integers, which gives the last inequality.
A simple ERM algorithm for the learning task would be to scan the sample set \(\mathcal{S}\) and to output \(k'=\max\{x_i : y_i =0\}+1\) resulting in running time \(O(\ell \prm)\). Hence,
 \[\prov \in_{1} \AD(\task, 1, \infty, \tilde O((\prm\log(1/\varepsilon) + \prm\log(1/\delta))/\varepsilon), \varepsilon, \delta, \tau-2\varepsilon-\delta, \tau),\]
 for any \(\tau  >2\varepsilon+\delta\).

%% file: sections/generation.tex
In this section we construct two generative learning tasks, \(\taskcon\) and \(\tasktime\), on which an Active Adversarial Defense exists while no Adversarial Defense does. In both tasks inputs will be associated with levels. A correct answer to an input of a higher level requires more samples or more sequential computation, and encryption hides the level of an input from the published model, so an inference phase can answer the input it actually receives, while a model fixed in advance cannot answer every input an attacker can produce.

In both tasks inputs and outputs are bit strings of the same length, \(\inspace = \outspace = \{0,1\}^{\poly(\prm)}\), for a polynomial that is fixed by cryptographic constructions. We use batch size \(q = 1\) throughout.

The constructions use cryptographic primitives defined in Appendix~\ref{app:crypto}: \(\taskcon\) uses IB-FHE (Definition~\ref{def:ibfhe}), strongly unforgeable signatures (Definition~\ref{def:signatures}) and zero-knowledge SNARKs (Definition~\ref{def:zk-snark}), and \(\tasktime\) uses IB-FHE, IVC (Definition~\ref{def:ivc}) and NPLs (Definition~\ref{def:npl}).

Let \(\mathsf{IB\text{-}FHE} = (\mathsf{Setup}, \mathsf{KeyGen}, \Enc, \Dec, \Eval)\) be an IB-FHE scheme (Definition~\ref{def:ibfhe}) for the class of polynomially-sized circuits, with message space \(\{0,1\}\), applied bit by bit to strings. For both constructions, we assume that the IB-FHE scheme admits
a predicate \(\mathsf{Adm}(\mathsf{MSK},\idt,c)\).
Within each task instance, fix one secret key per identity,
shared by the example oracle and \(h\), and write \(\sk_{\idt}\gets\mathsf{KeyGen}(\mathsf{MSK},\idt)\) for this fixed key.
If \(\mathsf{Adm}(\mathsf{MSK},\idt,c)=1\), then
\(u:=\Dec(\sk_{\idt},c)\) is well defined and
\[
    \Dec\bigl(
        \sk_{\idt},\Eval(\mathsf{pp}_{\mathsf{fhe}},C,c)
    \bigr)=C(u)
\]
for every clear-input circuit \(C\).
Under an independently sampled fresh identity, a fresh encryption is admissible except with negligible probability.

Also both tasks use the geometric distribution \(\mathsf{Geom}(\frac12)\), which samples \(\ell \in \N\) with probability \(2^{-\ell}\).

In both tasks, every input in the support of \(\dist\) also contains, in the clear, the public parameters of the task instance, which \(h\) ignores, so every party reads them off any input it receives or samples. For \(\taskcon\) these are public parameters \(\mathsf{pp}_{\mathsf{fhe}}\) and \(\mathsf{pp}_{\mathsf{snark}}\) and a verification key  \(\mathsf{vk}\). For \(\tasktime\) they are public parameters \(\mathsf{pp}_{\mathsf{fhe}}\) and \(\mathsf{pp}_{\mathsf{ivc}}\) and input \(\chi\).

\subsection{Separation in Sample Complexity}
\label{sec:separation-samples}

\subsubsection{\texorpdfstring{The Learning Task \(\taskcon\)}{The Learning Task L\textasciicircum data}}
Let \((\Gen, \mathsf{Sign}, \mathsf{Verify})\) be a strongly unforgeable signature scheme.
For the verification key \(\mathsf{vk}\) of a key pair \((\mathsf{vk}, \sk) \gets \Gen(1^\prm)\), sampled as part of the task below, let \(M\) be the machine, with \(\mathsf{vk}\) hard-wired, that on input \(\ell \in \N\) and a potential witness \((\sigma_i)_{i \in [\ell]}\) accepts if and only if the \(\sigma_i\) are pairwise distinct\footnote{
Every signature in \(\taskcon\) is on the message \(0\). We make
signatures pairwise distinct by signing a fresh nonce: \(\mathsf{Sign}(\sk,0)\)
stands for \((\omega,\mathsf{Sign}(\sk,\langle 0,\omega\rangle))\) with
\(\omega\gets\{0,1\}^{\prm}\), and \(\mathsf{Verify}\) checks this encoding.} and \(\mathsf{Verify}(\mathsf{vk},0,\sigma_i) = 1\) for every \(i \in [\ell]\). Let \(t_\ell\) be a fixed polynomial in \(\ell\) and \(\prm\), part of the task, that bounds the running time of \(M\) on input \(\ell\) and the lengths of the instance and the witness, so that \((M,\ell,t_\ell)\) is an instance of the universal relation (Definition~\ref{def:universal-relation}). The machine \(M\) depends on \(\prm\) and, through \(\mathsf{vk}\), on the task instance, and the bounds \(t_\ell\) depend on \(\prm\). We suppress these dependencies. Let \((\gensnark, \provsnark, \versnark)\) be a publicly verifiable zk-SNARK (Definition~\ref{def:zk-snark}) for the universal relation \(\mathsf{R}_U\). 

For \(\prm  \in \N\) we define the learning task \(\taskcon_\prm\) as the result of the following procedure: 
\[\taskcon_\prm =
 \left\{
(\dist_{\mathsf{vk},\sk,\mathsf{pp}_\mathsf{fhe},\mathsf{MSK},\mathsf{pp}_\mathsf{snark}},h_{\mathsf{vk},\sk,\mathsf{pp}_\mathsf{fhe},\mathsf{MSK},\mathsf{pp}_\mathsf{snark}})
\GivenExperiment
\StateExperiment{
(\mathsf{vk},\sk) \gets \Gen(1^\prm) \\
(\mathsf{pp}_\mathsf{fhe}, \mathsf{MSK}) \gets \mathsf{Setup}(1^\prm) \\
\mathsf{pp}_\mathsf{snark} \gets \gensnark(1^\prm, 2^\prm)\footnotemark
}
\right\}.
\]\footnotetext{The time bound \(2^\prm\) lets the SNARK prove every polynomial-time computation once \(\prm\) is large enough.}

Next, we define \(\dist_{\mathsf{vk},\sk,\mathsf{pp}_\mathsf{fhe},\mathsf{MSK},\mathsf{pp}_\mathsf{snark}}\).
For the rest of this subsection we omit the subscripts.
\(\dist\) will have two equally probable parts \(\dist_\mathsf{Clear}\) and \(\dist_\mathsf{Enc}\), that is,
\[
\dist = \frac12 \dist_\mathsf{Clear} + \frac12 \dist_\mathsf{Enc}.
\]

For every \((x,y) \in \mathsf{supp}(\dist_\mathsf{Clear})\) we have
\[
x = \sigma \| \ell  \| \pi  , y = \sigma \| \ell^+ \|  \pi',
\]
\(\mathsf{Verify}(\mathsf{vk},0,\sigma) = 1\), \(\pi\) is a zk-SNARK proof of knowledge of \(\ell\) signatures of \(0\).
Additionally, \(\pi'\) is a zk-SNARK proof of knowledge of \(\ell^+\) signatures of \(0\). 

More precisely, we set
\[
\dist_\mathsf{Clear} = \left\{
( \sigma \| \ell \| \pi, \sigma \| \ell^+ \|  \pi')
\GivenExperiment
\StateExperiment{
\ell \gets \mathsf{Geom}(\frac12) \\
\sigma \gets \mathsf{Sign}(\sk,0) \\
(\sigma'_i)_{i \in [\ell^+]} \gets \mathsf{Sign}(\sk,0) \\
(\sigma_i)_{i \in [\ell]} \gets \mathsf{Sign}(\sk,0) \\
\pi \gets \provsnark \left(\mathsf{pp_{snark}},(M,\ell,t_\ell),(\sigma_i)_{i \in [\ell]} \right) \\
\pi' \gets \provsnark \left(\mathsf{pp_{snark}},(M,\ell^+,t_{\ell^+}),(\sigma'_i)_{i \in [\ell^+]} \right) 
}
\right\}.
\]

The distribution \(\dist_\mathsf{Enc}\) is \(\dist_\mathsf{Clear}\) under an IB-FHE encryption.
More formally, let \(\mathcal{I}\) be a space of identities of size \(2^{\Omega(\prm)}\).
Let
\[
\dist_\mathsf{Enc} = \left\{
(x^{\mathsf{enc}}\|\idt_1\|\idt_2\|\sk_{\idt_2},y^{\mathsf{enc}})
\GivenExperiment
\StateExperiment{
\idt_1,\idt_2 \gets \mathcal{I} \\
(x,y) \gets \dist_\mathsf{Clear} \\
x^\mathsf{enc} \gets \Enc(\mathsf{pp}_{\mathsf{fhe}},\idt_1,x) \\
y^\mathsf{enc} \gets \Enc(\mathsf{pp}_{\mathsf{fhe}},\idt_1,y) \\
\sk_{\idt_2} \gets \mathsf{KeyGen}(\mathsf{MSK},\idt_2)
}
\right\}.
\]
Finally, every input and answer of \(\dist\), clear or encrypted, is padded with zeros to one common length. Also, every input and answer begins with one bit indicating whether it is clear or encrypted.

 We say that \((\sigma \| \ell  \| \pi, \sigma \| \ell^+ \|  \pi') \in \mathsf{supp}(\dist)\) is of hardness level \(\ell\).

Oracle \(h\)\footnote{Here we also omit a subscript.} is defined via the following algorithm

\begin{enumerate}[topsep=0pt, itemsep=0pt, parsep=0pt]
    \item Determine whether \(x\) has the clear or
    encrypted input format. If neither, return \(0\).\footnote{The attacker may submit inputs outside the support
    of the natural distribution. Invalid inputs incur no error,
    but invalid answers to valid inputs do.}

    \item If \(x\) is encrypted, parse
    \[
        x=x^{\mathsf{enc}}\|\idt_1\|\idt_2\|\sk_{\idt_2}.
    \]
    If \(\mathsf{Adm}(\mathsf{MSK},\idt_1,x^{\mathsf{enc}})=0\), return \(0\).
    Otherwise, retrieve the fixed key \(\sk_{\idt_1}\) and replace
    \(x\) by
    \[
        \Dec(\sk_{\idt_1},x^{\mathsf{enc}}).
    \]
    If decryption fails, return \(0\).
    \item Decompose \(x = \sigma \| \ell \| \pi \).
    If \(x\) has the wrong format, return \(0\).
    If
    \[
        \versnark(\mathsf{pp_{snark}},
                  (M,\ell,t_\ell),\pi)=0
        \quad\text{or}\quad
        \mathsf{Verify}(\mathsf{vk},0,\sigma)=0,
    \]
    return \(0\).

    \item If the original input was encrypted, require \(y\)
    to have the prescribed ciphertext format and replace it by
    \(\Dec(\sk_{\idt_1},y)\).
    If its format is invalid or decryption fails, return \(1\).

    \item Decompose \(y = \sigma' \| \ell' \| \pi'\).
    If \(y\) has the wrong format, return \(1\).

    \item Verify that
    \[
        \sigma=\sigma',
        \qquad
        \ell' \geq \ell^+,
        \qquad
        \versnark(\mathsf{pp_{snark}},
                  (M,\ell',t_{\ell'}),\pi')=1.
    \]
    If all checks succeed, return \(0\),
    and otherwise return \(1\).
\end{enumerate}
Hence, \(h\) decrypts encrypted inputs and
invalid inputs incur no error. For a valid input at level \(\ell\),
an answer is correct if it preserves the input's signature \(\sigma\)
and contains a valid zk-SNARK proving knowledge of \(\ell'\) distinct
signatures, where \(\ell' \geq \ell^+\).

In Theorem~\ref{thm:mitdetgen} we show that \(\taskcon\) is learnable: the training phase of its Active Adversarial Defense satisfies Correctness, so it learns \(\taskcon\).

\subsubsection{The Separation for $\taskcon$}
Assume that strongly unforgeable signatures, publicly verifiable
zero-knowledge SNARKs with adaptive proof of knowledge, and an IB-FHE scheme with the admissibility property exist.

Every signature and every honest proof that a party in \(\taskcon\) sees
can be produced by the following oracle, which never discloses the
witnesses of the proofs it issues.

\begin{definition}[Pool Oracle]\label{def:pool-oracle}
Let \(s,\ell_{\max}:\N\to\N\), let \((\mathsf{vk},\sk)\) be a key pair of the
signature scheme, and let \(\mathsf{pp}_{\mathsf{snark}}\) be public parameters
of the SNARK. The oracle \(\mathcal{O}_{s,\ell_{\max}}\) uses \(\sk\) and
\(\mathsf{pp}_{\mathsf{snark}}\) and, for every
label \(\iota\), lazily samples an independent pool
\(\boldsymbol{\sigma}^{\iota}=(\sigma^{\iota}_1,\sigma^{\iota}_2,\dots)\) with
\(\sigma^{\iota}_i\gets\mathsf{Sign}(\sk,0)\). It answers two kinds of
queries.
\begin{itemize}[topsep=0pt, itemsep=0pt, parsep=0pt]
    \item \(\mathsf{sig}\): return a fresh \(\sigma\gets\mathsf{Sign}(\sk,0)\).
    At most \(s\) such queries are answered.
    \item \(\mathsf{prove}(\iota,\ell)\) with \(\ell\leq \ell_{\max}\): return
    \(\pi\gets\provsnark\bigl(\mathsf{pp}_{\mathsf{snark}},(M,\ell,t_\ell),
    (\sigma^{\iota}_1,\dots,\sigma^{\iota}_\ell)\bigr)\).
\end{itemize}
Here \(M\) is the machine of Section~\ref{sec:separation-samples} with \(\mathsf{vk}\) hard-wired.
\end{definition}

An example
\((x,y)\gets\dist_{\mathsf{Clear}}\) with \(x=\sigma\|\ell\|\pi\) and
\(y=\sigma\|\ell^+\|\pi'\) is one \(\mathsf{sig}\) query for \(\sigma\) and
two \(\mathsf{prove}\) queries at levels \(\ell\) and \(\ell^+\) under two
fresh labels.

\begin{lemma}[Pool Lemma]\label{lem:pool}
Assume strongly unforgeable signatures and zk-SNARKs with
adaptive proof of knowledge. Let \(s,\ell_{\max}\) be polynomially bounded.
For every polynomial-size oracle circuit family
\(\mathsf{Adv}\),
\[
\Pr\left[
    \!\begin{array}{l}
        \ell^*>\max\{s,\ell_{\max}\}, \\
        \versnark\bigl(\mathsf{pp}_{\mathsf{snark}},(M,\ell^*,t_{\ell^*}),\pi^*\bigr)=1
    \end{array}\!
    \GivenExperiment
    \StateExperiment{
        (\mathsf{vk},\sk)\gets\Gen(1^\prm) \\
        \mathsf{pp}_{\mathsf{snark}}\gets\gensnark(1^\prm,2^\prm) \\
        (\ell^*,\pi^*)\gets\mathsf{Adv}^{\mathcal{O}_{s,\ell_{\max}}}(\mathsf{vk},\mathsf{pp}_{\mathsf{snark}})
    }
\right]\leq\negl(\prm).
\]
\end{lemma}

\begin{proof}
Let \(Q=\poly(\prm)\) bound the number of
\(\mathsf{prove}\) queries of \(\mathsf{Adv}\), and fix a label \(0\)
that \(\mathsf{Adv}\) never uses. For \(0\leq j\leq Q\), let
\(\mathsf{Hyb}_j\) be the experiment of the lemma with two changes: the
\(\mathsf{sig}\) queries are answered by \(\sigma^{0}_1,\sigma^{0}_2,\dots\) in order, and
the first \(j\) \(\mathsf{prove}\) queries \((\iota,\ell)\) are answered
with the witness \((\sigma^{0}_1,\dots,\sigma^{0}_\ell)\). Let \(p_j\) be the probability that \(\mathsf{Adv}\) wins in \(\mathsf{Hyb}_j\). Since \(\boldsymbol{\sigma}^{0}\) is independent of all other answers, \(\mathsf{Hyb}_0\) merely renames fresh signatures, so \(p_0\) is the probability in the lemma and our goal is to bound \(p_0\).

First we argue that \emph{\(|p_j-p_{j-1}|\leq\negl(\prm)\).} Except with negligible
probability all pool entries are pairwise distinct, so for the \(j\)-th
query \((\iota,\ell)\) both \(w_0=(\sigma^{\iota}_1,\dots,\sigma^{\iota}_\ell)\)
and \(w_1=(\sigma^{0}_1,\dots,\sigma^{0}_\ell)\) are witnesses for
\((M,\ell,t_\ell)\). The distinguisher of
Definition~\ref{def:zk-snark} samples \((\mathsf{vk},\sk)\) and all pools,
runs \(\mathsf{Hyb}_{j-1}\) with \(\mathsf{Adv}\), answering every query
itself except the \(j\)-th \(\mathsf{prove}\) query, for which it outputs
\((M,\ell,t_\ell)\) with \(w_0\), respectively \(w_1\), and forwards the
proof it receives, and accepts if \(\mathsf{Adv}\) wins. By zero
knowledge, both \(p_{j-1}\) and \(p_j\) are within
\(\negl(\prm)\) of the probability that \(\mathsf{Adv}\) wins when the \(j\)-th proof is produced by the simulator. That probability does not depend on the witness, since the simulator never sees it. 

Now we argue that \emph{\(p_Q\leq\negl(\prm)\).} In \(\mathsf{Hyb}_Q\) every answer is
computed from \(\mathsf{vk}\), \(\mathsf{pp}_{\mathsf{snark}}\) and
\(\sigma^{0}_1,\dots,\sigma^{0}_{\max\{s,\ell_{\max}\}}\). Let
\(z\) consist of \(\mathsf{vk}\), \(\sigma^{0}_1,\dots,\sigma^{0}_{\max\{s,\ell_{\max}\}}\) and the coins of \(\mathsf{Adv}\) and of the proofs that answer its \(\mathsf{prove}\) queries in \(\mathsf{Hyb}_Q\), and let
\(\provsnark^*\) be the deterministic prover that, on \(\mathsf{pp}_{\mathsf{snark}}\) and auxiliary input \(z\), runs
\(\mathsf{Hyb}_Q\) with \(\mathsf{Adv}\) on these coins and outputs
\(((M,\ell^*,t_{\ell^*}),\pi^*)\), and let \(\mathcal{E}\) be its
extractor. A forger obtains the signatures in \(z\) from \(\max\{s,\ell_{\max}\}\) signing queries, samples
the coins in \(z\) and \(\mathsf{pp}_{\mathsf{snark}}\gets\gensnark(1^\prm,2^\prm)\), runs
\(\provsnark^*\) and \(\mathcal{E}\) on \((\mathsf{pp}_{\mathsf{snark}},z)\), and outputs
an entry of the extracted \(w\) not among its queries. When
\(\mathsf{Adv}\) wins and extraction succeeds, \(w\) consists of
\(\ell^*>\max\{s,\ell_{\max}\}\) pairwise distinct valid signatures, so such an entry exists
and is a strong forgery. 

Hence \(p_Q\leq\negl(\prm)\), and
\(p_0\leq p_Q+Q\cdot\negl(\prm)=\negl(\prm)\).
\end{proof}
\begin{theorem}[Sample-Complexity Separation]
\label{thm:mitdetgen}
There exists
a learning task \(\task\) such that
for every distinguishability parameter
\(c_{\AAD} : \N \to (0,1)\) and every sample complexity \(K : \N \to \N\) bounded by a
polynomial in \(\prm\) with \(K(\prm)\geq\prm\) the following holds.
\begin{enumerate}[label=(\roman*), leftmargin=2em, topsep=0pt, itemsep=0pt, parsep=0pt]
    \item There is an Active Adversarial Defense \((\provtrain, \provfix)\)
    with sample complexities \(\ktrain = K\) and \(\kfix = O(\sqrt{K})\), such that
    \[
    (\provtrain, \provfix) \in_{1} \AAD\Big(\task, 1, \poly(\prm), \poly(\prm), \poly(\prm), \negl(\prm), \negl(\prm), c_{\AAD}, \negl(\prm)\Big),
    \]
    where the attackers are restricted to sample
    complexity \(K_\ver \leq K/4\).
    \item For every \(\prov\) of polynomial size \(S_\prov\) with sample
    complexity \(K_\prov \leq K\) whose model class has evaluation circuits
    of size \(S_E\),
    \[
    \prov \notin_{\varphi} \AD\Big(\task, 1, O\big(\sqrt{K}\,\log^2\prm\,(S_E + \poly(\prm))\big), S_\prov, \varepsilon, \delta, \tfrac{1}{p(\prm)}, \tfrac14 - \tfrac{1}{p(\prm)}\Big),
    \]
    for any polynomials \(p > 4\), \(\varepsilon, \delta : \N \to (0,1)\) and \(\varphi > 0\),
    witnessed by an attacker with \(K_\ver = O(\sqrt{K}\,\log^2\prm)\).
\end{enumerate}
\end{theorem}

\begin{proof}
The task is \(\task = \taskcon\). 
\paragraph{The Attack.}
Let \(m=10\sqrt K\). Upon receiving \(z_f\) and \(z_R\), the attacker ignores
\(z_R\), evaluates \(f\) with the evaluation circuit of its class,
and proceeds as follows. 

\begin{enumerate}[topsep=0pt, itemsep=0pt, parsep=0pt]
    \item Draw \(\log^2\prm\) samples from \(\dist\) and let
    \(\sigma\) be a signature contained in one unencrypted sample.
    If no such sample exists, return a sample from
    \(\dist_{\inspace}\).

    \item Compute
    \[
        \pi_1\gets\provsnark
        \bigl(\mathsf{pp}_{\mathsf{snark}},(M,1,t_1),\sigma\bigr),
    \]
    and set \(\ell_1:=1\) and
    \(x_1:=\sigma\|\ell_1\|\pi_1\).

    \item Draw \(4(2m\log^2\prm+2)\) samples from \(\dist\). If fewer
    than \(2m\log^2\prm+2\) of them are encrypted, return a fresh input
    from \(\dist_{\inspace}\). Otherwise take the identity--key pairs
    from the auxiliary fields of the first \(2m\log^2\prm+2\) encrypted
    samples. If two of these identities coincide, return a fresh input
    from \(\dist_{\inspace}\).

    \item For \(j=1,\dots,m\), perform \(\log^2\prm\) trials, each as follows:
    \begin{itemize}[topsep=0pt, itemsep=0pt, parsep=0pt]
        \item Take the next two unused pairs
        \((\idt_1,\sk_{\idt_1})\) and
        \((\idt_2,\sk_{\idt_2})\).

        \item Let
        \[
            x':=\Enc(\mathsf{pp}_{\mathsf{fhe}},\idt_1,x_j)\|\idt_1\|
                \idt_2\|\sk_{\idt_2},
            \qquad
            y_j:=\Dec(\sk_{\idt_1},f(x')).
        \]
        Decompose \(y_j=\sigma'\|\ell'\|\pi'\).
        The trial succeeds if \(y_j\) has the right format,
        \[
            \sigma'=\sigma,\qquad
            \ell'\geq \ell_j^+,
            \qquad
            \versnark\bigl(
                \mathsf{pp}_{\mathsf{snark}},
                (M,\ell',t_{\ell'}),\pi'
            \bigr)=1.
        \]
        Decryption failures count as failed trials.
    \end{itemize}
    If any trial succeeds, let \(x_{j+1}\) be the
    output \(y_j\) of the first successful trial, and let
    \(\ell_{j+1}\) be its level. If all trials fail:
    \begin{itemize}[topsep=0pt, itemsep=0pt, parsep=0pt]
        \item With probability \(\frac12\), take the
        last two pairs and return
        \[
            \Enc(\mathsf{pp}_{\mathsf{fhe}},\idt_1,x_j)\|\idt_1\|
            \idt_2\|\sk_{\idt_2}.
        \]
        \item Otherwise, draw \(\log^2\prm\) samples from \(\dist\) and
        return the input of the first unencrypted one, or a fresh input
        from \(\dist_{\inspace}\) if there is none.
    \end{itemize}

    \item Return a fresh input from \(\dist_{\inspace}\).
\end{enumerate}

\paragraph{Analysis of the Attack.}
Training and the attack together use at most
\[N=K+4(2m\log^2\prm+2)+2\log^2\prm\leq 2K\]
examples, where the inequality holds for all sufficiently large
\(\prm\).
For \(\ell\leq N\) we have \(\ell^+\leq 2N\), so an example \((x,y)\)
contains a proof of knowledge of more than \(2N\) signatures only if its level exceeds \(N\). Such an example exists with probability at most \(N\cdot2^{-N}\), which is negligible in \(\prm\). We condition on this not happening, so that every example can be
produced through the oracle \(\mathcal{O}_{N,2N}\).

Let \(\mathsf{Adv}\) be a circuit that simulates the training and the attack having oracle access to
\(\mathcal{O}_{N,2N}\).
\begin{enumerate}[topsep=0pt, itemsep=0pt, parsep=0pt]
    \item Run \((\mathsf{pp}_{\mathsf{fhe}},\mathsf{MSK})\gets\mathsf{Setup}(1^\prm)\).
    \item Answer every query of training and of the attack to \(\ex\)
    by sampling \(\ell\gets\mathsf{Geom}(\frac12)\) conditioned on
    \(\ell\leq N\) and making queries to \(\mathcal{O}_{N,2N}\). Sample \(\idt_1,\idt_2\gets\mathcal{I}\) and
    \(\sk_{\idt_2}\gets\mathsf{KeyGen}(\mathsf{MSK},\idt_2)\) to generate encrypted examples.
    \item Run the training of \(\prov\), which publishes \(f\), and
    then the attack on \(f\).
    \item If all \(m\) rounds succeed, output \((\ell_{m+1},\pi_{m+1})\),
    where \(x_{m+1}=\sigma\|\ell_{m+1}\|\pi_{m+1}\).
\end{enumerate}
Lemma~\ref{lem:pool} rules out
\(\ell_{m+1}>2N\), except with negligible probability. 
On the other hand, if all \(m\) rounds succeeded, then
\(\ell_{j+1}\geq \ell_j^+\) would give
\(\ell_{m+1}\geq(1+m/5)^2>4K\geq 2N\).
Thus, some round has no successful trial, except with negligible
probability.

Fix a round \(j\), and fix the task instance, the outcome of
training and the history of the attack before round \(j\). This fixes
\(f\) and \(x_j=\sigma\|\ell_j\|\pi_j\), and the probability that a trial on a
fresh encryption of \(x_j\) succeeds is the number
\[
p_j=\Pr\left[
    \!\begin{array}{l}
        y=\sigma\|\ell'\|\pi' \text{ with } \ell'\geq \ell_j^+, \\
        \versnark\bigl(\mathsf{pp}_{\mathsf{snark}},(M,\ell',t_{\ell'}),\pi'\bigr)=1
    \end{array}\!
    \GivenExperiment
    \StateExperiment{
        \idt_1,\idt_2\gets\mathcal{I} \\
        \sk_{\idt_1}\gets\mathsf{KeyGen}(\mathsf{MSK},\idt_1) \\
        \sk_{\idt_2}\gets\mathsf{KeyGen}(\mathsf{MSK},\idt_2) \\
        x'\gets\Enc(\mathsf{pp}_{\mathsf{fhe}},\idt_1,x_j)\|\idt_1\|\idt_2\|\sk_{\idt_2} \\
        y\gets\Dec(\sk_{\idt_1},f(x'))
    }
\right].
\]
Up to a negligible collision cost, the distinct identities of the
trials are independent fresh identities, so the \(\log^2\prm\) trials of
round \(j\) are independent, and if \(p_j>1/2\) they all fail with
probability at most \(2^{-\log^2\prm}\). By a union bound over the \(m\)
rounds, except with probability \(m2^{-\log^2\prm}=\negl(\prm)\), the
round in which all trials fail has \(p_j\leq 1/2\). On this event,  with probability \(\frac12\) the attack returns
\[
x_\ver=\Enc(\mathsf{pp}_{\mathsf{fhe}},\idt_1,x_j)\|\idt_1\|\idt_2\|\sk_{\idt_2}
\]
for the last two pairs, which no trial used. Given the history,
\(x_\ver\) is distributed as \(x'\) in the experiment of \(p_j\), so the
event of \(p_j\) fails for the answer \(f(x_\ver)\) with probability
\(1-p_j\geq 1/2\). Since \(x_j\) is a valid question, on an admissible
encryption of \(x_j\) the oracle \(h\) checks exactly this event and
returns \(1\) when it fails, and \(x_\ver\) is admissible except with
negligible probability. Hence, the expected error is
\[
\mathbb{E}\left[h(x_\ver,f(x_\ver))
\GivenExperiment
\StateExperiment{
    (\dist,h)\gets\task_\prm \\
    (z_f,z_R,\priv)\gets\prov^{\ex} \\
    x_\ver\gets\ver_\prm^{\ex}(z_f,z_R)
}\right]
\geq\bigl(1-\negl(\prm)\bigr)\cdot\tfrac12\bigl(\tfrac12-\negl(\prm)\bigr)
\geq\tfrac14-\negl(\prm).
\]

It remains to show that the expected absolute flag-rate gap
\begin{multline*}
\mathbb{E}\Bigl[\Bigl|
\Pr\left[R(\priv,x_\ver)=1\GivenExperiment x_\ver\gets\ver_\prm^{\ex}(z_f,z_R)\right]
-\Pr\left[R(\priv,x)=1\GivenExperiment x\gets\dist_{\inspace}\right]
\Bigr| \\
\;\Big\vert\; (\dist,h)\gets\task_\prm,\; (z_f,z_R,\priv)\gets\prov^{\ex}\Bigr],
\end{multline*}
is negligible. Fix the task instance and \((z_f,z_R,\priv)\). A
natural input is clear with probability \(\frac12\) and otherwise
\(\Enc(\mathsf{pp}_{\mathsf{fhe}},\idt_1,x)\|\idt_1\|\idt_2\|\sk_{\idt_2}\)
for a fresh natural clear question \(x\), and \(x_\ver\) has the same
distribution with \(x_j\) in place of \(x\) if a round fails. The following
IND-sID-CPA adversary shows indistinguishability. It
chooses \(\idt^*\gets\mathcal{I}\) before setup, simulates the task, the
training of \(\prov\) and the attack with \(\idt_1=\idt^*\) in \(x_\ver\)
and with \(\mathsf{KeyGen}(\mathsf{MSK},\idt)\), \(\idt\neq\idt^*\), from
the challenger, submits \(x_j\) and \(x\), and outputs \(R(\priv,x_\ver)\)
with the challenge ciphertext in place of
\(\Enc(\mathsf{pp}_{\mathsf{fhe}},\idt^*,x_j)\), flipped if the inner
difference, estimated from \(\poly(\prm)\) runs with the same
\((z_f,z_R,\priv)\), is negative. It never needs \(\sk_{\idt^*}\), and it
aborts if \(\idt^*\) occurs elsewhere, which has probability
\(\poly(\prm)\,2^{-\Omega(\prm)}\). So a non-negligible gap gives a non-negligible advantage, at least half the gap minus \(\negl(\prm)\).

Fix a polynomial \(p > 4\). For all sufficiently large
\(\prm\) the expected error of the attack exceeds \(\tfrac14 - \tfrac1p\)
and the gap is below \(\tfrac1p\), so the attacker violates both
Robustness with \(s_{\AD} = \tfrac14 - \tfrac1p\) and
Distinguishability with \(c_{\AD} = \tfrac1p\), whatever
\(\varepsilon\) and \(\delta\) are. The AD conditions therefore
hold at finitely many \(\prm\) only. Thus,
\(\prov\) is not a successful AD. 

The attack uses \(O(m\log^2\prm)=O(\sqrt K\log^2\prm)\)
samples and has size \(O(m\log^2\prm\,(S_E+\poly(\prm)))\).
\paragraph{The Active Adversarial Defense.}
\(\provtrain\) runs the following algorithm:
\begin{enumerate}[topsep=0pt, itemsep=0pt, parsep=0pt]
    \item Draw \(K\) samples and let \((\sigma_i)_{i \in [K/4]}\) be the signatures of the first \(K/4\)
    clear examples. If there are fewer than \(K/4\) clear
    examples, or two of the \(\sigma_i\) coincide, publish the constant
    model \(y_\bot\), set \(\priv=\bot\), and stop.
    \item For every \(\ell\in \{\sqrt K/2, \sqrt K, \dots, K/4\}\) set
    \[\pi_\ell\gets\provsnark\bigl(\mathsf{pp}_{\mathsf{snark}},(M,\ell,t_\ell),
    (\sigma_i)_{i \in [\ell]}\bigr).\]
    \item Let \(f'\) be the algorithm that on input
    \(x=\sigma\|\ell\|\pi\) returns \(\sigma\|\ell'\|\pi_{\ell'}\) for the smallest
    \(\ell'\) with \(\ell'\geq \ell^+\), and \(y_\bot\) if
    there is no such \(\ell'\) or \(x\) is malformed.
    \item Let \(f\) be the algorithm that on a clear input returns
    \(f'(x)\), and on an encrypted input
    \(x=x^{\mathsf{enc}}\|\idt_1\|\idt_2\|\sk_{\idt_2}\) returns
    \(\Eval(\mathsf{pp}_{\mathsf{fhe}},f',x^{\mathsf{enc}})\).
    \item Publish \(z_f\) and set \(\priv=(\sigma_i)_{i \in [K/4]}\).
\end{enumerate}
\(\provfix\) receives \(\priv\) and \(x\in\inspace\) and runs the
following algorithm:
\begin{enumerate}[topsep=0pt, itemsep=0pt, parsep=0pt]
    \item Draw \(4\sqrt K\) samples and let \((\sigma^{\mathsf{Inference}}_i)_{i \in [\sqrt K/2]}\) be the signatures of
    the first \(\sqrt K/2\) clear examples. If
    \(\priv=\bot\), a signature repeats in
    \((\sigma_i)_{i \in [K/4]}\cup(\sigma^{\mathsf{Inference}}_i)_{i \in [\sqrt K/2]}\),
    or there are fewer than \(\sqrt K/2\) clear examples, return \((y_\bot,r=0)\).
    \item Compute \(\pi'\gets\provsnark\bigl(\mathsf{pp}_{\mathsf{snark}},(M,K/4+\sqrt K/2,t_{K/4+\sqrt K/2}),
    (\sigma_i)_{i \in [K/4]}\cup(\sigma^{\mathsf{Inference}}_i)_{i \in [\sqrt K/2]}\bigr)\).
    \item Let \(f^+\) be the algorithm that on input
    \(x=\sigma\|\ell\|\pi\) with \(1\leq \ell\leq K/4\) returns
    \(\sigma\|(K/4+\sqrt K/2)\|\pi'\), and \(y_\bot\) otherwise.
    \item Return \((f^+(x),r=0)\) on a clear input, and
    \((\Eval(\mathsf{pp}_{\mathsf{fhe}},f^+,x^{\mathsf{enc}}),r=0)\) on
    an encrypted input.
\end{enumerate}
Here \(\provfix\) collects \(\sqrt K/2\) fresh signatures, joins them with the
\(K/4\) retained ones into a single proof \(\pi'\) at level \(K/4+\sqrt K/2\), and
answers every valid question of level at most \(K/4\) with it. Training
also publishes the description \(z_R\) of the rule \(R \equiv 0\), so the defense never flags an input.

\paragraph{Correctness and Completeness.}
By the Chernoff bound, training
and inference return
\(y_\bot\) with negligible probability.
Otherwise, let \(x=\sigma\|\ell\|\pi\) be a valid clear question. If
\(\ell\leq K/8\), then \(\ell^+\leq K/4\), so \(f'(x)=\sigma\|\ell'\|\pi_{\ell'}\)
for the least \(\ell'\in\{\sqrt K/2, \sqrt K, \dots, K/4\}\) with
\(\ell'\geq \ell^+\), and \(h(x,f'(x))=0\) by completeness of the SNARK. If
\(\ell\leq K/4\), then \(\ell^+\leq K/4+\sqrt K/2\), so \(h(x,f^+(x))=0\). An
encrypted natural question is inadmissible, and then \(h=0\), or \(h\) decrypts
to a valid clear question \(x\), and then
\(\Eval(\mathsf{pp}_{\mathsf{fhe}},f',\cdot)\) and
\(\Eval(\mathsf{pp}_{\mathsf{fhe}},f^+,\cdot)\) decrypt to \(f'(x)\) and
\(f^+(x)\) by perfect correctness. Since \(\Pr[\ell>K/8]=2^{-K/8}\) and
\(\Pr[\ell>K/4]=2^{-K/4}\) for \(\ell\gets\mathsf{Geom}(\frac12)\), and
\(\err_{x,h}(y)=h(x,y)\) for batch size \(1\), we obtain
\[
\Pr\left[\errf>2^{-K/8}\GivenExperiment\StateExperiment{
    (\dist,h)\gets\task_\prm \\
    z_f\gets\provtrain^{\ex}}\right]\leq\negl(\prm),
\]
\[
\Pr\left[\err_{x,h}(y)=1\GivenExperiment\StateExperiment{
    (\dist,h)\gets\task_\prm \\
    (z_f,z_R,\priv)\gets\provtrain^{\ex} \\
    x\gets\dist_{\inspace} \\
    (y,r)\gets\provfix^{\ex}(\priv,x)}\right]
\leq \negl(\prm).
\]
As \(K\geq\prm\) gives \(2^{-K/8}\leq 2^{-\prm/8}\), and
\(\err_{x,h}(y)\in\{0,1\}\), these bounds give Correctness and
Completeness with negligible \(\varepsilon\) and \(\delta\).

\paragraph{Robustness.}
Fix an attacker \(\ver\) of polynomial size with \(K_\ver\leq K/4\).
The published model contains only proofs at levels at
most \(K/4\), computed on prefixes of the retained signatures, and the
examples of \(\ver\) contain a proof above \(K/4\) with
probability at most \(K/4\cdot 2^{-K/8}=\negl(\prm)\), since
\(\ell^+ \leq 2\ell\). We condition on this not happening.

By the analysis of the attack, with \(\mathcal{O}_{K/4,K/4}\) in place of
\(\mathcal{O}_{N,2N}\), the question of \(x_\ver\), decrypted if \(x_\ver\) is encrypted, is valid with level above \(K/4\) only with negligible
probability. Here the circuit of Lemma~\ref{lem:pool} obtains the
published proofs as \(\mathsf{prove}(\iota_{\mathsf{Train}},\ell)\) under one
label \(\iota_{\mathsf{Train}}\), since \(f\) depends on training only
through the retained \(K/4\) fresh signatures, so it makes at most \(K/4\)
\(\mathsf{sig}\) queries and only \(\mathsf{prove}\) queries of level at
most \(K/4\).

If \(x_\ver\) is encrypted and inadmissible, or its question is
invalid, then \(h(x_\ver,\cdot)=0\). Otherwise its question has level at
most \(K/4\), and the argument of Correctness and Completeness gives
\(h(x_\ver,y)=0\) unless training or inference returns \(y_\bot\), with the
admissibility in place of perfect correctness
when \(x_\ver\) is encrypted. Hence,
\[
\mathbb{E}\left[h(x_\ver,y)\GivenExperiment\StateExperiment{
    (\dist,h)\gets\task_\prm \\
    (z_f,z_R,\priv)\gets\provtrain^{\ex} \\
    x_\ver\gets\ver_\prm^{\ex}(z_f,z_R) \\
    (y,r)\gets\provfix^{\ex}(\priv,x_\ver)}\right]
\leq\negl(\prm).
\]
The negligible function is the same for all attackers of the same
size, since for each \(\prm\) the maximal expected error is attained by one
circuit and Lemma~\ref{lem:pool} applies to that family. As Correctness,
Completeness and Robustness hold for all sufficiently large \(\prm\), the
defense has frequency \(1\).

\paragraph{Efficiency.}
Training uses \(K\) samples and inference uses \(4\sqrt K\)
fresh samples. The model contains \(\sqrt K/2\) succinct proofs,
so its description length \(m_{\fclass}\) is \(O(\sqrt K\,\poly(\prm))\), and so is the size \(S_E\) of its evaluation circuits, since on an encrypted input \(\Eval\) adds only a factor \(p_{\Eval}(\prm)\) to the size of \(f'\).
All computations have polynomial running time.
\end{proof}

\subsection{Separation in Running Time}
\label{sec:separation-time}

\subsubsection{\texorpdfstring{The Learning Task \(\tasktime\)}{The Learning Task L\textasciicircum time}}
Let \(\mathsf{NPL} = (\mathsf{Samp}, M)\) be an average-case NPL ensemble with gap \(\eta \in (\frac12, 1)\), and let \(S\) be a size parameter bounded by a polynomial in \(\prm\). We use the ensemble at the difficulty \(\tau = \prm \cdot S^{1/\eta}\), and since \(\tau\) is fixed throughout this section, we write \(\mathsf{Samp}(1^\prm)\) for \(\mathsf{Samp}(1^\prm,\tau)\) and \(M(\chi)\) for \(M(\tau,\chi)\). 

Let \(\mathsf{IVC} = (\mathsf{G}, \mathsf{P}, \mathsf{V}, \mathsf{Update})\) be an IVC scheme for the machine \(M\), and write \(M^{(\ell)}(\chi)\) for its configuration on input \(\chi\) after \(\ell\) steps. Throughout, the prover \(\mathsf{P}(\mathsf{pp}_{\mathsf{ivc}}, (\chi,\ell,c))\) is implemented as \(\pi_0 = \bot\) followed by \(\ell\) calls to \(\mathsf{Update}\) along the computation, with fresh randomness for every proof, so that the proofs of the example distribution and those produced by the parties and reductions below have the same distribution.

For \(\prm  \in \N\) we define the learning task \(\tasktime_\prm\) as the result of the following procedure:
\[\tasktime_\prm =
 \left\{
(\dist_{\mathsf{pp}_{\mathsf{fhe}},\mathsf{MSK},\mathsf{pp}_{\mathsf{ivc}},\chi},h_{\mathsf{pp}_{\mathsf{fhe}},\mathsf{MSK},\mathsf{pp}_{\mathsf{ivc}},\chi})
\GivenExperiment
\StateExperiment{
\mathsf{pp}_{\mathsf{ivc}} \gets \mathsf{G}(1^\prm) \\
(\mathsf{pp}_{\mathsf{fhe}}, \mathsf{MSK}) \gets \mathsf{Setup}(1^\prm) \\
\chi \gets \mathsf{Samp}(1^\prm)
}
\right\}.
\]

As before, we omit the subscripts of \(\dist\) and \(\dist\) will have two equally probable parts \(\dist_\mathsf{Clear}\) and \(\dist_\mathsf{Enc}\).
We set
\[
\dist_\mathsf{Clear} = \left\{
(c \| \ell \| \pi, c' \| \ell^+ \|  \pi')
\GivenExperiment
\StateExperiment{
\ell \gets \mathsf{Geom}(\frac12) \text{ conditioned on } \ell \leq 2\tau \\
c \gets M^{(\ell)}(\chi) \\
c' \gets M^{(\ell^+)}(\chi) \\
\pi \gets \mathsf{P}(\mathsf{pp}_{\mathsf{ivc}},(\chi,\ell,c)) \\
\pi' \gets \mathsf{P}(\mathsf{pp}_{\mathsf{ivc}},(\chi,\ell^+,c')) \text{ with independent randomness}
}
\right\}.
\]
Conditioning \(\ell \leq 2\tau\) changes the distribution by \(2^{-2\tau}\), which is negligible, and we ignore this term below.

The distribution \(\dist_\mathsf{Enc}\) is obtained from \(\dist_\mathsf{Clear}\) exactly as in Section~\ref{sec:separation-samples}.

Oracle \(h\) is defined via the following algorithm. 
\begin{enumerate}[topsep=0pt, itemsep=0pt, parsep=0pt]
    \item Determine whether \(x\) has the clear or
    encrypted input format. If neither, return \(0\).
    \item If \(x\) is encrypted, parse
    \[
        x = x^{\mathsf{enc}} \| \idt_1 \| \idt_2
            \| \sk_{\idt_2}.
    \]
    If \(\mathsf{Adm}(\mathsf{MSK},\idt_1,x^{\mathsf{enc}})=0\),
    return \(0\).
    Otherwise, retrieve the fixed key \(\sk_{\idt_1}\) and replace \(x\) by
    \(\Dec(\sk_{\idt_1},x^{\mathsf{enc}})\).
    If decryption fails, return \(0\).

    \item Decompose \(x = c \| \ell \| \pi\).
    If \(x\) has the wrong format or \(\ell > 2\tau\), return \(0\).
    If
    \[
        \mathsf{V}(\mathsf{pp}_{\mathsf{ivc}},
                   (\chi,\ell,c),\pi)=0,
    \]
    return \(0\).

    \item If the original input was encrypted, require \(y\)
    to have the prescribed ciphertext format and replace it by
    \(\Dec(\sk_{\idt_1},y)\).
    If its format is invalid or decryption fails, return \(1\).

    \item Decompose \(y = c' \| \ell' \| \pi'\).
    If \(y\) has the wrong format, return \(1\).

    \item Verify that
    \[
        \ell^+ \leq \ell' \leq 4\tau,
        \qquad
        \mathsf{V}(\mathsf{pp}_{\mathsf{ivc}},
                   (\chi,\ell',c'),\pi')=1.
    \]
    If all checks succeed, return \(0\),
    and otherwise return \(1\).
\end{enumerate}
In other words, \(h\) decrypts encrypted inputs such that invalid inputs incur no error. For a valid input at level \(\ell\),
an answer is correct if it contains a valid IVC proof for a
configuration at some level \(\ell'\) with \(\ell^+ \leq \ell' \leq 4\tau\).

In Theorem~\ref{thm:mitdetgentime} we show that \(\tasktime\) is learnable: the training phase of its Active Adversarial Defense satisfies Correctness, so it learns \(\tasktime\).

\subsubsection{The Separation for $\tasktime$}
Let \(\eta \in (\tfrac12, 1)\), and assume that an average-case non-parallelizing language ensemble with gap \(\eta\), an IVC scheme and an IB-FHE scheme with the admissibility property exist.

Let \(g(\prm)\geq\prm\) be a polynomial that bounds the length of the
inputs and answers of \(\tasktime\) and the size of each of the following
circuits: one step of \(M\) together with \(\mathsf{Update}\), \(\mathsf{G}\),
\(\mathsf{Setup}\), \(\mathsf{KeyGen}\), \(\Enc\), \(\Dec\), \(\mathsf{V}\),
sampling the level of an example, comparing two levels or identities, and
storing a certified configuration together with the test, in the clear or
under \(\Eval\), whether it is a valid answer for an input.

The following lemma is the only place where the hardness of the non-parallelizing language enters.

\begin{lemma}[Hardness of Reaching the Terminal Level]
\label{lem:terminal-level-hardness}
Let \(S\) be polynomial in \(\prm\) and \(\tasktime\) be the task of Section~\ref{sec:separation-time}, with
difficulty \(\tau=\prm\cdot S^{1/\eta}\). Let \(\ver\) be a circuit of size
at most \(\tau^\eta-2\prm\, g(\prm)\) with access to \(\ex\), where \(\dist\) is
the distribution of the task instance. Then
\[
\Pr\left[
    \!\begin{array}{l}
        \tau\leq \ell\leq 4\tau, \\
        \mathsf{V}(\mathsf{pp}_{\mathsf{ivc}},(\chi,\ell,c),\pi)=1
    \end{array}\!
    \GivenExperiment
    \StateExperiment{
        \mathsf{pp}_{\mathsf{ivc}}\gets\mathsf{G}(1^\prm) \\
        (\mathsf{pp}_{\mathsf{fhe}},\mathsf{MSK})\gets\mathsf{Setup}(1^\prm) \\
        \chi\gets\mathsf{Samp}(1^\prm) \\
        (\ell,c,\pi)\gets\ver^{\ex}(\chi)
    }
\right]\leq\negl(\prm),
\]
and the same holds if \(\ver\) outputs
\(\Enc(\mathsf{pp}_{\mathsf{fhe}},\idt,c\|\ell\|\pi)\) for an identity \(\idt\) in
place of \((\ell,c,\pi)\).
\end{lemma}

\begin{proof}
Let \(p\) be the probability in the lemma, and let \(\mathsf{Adv}\) be
the circuit that on input \(\chi\)
\begin{enumerate}[topsep=0pt, itemsep=0pt, parsep=0pt]
    \item Samples \(\mathsf{pp}_{\mathsf{ivc}}\gets\mathsf{G}(1^\prm)\) and
    \((\mathsf{pp}_{\mathsf{fhe}},\mathsf{MSK})\gets\mathsf{Setup}(1^\prm)\).
    \item Answers all queries of \(\ver\) to \(\ex\) in parallel, as in
    \(\dist_{\mathsf{Clear}}\) and \(\dist_{\mathsf{Enc}}\) but with levels
    truncated at \(\prm\), so that each example takes at most
    \(\prm^+\) steps of \(M\) and of \(\mathsf{Update}\).
    \item Runs \(\ver\), decrypts its output with \(\mathsf{MSK}\) if it
    is encrypted, and checks \(\tau\leq \ell\leq 4\tau\) and
    \(\mathsf{V}(\mathsf{pp}_{\mathsf{ivc}},(\chi,\ell,c),\pi)=1\). If both hold, it
    returns \(1\) if \(c\) is an accepting configuration and \(0\) otherwise.
    Else it returns a uniform bit.
\end{enumerate}
Truncation changes the view of \(\ver\) by at most
\(\poly(\prm)\,2^{-\prm}\). Step 1 has depth at most \(2g(\prm)\),
step 2 at most \((\prm^+ +2)\,g(\prm)\), since the two proof chains
of an example run in parallel, and the checks of step 3 at most
\(5g(\prm)\). So for \(\prm\geq 16\), \(\mathsf{Adv}\) has depth at most
\(|\ver|+(\prm^+ +9)\,g(\prm)\leq|\ver|+2\prm\, g(\prm)\leq\tau^\eta\).

In the experiment of the lemma,
\[
\Pr\left[
    \!\begin{array}{l}
        \tau\leq \ell\leq 4\tau,\ c\neq M^{(\ell)}(\chi), \\
        \mathsf{V}(\mathsf{pp}_{\mathsf{ivc}},(\chi,\ell,c),\pi)=1
    \end{array}\!
\right]
\leq 4\tau\cdot\max_{\tau\leq \ell^*\leq 4\tau}
\Pr\left[
    \!\begin{array}{l}
        \ell=\ell^*,\ c\neq M^{(\ell)}(\chi), \\
        \mathsf{V}(\mathsf{pp}_{\mathsf{ivc}},(\chi,\ell,c),\pi)=1
    \end{array}\!
\right].
\]
For the maximizing \(\ell^*\), the prover that runs this experiment and
outputs \(((\chi,\ell,c),\pi)\) only if \(\ell=\ell^*\) breaks the soundness of IVC
with the maximum, so the maximum is negligible, and so is the left side.
So, except with negligible probability, \(c=M^{(\ell)}(\chi)\), which for
\(\ell\geq\tau\) is halted, and accepting if and only if \(M(\chi)=1\). Hence,
\[
\Pr\left[\mathsf{Adv}(\chi)=M(\chi)\GivenExperiment\chi\gets\mathsf{Samp}(1^\prm)\right]
\geq\tfrac12+\tfrac p2-\negl(\prm),
\]
and hence \(p\leq\negl(\prm)\).
\end{proof}

\begin{theorem}[Running-Time Separation]\label{thm:mitdetgentime}
For every \(S : \N \to \N\) bounded by a polynomial in \(\prm\) with \(S(\prm)\geq 4\prm\, g(\prm)\) there exists a learning task \(\task\) such that for every distinguishability parameter \(c_{\AAD} : \N \to (0,1)\), every constant \(a > 0\), and every \(S_E : \N \to \N\) with
\[2g(\prm) \leq S_E \leq S^{1-\frac{1}{2\eta}}/\bigl(g(\prm)\log^2\prm\bigr)\]
the following holds.
\begin{enumerate}[label=(\roman*), leftmargin=2em, topsep=0pt, itemsep=0pt, parsep=0pt]
    \item There is an Active Adversarial Defense \((\provtrain, \provfix)\), whose published model belongs to a class with evaluation circuits of size \(S_E\), such that
    \[
    (\provtrain, \provfix) \in_{1} \AAD\Big(\task, 1, aS, S, \poly(\prm)\cdot S^{\frac{1}{2\eta}}, \negl(\prm), \negl(\prm), c_{\AAD}, \negl(\prm)\Big).
    \]
    \item For every \(\prov\) of size \(S\) whose model class has evaluation circuits of size \(S_E\),
    \[
    \prov \notin_{\varphi} \AD\Big(\task, 1, \poly(\prm)\cdot S^{\frac{1}{2\eta}} \cdot S_E, S, \varepsilon, \delta, \tfrac{1}{p(\prm)}, \tfrac14 - \tfrac{1}{p(\prm)}\Big),
    \]
    for all polynomials \(p > 4\), \(\varepsilon, \delta : \N \to (0,1)\) and \(\varphi > 0\).
\end{enumerate}
\end{theorem}

\begin{proof}
The task is \(\task=\tasktime\). As we discussed in Section~\ref{sec:overview}, the proof is very similar to that of Theorem~\ref{thm:mitdetgen}.

\paragraph{The Attack.}
Let \(m=5\sqrt{\tau}\). \(\ver\) performs the following
\begin{enumerate}[topsep=0pt, itemsep=0pt, parsep=0pt]
    \item Draw \(\log^2\prm\) samples from \(\dist\) and let
    \(x_1=c_1\|\ell_1\|\pi_1\) be the input of one unencrypted sample. If no
    such sample exists, return a sample from \(\dist_{\inspace}\).
    \item Draw \(4(2m\log^2\prm+2)\) samples from \(\dist\). If fewer
    than \(2m\log^2\prm+2\) of them are encrypted, return a fresh input
    from \(\dist_{\inspace}\). Otherwise take the identity--key pairs
    from the auxiliary fields of the first \(2m\log^2\prm+2\) encrypted
    samples. If two of these identities coincide, return a fresh input
    from \(\dist_{\inspace}\).

    \item For \(j=1,\dots,m\), perform \(\log^2\prm\) trials, each as follows:
    \begin{itemize}[topsep=0pt, itemsep=0pt, parsep=0pt]
        \item Take the next two unused pairs
        \((\idt_1,\sk_{\idt_1})\) and
        \((\idt_2,\sk_{\idt_2})\).

        \item Let
        \[
        x':=\Enc(\mathsf{pp}_{\mathsf{fhe}},\idt_1,x_j)\|\idt_1\|
            \idt_2\|\sk_{\idt_2},
        \qquad
        y_j:=\Dec(\sk_{\idt_1},f(x')).
        \]
        Decompose \(y_j=c'\|\ell'\|\pi'\).
        The trial succeeds if \(y_j\) has the right format,
        \[
        \ell_j^+ \leq \ell' \leq 4\tau,
        \qquad
        \mathsf{V}(\mathsf{pp}_{\mathsf{ivc}},
                    (\chi,\ell',c'),\pi')=1.
        \]
        Decryption failures count as failed trials.
    \end{itemize}
    If any trial succeeds, let \(x_{j+1}\) be the output
    \(y_j\) of the first successful trial, and let \(\ell_{j+1}\) be its level. If all trials fail:
    \begin{itemize}[topsep=0pt, itemsep=0pt, parsep=0pt]
        \item With probability \(\frac12\), take the
        last two pairs and return
        \[
        \Enc(\mathsf{pp}_{\mathsf{fhe}},\idt_1,x_j)\|\idt_1\|
        \idt_2\|\sk_{\idt_2}.
        \]
        \item Otherwise, draw \(\log^2\prm\) samples from \(\dist\) and
        return the input of the first unencrypted one, or a fresh input
        from \(\dist_{\inspace}\) if there is none.
    \end{itemize}
    \item Return a fresh input from \(\dist_{\inspace}\).
\end{enumerate}
Notice that the final encrypted input is freshly generated
and is not one of the ciphertexts tested during the search.

\paragraph{Analysis of the Attack.}
Lemma~\ref{lem:terminal-level-hardness} applies to every circuit of size at most \(\tau^\eta-2\prm\, g(\prm)=\prm^{\eta}\,S-2\prm\, g(\prm)\). As \(S\geq 4\prm\, g(\prm)\), for all sufficiently large \(\prm\) both \((1+a)S\) and \(2S\) are at most this bound. This is where item (ii) uses the lower bound on \(S\). The upper bound on \(S_E\) makes the attack of item (ii) of size at most \(S\), as shown below.

Fix an arbitrary defender \(\prov\) of size at most \(S\)
whose published model \(f\) has evaluation cost at most
\(S_E\).
Each trial costs at most \(S_E+4g(\prm)\), for evaluating \(f\), \(\Enc\),
\(\Dec\), \(\mathsf{V}\) and the comparisons. Step 2 parses
\(4(2m\log^2\prm+2)\) samples and sorts the \(2m\log^2\prm+2\) identities
with a sorting network, at cost \(O(m\log^4\prm\; g(\prm))\), since
\(m\leq\poly(\prm)\). With \(m=5\sqrt{\prm}\,S^{1/(2\eta)}\), the attack has size at
most
\[
m\log^2\prm\,\bigl(S_E+O(g(\prm)\log^2\prm)\bigr)
\leq\frac{5S}{\sqrt\prm}+O\bigl(\sqrt\prm\,S^{\frac{1}{2\eta}}g(\prm)\log^4\prm\bigr)\leq S
\]
for all sufficiently large \(\prm\), by the upper bound on \(S_E\),
\(g(\prm)\geq\prm\), and \(S^{1-\frac{1}{2\eta}}\geq 2g(\prm)^2\log^2\prm\), which
follows from the range of \(S_E\). As \(S_E\geq 2g(\prm)\), this size is also
\(\poly(\prm)\,S^{1/(2\eta)}S_E\).

Suppose that all \(m\) rounds find a valid successor.
Then their levels satisfy
\(
\ell_{j+1}\geq \ell_j^+,
\)
and therefore
\(
\ell_{m+1}\geq(1+m/5)^2>\tau.
\)
The defender \(\prov\) followed by the attack, of total size at most \(2S \leq \prm^{\eta}\,S - 2\prm\, g(\prm)\), would
then produce a valid \((\chi,\ell,c)\) at a level at least \(\tau\), contradicting Lemma~\ref{lem:terminal-level-hardness}. Thus, except with
negligible probability, the attack finds a round in
which all \(\log^2\prm\) trials fail. By the same application of
Lemma~\ref{lem:terminal-level-hardness}, \(\ell_j<\tau\leq 2\tau\) for every round,
except with negligible probability, so \(x_j\) is a valid question.

As in the proof of Theorem~\ref{thm:mitdetgen}, with the check
\(\ell_j^+\leq \ell'\leq 4\tau\),
\(\mathsf{V}(\mathsf{pp}_{\mathsf{ivc}},(\chi,\ell',c'),\pi')=1\) of a trial, which
is the check of \(h\) on a valid question at level \(\ell_j\), the expected error is
\[
\mathbb{E}\left[h(x_\ver,f(x_\ver))
\GivenExperiment
\StateExperiment{
    (\dist,h)\gets\task_\prm \\
    (z_f,z_R,\priv)\gets\prov^{\ex} \\
    x_\ver\gets\ver_\prm^{\ex}(z_f,z_R)
}\right]
\geq\tfrac14-\negl(\prm),
\]
and the expected absolute flag-rate gap is negligible, since \(x_\ver\)
and a natural input differ only in the plaintext \(x_j\) of the encrypted
part.
Fix a polynomial \(p > 4\). For all sufficiently large \(\prm\) the
attacker violates both Robustness with
\(s_{\AD} = \tfrac14 - \tfrac1p\) and Distinguishability with
\(c_{\AD} = \tfrac1p\), whatever \(\varepsilon\) and
\(\delta\) are, so the AD conditions hold at finitely many \(\prm\)
only. Thus, \(\prov\) is not a successful AD. The attack has size
\(\poly(\prm)\,S^{1/(2\eta)} S_E\). This proves item (ii).

\paragraph{The Active Adversarial Defense.}
\(\provtrain\) runs the following algorithm:
\begin{enumerate}[topsep=0pt, itemsep=0pt, parsep=0pt]
    \item Run \(M\) on \(\chi\), together with the IVC updates,
    for \(S/(2g(\prm))\) steps. Record \(N=S_E/(2g(\prm))\) approximately evenly spaced certified configurations \(y_1, \dots, y_N\), including the final completed checkpoint.
    \item Let \(f'\) be the algorithm that, on a clear input \(x = c \| \ell \| \pi\), returns the lowest level tuple in \((y_1,\dots,y_N)\) that is a valid answer for \(x\) according to \(\tasktime\), and a fixed element of \(\outspace\) if there is none or \(x\) is malformed. Set \(f\) as the algorithm that on a clear input returns \(f'(x)\), and on an encrypted input \(x = x^{\mathsf{enc}} \| \idt_1 \| \idt_2 \| \sk_{\idt_2}\) returns \(\Eval(\mathsf{pp}_{\mathsf{fhe}}, f', x^{\mathsf{enc}})\).
    \item Publish \(z_f\) together with the description \(z_R\) of the rule \(R \equiv 0\), and keep \(\priv = \bot\).
\end{enumerate}

\(\provfix\) receives as input \(x \in \inspace\).
Consider the following algorithm \(f^+\):
\begin{enumerate}[topsep=0pt, itemsep=0pt, parsep=0pt]
    \item Parse \(x = c \| \ell \| \pi\).
    \item If the input is malformed or its level exceeds \(\tau\), it returns a fixed element of \(\outspace\).
    \item Run \(M\) for \(\sqrt \ell\) steps from \(c\), reaching
    \(c'=M^{(\ell^+)}(\chi)\), extend \(\pi\) to a proof \(\pi'\) for
    \((\chi,\ell^+,c')\) by \(\sqrt \ell\) calls of \(\mathsf{Update}\), and return
    \(c'\|\ell^+\|\pi'\).
\end{enumerate}
\(\provfix\) runs \(f^+\) on a clear input and \(\Eval(\mathsf{pp}_{\mathsf{fhe}}, f^+, x^{\mathsf{enc}})\) on an encrypted one, and outputs the flag \(r = R(\priv, x) = 0\) of the published rule.

\paragraph{Correctness and Completeness.}
Training runs \(S/(2g(\prm))\) steps, so for \(S\geq 4\prm\, g(\prm)\) the
last recorded configuration has level at least \(2\prm\) and at most
\(\tau\). Let \(x=c\|\ell\|\pi\) be a valid clear question. If \(\ell\leq\prm\),
then \(\ell^+\leq 2\ell\leq 2\prm\), so \(f'(x)\) is a valid answer. If
\(\ell\leq\tau\), then \(\ell^+\leq 2\tau\leq 4\tau\), so \(f^+(x)\) is a valid
answer. An encrypted natural question is inadmissible, and then \(h=0\),
or \(h\) decrypts it to a valid clear question \(x\), and then
\(\Eval(\mathsf{pp}_{\mathsf{fhe}},f',\cdot)\) and
\(\Eval(\mathsf{pp}_{\mathsf{fhe}},f^+,\cdot)\) decrypt to \(f'(x)\) and
\(f^+(x)\) by perfect correctness. Since \(\Pr[\ell>\prm]=2^{-\prm}\) and
\(\Pr[\ell>\tau]=2^{-\tau}\) for natural levels, and \(\err_{x,h}(y)=h(x,y)\)
for batch size \(1\), we obtain
\[
\Pr\left[\errf\leq 2^{-\prm}\GivenExperiment\StateExperiment{
    (\dist,h)\gets\task_\prm \\
    z_f\gets\provtrain^{\ex}}\right]=1,
\]
\[
\Pr\left[\err_{x,h}(y)=1\GivenExperiment\StateExperiment{
    (\dist,h)\gets\task_\prm \\
    (z_f,z_R,\priv)\gets\provtrain^{\ex} \\
    x\gets\dist_{\inspace} \\
    (y,r)\gets\provfix^{\ex}(\priv,x)}\right]
\leq 2^{-\tau}.
\]
As \(2^{-\prm}\) and \(2^{-\tau}\) are negligible, and
\(\err_{x,h}(y)\in\{0,1\}\), these bounds give Correctness and
Completeness with negligible \(\varepsilon\) and \(\delta\).

\paragraph{Robustness.}
Fix an attacker \(\ver\) of size at most \(aT\). Training followed by
\(\ver\) has size at most \((1+a)S\leq\prm^{\eta}\,S-2\prm\, g(\prm)\), so by
Lemma~\ref{lem:terminal-level-hardness} the question of \(x_\ver\),
decrypted if \(x_\ver\) is encrypted, is valid with level at least \(\tau\)
only with negligible probability. If \(x_\ver\) is encrypted and
inadmissible, or its question is invalid, then \(h(x_\ver,\cdot)=0\).
Otherwise its question has level below \(\tau\), and the argument of
Correctness and Completeness gives \(h(x_\ver,y)=0\), with the
admissibility of Section~\ref{sec:separation} in place of perfect
correctness when \(x_\ver\) is encrypted. Hence
\[
\mathbb{E}\left[h(x_\ver,y)\GivenExperiment\StateExperiment{
    (\dist,h)\gets\task_\prm \\
    (z_f,z_R,\priv)\gets\provtrain^{\ex} \\
    x_\ver\gets\ver_\prm^{\ex}(z_f,z_R) \\
    (y,r)\gets\provfix^{\ex}(\priv,x_\ver)}\right]
\leq\negl(\prm),
\]
with one negligible function for all attackers of size \(aT\), as in
the proof of Theorem~\ref{thm:mitdetgen}. As Correctness, Completeness and
Robustness hold for all sufficiently large \(\prm\), the defense has
frequency \(1\).

\paragraph{Efficiency.}
One step of \(M\) together with \(\mathsf{Update}\) costs at most \(g(\prm)\), so the \(S/(2g(\prm))\) steps of training cost at most \(S/2\), and recording the \(N\) configurations costs at most \(N g(\prm)=S_E/2\leq S/2\). So training has size at most \(S\).
Inference reads its input and performs at most \(\sqrt{\tau}\) steps with their updates, which is a circuit \(f^+\) of size \(\poly(\prm)\,S^{1/(2\eta)}\) and depth \(\poly(\prm)\,S^{1/(2\eta)}\). On an encrypted input it evaluates \(f^+\) homomorphically, which by the efficiency of \(\Eval\) costs \(|f^+| \cdot p_{\Eval}(\prm)\), a per-gate overhead that does not depend on the depth \(\poly(\prm)\,S^{1/(2\eta)}\) of \(f^+\). A leveled scheme, whose per-gate cost grows with that depth, would add a power of \(S\) here. Hence its size is \(\poly(\prm)\,S^{1/(2\eta)}\).

With \(N\) as in step 1 of \(\provtrain\), the evaluation circuit of \(f\) dispatches on
the format of the input and selects the lowest-level stored tuple that
is a valid answer, in the clear or under \(\Eval\), so it has size at
most \(g(\prm)(N+1)\leq S_E\), provided \(S_E\geq 2g(\prm)\).
Thus \(f\) satisfies the evaluation bound.
\end{proof}

%% file: sections/neither.tex
The task \(\task\) of Theorem~\ref{thm:mitdetgentime} admits
no Adversarial Defense of size \(S\), by item (ii) of that theorem. It admits
no Transferable Attack of size \(aS\) either, because the inference phase
\(\provfix\) of its Active Adversarial Defense is a defender on its own. It
runs \(f^+\), in the clear or homomorphically, outputs the flag \(0\),
consults neither \(f\) nor \(\priv\), and has size
\(\poly(\prm)\cdot S^{1/(2\eta)}\). For an attacker \(\ver\) of size
\(aS\leq\tau^{\eta}-2\prm\, g(\prm)\), where \(\tau=\prm\cdot S^{1/\eta}\), Lemma~\ref{lem:terminal-level-hardness}
shows that the question of \(x_\ver\) is valid with level at least
\(\tau\) only with negligible probability, and \(\provfix\)
answers every valid question of lower level correctly, as in the Robustness
paragraph of the proof of Theorem~\ref{thm:mitdetgentime}. So the expected
error of \(\provfix\) on \(x_\ver\) is negligible, and
Transferability fails at all but finitely many \(\prm\).

\begin{corollary}[Neither a Transferable Attack nor an Adversarial Defense]\label{cor:neithertime}
For every \(S : \N \to \N\) that is bounded by a polynomial in \(\prm\) and satisfies \(S(\prm) \geq 4\prm\, g(\prm)\), with \(\eta\) and \(g\) as in Section~\ref{sec:separation-time}, there is a learning task \(\task\) that  admits neither an Adversarial Defense nor a Transferable Attack at the following budgets: for every indistinguishability parameter \(s_{\TA} : \N \to (0,1)\), every constant \(a > 0\), and every \(S_E : \N \to \N\) with
\[2g(\prm) \leq S_E \leq S^{1-\frac{1}{2\eta}}/\bigl(g(\prm)\log^2\prm\bigr)\]
\begin{enumerate}[label=(\roman*), leftmargin=2em, topsep=0pt, itemsep=0pt, parsep=0pt]
    \item For every circuit family \(\ver\) of size \(aS\), every polynomial \(p > 1\) and every \(\varphi > 0\)
    \[
    \ver \notin_{\varphi} \TA\Big(\task, 1, aS, \poly(\prm)\cdot S^{\frac{1}{2\eta}}, \tfrac{1}{p(\prm)}, s_{\TA}\Big).
    \]
    \item For every \(\prov\) of size \(S\) whose model class has evaluation circuits of size \(S_E\),
    \[
    \prov \notin_{\varphi} \AD\Big(\task, 1, \poly(\prm)\cdot S^{\frac{1}{2\eta}} \cdot S_E, S, \varepsilon, \delta, \tfrac{1}{p(\prm)}, \tfrac14 - \tfrac{1}{p(\prm)}\Big),
    \]
    for all polynomials \(p > 4\), \(\varepsilon, \delta : \N \to (0,1)\) and \(\varphi > 0\).
\end{enumerate}
\end{corollary}

%% file: sections/ai_use_statement.tex
The authors reviewed all work prepared with the assistance of generative AI (ChatGPT 6, Claude Fable 5, Opus 5.5) and take full responsibility for the content of this paper, including every passage and figure so prepared.

\textbf{Research Contributions.} All research contributions are the authors' own. The problem formulation, the definitions, the statements of the theorems and corollaries, and every proof were conceived by the authors.

\textbf{Writing.} The authors used generative AI assistants to help draft and revise parts of the exposition, and for copy-editing of English usage and grammar. Every passage prepared with such assistance was read, checked against the underlying mathematics, edited where needed and approved by the authors.

\textbf{Figures.} The authors used generative AI assistance to produce the code that draws the figures. What each figure shows, and its correspondence to the results it illustrates, was specified and verified by the authors.

\textbf{Not Applicable.} This work is theoretical. It involves no experiments, no datasets and no released software, so no generative AI was used for experimentation, data collection, data analysis or the preparation of research artifacts. Generative AI was not used for translation.

%% file: appendix/trichotomy.tex
This appendix derives the values for \citet{gluch2026good} in Table~\ref{tab:trichotomy}. As in Lemma~\ref{lem:dichotomy}, we consider binary classification with batch size \(q = 1\). 

In our notation, the formal version of their main theorem, about Watermarks (WMs), Adversarial Defenses (ADs) and Transferable Attacks (TAs), gives the following. We do not define WMs here, since the details are irrelevant for our discussion. The only fact that we need to know is that a WM implies a TA. All constants in this appendix, including the transferability \(10/48\) of the TA obtained from a WM and the indistinguishability \(28/48\), come from the proof of their Theorem~5, and Corollary~\ref{cor:gluch-free} replaces the threshold \(19/24\) of that proof by \(1-c\).
\begin{theorem}[{Consequence of \citealp[Theorem~5]{gluch2026good}}]\label{thm:gluch-trichotomy}
Let \(\varepsilon \in (0, 1/7)\), \(\delta \in (0, 1/48]\) and \(S : \N \to \N\). Let \(\task\) be a learning task that is learnable to error \(\varepsilon\) with confidence \(1 - \delta\) by circuits of size \(O(\sqrt{S(\prm)}/\log S(\prm))\), and let \(\fclass\) be a family of function classes with \(\sqrt{S(\prm)}/\log S(\prm) = \Omega(m_{\fclass}(\prm) + \prm)\). Then, with frequency \(1/3\), at least one of the following exists:
\begin{enumerate}[label=(\roman*), leftmargin=2em, topsep=0pt, itemsep=0pt, parsep=0pt]
    \item \(\mathsf{WM}\bigl(\task,1, S(\prm), o(\sqrt{S(\prm)}/\log S(\prm)), \varepsilon, 20/48, 42/48, 38/48\bigr)\),
    \item \(\AD\bigl(\task, 1, o(\sqrt{S(\prm)}/\log S(\prm)), O(S(\prm)), \varepsilon, 1/48, c_{\AD}, 44/48 + c_{\AD}\bigr)\) for every \(c_{\AD} \in (0, 4/48)\),
    \item \(\TA\bigl(\task, 1, S(\prm), S(\prm), 6/48, 28/48\bigr)\).
\end{enumerate}
\end{theorem}
Items (ii) and (iii) restate their AD and TA as our AD and TA. For the AD, an attacker whose inputs are flagged at a rate within \(c_{\AD}\) of the rate on natural inputs is answered wrongly with probability at most their soundness bound \(11/24\) plus their false-positive rate \(1 - 13/24\) plus \(c_{\AD}\), and their flag can serve as our published rule \(R\), since its description depends only on \(\prm\) and on \(f\).

Fix the assumptions of Theorem~\ref{thm:gluch-trichotomy}. Since the WM of Theorem~\ref{thm:gluch-trichotomy} implies a 
\[\TA\bigl(\task, 1, S(\prm), o(\sqrt{S(\prm)}/\log S(\prm)), 10/48, 28/48\bigr),\] we obtain the following.

\begin{corollary}[Dichotomy from the Trichotomy]\label{cor:gluch-dichotomy}
With frequency \(1/2\), at least one of the following exists:
\begin{enumerate}[label=(\roman*), leftmargin=2em, topsep=0pt, itemsep=0pt, parsep=0pt]
    \item \(\TA\bigl(\task, 1, S(\prm), o(\sqrt{S(\prm)}/\log S(\prm)), 6/48, 28/48\bigr)\),
    \item \(\AD\bigl(\task, 1, o(\sqrt{S(\prm)}/\log S(\prm)), O(S(\prm)), \varepsilon, 1/48, c_{\AD}, 44/48 + c_{\AD}\bigr)\) for every \(c_{\AD} \in (0, 4/48)\).
\end{enumerate}
\end{corollary}

Replacing the threshold \(19/24\) of their proof by \(1 - c\) we obtain a final corollary.
\begin{corollary}[Dichotomy with a Free Threshold]\label{cor:gluch-free}
For every \(c \in (0, 1/4 - \delta)\), with frequency \(1/2\), at least one of the following exists:
\begin{enumerate}[label=(\roman*), leftmargin=2em, topsep=0pt, itemsep=0pt, parsep=0pt]
    \item \(\TA\bigl(\task, 1, S(\prm), o(\sqrt{S(\prm)}/\log S(\prm)), c, 1 - 2c\bigr)\),
    \item \(\AD\bigl(\task, 1, o(\sqrt{S(\prm)}/\log S(\prm)), O(S(\prm)), \varepsilon, \delta, c_{\AD}, 4c + 4\delta + c_{\AD}\bigr)\) for every \(c_{\AD} \in (0, 1 - 4c - 4\delta)\).
\end{enumerate}
\end{corollary}

Larger transferability and smaller indistinguishability make a TA stronger, and smaller robustness and larger distinguishability make an AD stronger. In Table \ref{tab:trichotomy} the values for \citet{gluch2026good} are those of Corollary~\ref{cor:gluch-free}. The first two rows take their constants \(c = 5/24\) and \(\delta = 1/48\), so that
\[
c_{\TA} = \tfrac{10}{48} \approx 0.21, \qquad
s_{\TA} = \tfrac{28}{48} \approx 0.58, \qquad
s_{\AD} = 4c + 4\delta = \tfrac{44}{48} \approx 0.92.
\]
The last two rows match the transferability \(c = 0.10\) of Corollary~\ref{cor:dichotomy} and take \(\delta = 0.01\), so that
\[
c_{\TA} = 0.10, \qquad
s_{\TA} = 0.80, \qquad
s_{\AD} = 4c + 4\delta = 0.44.
\]
In both pairs of rows the robustness \(4c + 4\delta + c_{\AD}\) of their AD holds for every \(c_{\AD}\), and the table lists the limit \(c_{\AD} \to 0\).
The rows for our results read Lemma~\ref{lem:dichotomy} at \(\varepsilon = 0.01\) and Corollary~\ref{cor:dichotomy} at \(\varepsilon = \delta = 0.01\), where the corollary gives \(c_{\TA} = 10\varepsilon\), \(s_{\TA} = \varepsilon\), \(s_{\AD} = 25\varepsilon\) and \(c_{\AD} = \varepsilon^2/2\). The row for Lemma~\ref{lem:dichotomy} takes \(s = 0.02\), \(\alpha = 0.1\), \(\kappa = 0.985\) and \(\delta = 0.01\).


%% file: appendix/games.tex
As in \citet{gluch2026good}, the dichotomy, Lemma~\ref{lem:dichotomy}, is proved by framing interactions between Alice and Bob as zero-sum games. In this appendix we define such games and focus on statements crucial for our results.

Consider the following game specified by an \(N_r \times N_c\) matrix \(M\). The row player \(\mathsf{Min}\) chooses a distribution \(p\) over rows. Analogously, we define a column player \(\mathsf{Max}\) that chooses a distribution \(q\) over columns. Then a row and a column are drawn randomly according to \(p\) and \(q\), respectively, and \(\mathsf{Min}\) pays \(M_{ij}\) to \(\mathsf{Max}\). As the name suggests, the goal of the row player is to minimize the expected payment
\begin{align}\label{eq:exp-payment}
    \sum_{ij}p(i) q(j) M_{ij}.
\end{align}
In the case when both players play optimally, the expected payment is called the value of the game denoted by \(\operatorname{val}(M)\).

Now assume that the row player chooses \(p\) first and reveals it to \(\mathsf{Max}\) before they make their decision. To win the game, \(\mathsf{Max}\) will pick \(q\) that maximizes (\ref{eq:exp-payment}). Notice that
\begin{align*}
     \sum_{j}q(j) \sum_i p(i) M_{ij} \leq \max_j \sum_i p(i) M_{ij},
\end{align*}
meaning that \(\mathsf{Max}\) picks a single column \(j\) that maximizes \(\sum_i p(i) M_{ij}\).
The Min-Max Theorem \citep{v1928theorie} says that reversing the roles does not affect the value of the game:
\begin{align*}
    \operatorname{val}(M)=\min_p \max_j \sum_i p(i) M_{ij} = \max_q \min_i \sum_j q(j) M_{ij}.
\end{align*}
In our construction of Transferable Attacks and Adversarial Defenses we consider a game, where each row is represented by a computationally bounded attack \(\ver\) that generates adversarial \(x \in \{0,1\}^\prm\) and each column is basically represented by a defender \(\prov\) that on input \(x\) outputs a pair \((y,r) \in \{0,1\}^2\). The matrix value \(M_{\ver,\prov}\) corresponds to the ability of \(\prov\) to classify \(x\) correctly by \(y\) and to distinguish adversarial \(x\) from natural inputs by outputting \(r=1\). If the value of this game is smaller than a certain threshold \(t\), then there is a distribution \(p\) over \(\ver\)'s such that for all \(\prov\)
\begin{align*}
    \sum_\ver p(\ver) M_{\ver,\prov} < t.
\end{align*}
We define the matrix \(M\) so that the attack \(\ver_N\) that samples \(\ver\) according to \(p\) has the (extended) matrix value \(M_{\ver_N,\prov}\) smaller than \(t\) for all defenders \(\prov\).

One caveat is that \(\ver_N\) might be a huge circuit since there is a large number of rows to choose from. \citet{lipton1994simple} showed that we can approximate the value of the game by cleverly picking a strategy \(p\) that is a uniform distribution over \(k\) rows, where \(k\) is proportional to the logarithm of the number of columns: For any \(\gamma >0\) and \(k=\ln(N_c)/(2\gamma^2)\) there exists a uniform distribution \(p\) over \(k\) rows such that for all \(\prov\)
\begin{align*}
    \sum_\ver p(\ver) M_{\ver,\prov} <t + \gamma(\max_{\ver,\prov}M_{\ver,\prov}-\min_{\ver,\prov}M_{\ver,\prov}).
\end{align*}
Similarly, if the value of the game is greater than or equal to \(t\), one can show that for any \(\gamma >0\) and \(k=\ln(N_r)/(2\gamma^2)\) there exists a uniform distribution \(q\) over \(k\) columns such that for all \(\ver\)
\begin{align*}
    \sum_\prov q(\prov) M_{\ver,\prov} \geq t - \gamma(\max_{\ver,\prov}M_{\ver,\prov}-\min_{\ver,\prov}M_{\ver,\prov}).
\end{align*}

%% file: appendix/crypto.tex
\subsection{Public-Key Encryption}\label{app:pke}
A public-key encryption scheme (PKE) \citep{dif77, rivest1978method} allows for secure communication without the need to exchange secret keys beforehand. If Charlie wants to receive a message from Dave, Charlie generates a pair of public and secret keys and publishes the public key. Dave then encrypts the message with Charlie's public key and sends the ciphertext over a public channel. Correctness ensures that Charlie can decrypt Dave's message with the secret key. Security means indistinguishability under chosen-plaintext attacks (IND-CPA) \citep{Goldwasser1984ProbabilisticE}: a polynomial-size adversary that knows the public key cannot distinguish encryptions of zero from encryptions of one, except with a negligible advantage.
\begin{definition}[PKE]\label{def:pke}
    Let \(\lambda\in \N\) be a security parameter. A public-key encryption scheme with message space \(\{0,1\}\) and ciphertext set family \(\mathcal{C}=\{\mathcal{C}_\lambda\}_\lambda\) consists of a tuple of probabilistic polynomial-time algorithms \((\Gen, \Enc, \Dec)\) with the following properties:
    \begin{itemize}[topsep=0pt, itemsep=0pt, parsep=0pt]
        \item \textbf{Key Generation.} \((\pk, \sk) \gets \Gen(1^\lambda)\) takes as input the security parameter \(\lambda\) and outputs a public key \(\pk\) and a secret key \(\sk\), both in \(\{0,1\}^{\poly(\lambda)}\).
        \item \textbf{Encryption.} \(c \gets \Enc(\pk, m)\) takes as input the public key \(\pk\) and a plaintext \(m \in \{0,1\}\), and outputs a ciphertext \(c\in \mathcal{C}_\lambda\).
        \item \textbf{Decryption.} \(m' = \Dec(\sk, c)\) takes as input the secret key \(\sk\) and a ciphertext \(c\), and outputs a plaintext \(m'\in \{0,1\}\).
    \end{itemize}
    The scheme must satisfy the following properties:
    \begin{itemize}[topsep=0pt, itemsep=0pt, parsep=0pt]
        \item \textbf{Correctness.} For all \(\lambda\) and \(m \in \{0,1\}\), decryption always succeeds:
        \[
            \Pr\left[\Dec(\sk, \Enc(\pk, m))=m \GivenExperiment (\pk, \sk) \gets \Gen(1^\lambda)\right]=1.
        \]
        \item \textbf{IND-CPA Security.} For every polynomial-size adversary \(\mathsf{Adv}\) there exists a negligible function \(\negl\) such that for all \(\lambda\)
        \begin{multline}\label{eq:ind-cpa}
            \Big|\Pr\left[\mathsf{Adv}(\pk, \Enc(\pk, 0))=1 \GivenExperiment (\pk, \sk)\gets \Gen(1^\lambda)\right] \\
            -\Pr\left[\mathsf{Adv}(\pk, \Enc(\pk, 1))=1 \GivenExperiment (\pk, \sk)\gets \Gen(1^\lambda)\right]\Big| \leq \negl(\lambda).
        \end{multline}
    \end{itemize}
\end{definition}
Throughout the paper without loss of generality we assume that after fixing security parameter \(\lambda\), both \(\Gen\) and \(\Enc\) have fixed output length.

In the above definition the indistinguishability property (\ref{eq:ind-cpa}) is equivalent to
\begin{align}\label{eq:ind-cpa-guess}
    \Pr\left[\mathsf{Adv}(\pk,\Enc(\pk, y)) \neq y \GivenExperiment \StateExperiment{(\pk, \sk)\gets \Gen(1^\lambda) ,\; y\gets \{0,1\}}\right] \geq 1/2- \negl(\lambda).
\end{align}
We use both interpretations in our constructions of Transferable Attacks in Section~\ref{sec:constructions}.

\subsection{Identity-Based Fully Homomorphic Encryption}\label{app:ibfhe}
Identity-based encryption \citep{shamir1984identity, boneh2001identity} lets any string serve as a public key, and fully homomorphic encryption \citep{Gen09} allows computing on ciphertexts. Identity-based fully homomorphic encryption (IB-FHE) \citep{gentryIB-FHE, clear2014bootstrappable} combines the two. Its security is indistinguishability under selective-identity chosen-plaintext attacks (IND-sID-CPA) \citep{canetti2003forward}, where the adversary fixes the identity it attacks before it sees the public parameters.
\begin{definition}[Identity-Based Fully Homomorphic Encryption]\label{def:ibfhe}
An IB-FHE scheme with message space \(\mathcal{M}\), identity space \(\mathcal{I}\), and a class of circuits \(\mathfrak{C} \subseteq \{\mathcal{M}^* \to \mathcal{M}\}\) consists of a tuple of probabilistic polynomial-time algorithms \(\mathsf{IB\text{-}FHE} = (\mathsf{Setup}, \mathsf{KeyGen}, \Enc, \Dec, \Eval)\) with the following properties:
\begin{itemize}[topsep=0pt, itemsep=0pt, parsep=0pt]
    \item \textbf{Setup.} \((\mathsf{pp}, \mathsf{MSK}) \gets \mathsf{Setup}(1^\lambda)\) takes as input the security parameter \(\lambda\) and outputs public parameters \(\mathsf{pp}\) and a master secret key \(\mathsf{MSK}\).
    \item \textbf{Key Generation.} \(\sk_{\idt} \gets \mathsf{KeyGen}(\mathsf{MSK}, \idt)\) takes as input the master secret key \(\mathsf{MSK}\) and an identity \(\idt \in \mathcal{I}\), and outputs a secret key \(\sk_{\idt}\).
    \item \textbf{Encryption.} \(c \gets \Enc(\mathsf{pp}, \idt, m)\) takes as input the public parameters \(\mathsf{pp}\), an identity \(\idt\), and a message \(m \in \mathcal{M}\), and outputs a ciphertext \(c\).
    \item \textbf{Decryption.} \(m' = \Dec(\sk_{\idt}, c)\) takes as input a secret key \(\sk_{\idt}\) and a ciphertext \(c\), and outputs a plaintext \(m'\) if \(c\) is a valid encryption under \(\idt\). Otherwise, it outputs \(\bot\).
    \item \textbf{Evaluation.} \(c' \gets \Eval(\mathsf{pp}, C, c_1, \dots, c_k)\) takes as input the public parameters \(\mathsf{pp}\), a circuit \(C \in \mathfrak{C}\), and a collection of ciphertexts \((c_1, \dots, c_k)\), and outputs a new ciphertext \(c'\).
\end{itemize}
The scheme must satisfy the following properties:
\begin{itemize}[topsep=0pt, itemsep=0pt, parsep=0pt]
    \item \textbf{Correctness.} For all \(\lambda\), identities \(\idt \in \mathcal{I}\), circuits \(C \in \mathfrak{C}\), and messages \(m_1, \dots, m_k \in \mathcal{M}\),
    \[
    \Pr\left[ \Dec(\sk_{\idt}, \Eval(\mathsf{pp}, C, c_1, \dots, c_k)) = C(m_1, \dots, m_k) \GivenExperiment \StateExperiment{(\mathsf{pp}, \mathsf{MSK}) \gets \mathsf{Setup}(1^\lambda) \\ \sk_{\idt} \gets \mathsf{KeyGen}(\mathsf{MSK}, \idt) \\ c_i \gets \Enc(\mathsf{pp}, \idt, m_i),\ i \in [k]} \right] = 1.
    \]
    \item \textbf{Efficiency.} There is a polynomial \(p_{\Eval}\) such that for every circuit \(C\), \(\Eval(\mathsf{pp}, C, c_1, \dots, c_k)\) runs in time \(|C| \cdot p_{\Eval}(\lambda)\). The per-gate cost does not depend on \(C\), in particular not on its depth.\footnote{Leveled schemes such as \citet{gentryIB-FHE} have a per-gate cost polynomial in the depth bound, which would add a power of the depth to the cost of evaluating a deep circuit. Bootstrapping removes the dependence under a circular-security assumption. In our tasks the bootstrapping key of the identity of a sample can be supplied with the sample, since the example oracle holds the master secret key. We assume the depth-independent bound and do not pursue the construction.}
    \item \textbf{Compactness.} There exists a polynomial \(p\) such that for all \(\lambda\), \(C\), and ciphertexts \(c_i\),
    \[
    \Pr\left[ |\Eval(\mathsf{pp}, C, c_1, \dots, c_k)| \leq p(\lambda) \GivenExperiment (\mathsf{pp}, \mathsf{MSK}) \gets \mathsf{Setup}(1^\lambda) \right] = 1.
    \]
    \item \textbf{IND-sID-CPA Security.} For every polynomial-size adversary \(\mathsf{Adv}\),
    \[
    \left| \Pr \left[ \mathsf{Game}_0^{\mathsf{Adv}}(\lambda) = 1 \right] - \Pr \left[ \mathsf{Game}_1^{\mathsf{Adv}}(\lambda) = 1 \right] \right| \leq \negl(\lambda),
    \]
    where \(\mathsf{Game}_b^{\mathsf{Adv}}(\lambda)\) is defined as follows:
    \begin{enumerate}[topsep=0pt, itemsep=0pt, parsep=0pt]
        \item \(\mathsf{Adv}\) chooses a challenge identity \(\idt^*\).
        \item \((\mathsf{pp}, \mathsf{MSK}) \gets \mathsf{Setup}(1^\lambda)\) is generated, and \(\mathsf{pp}\) is given to \(\mathsf{Adv}\).
        \item \(\mathsf{Adv}\) may adaptively query \(\mathsf{KeyGen}(\mathsf{MSK}, \idt)\) for identities \(\idt \neq \idt^*\).
        \item \(\mathsf{Adv}\) submits \(m_0, m_1 \in \mathcal{M}\) of equal length.
        \item The challenger returns \(c^* \gets \Enc(\mathsf{pp}, \idt^*, m_b)\).
        \item \(\mathsf{Adv}\) may again query \(\mathsf{KeyGen}(\mathsf{MSK}, \idt)\) for identities \(\idt \neq \idt^*\).
        \item \(\mathsf{Adv}\) outputs a bit \(b'\), which is the output of the game.
    \end{enumerate}
\end{itemize}
\end{definition}

\subsection{Strongly Unforgeable Signatures}\label{app:signatures}
A strongly unforgeable signature scheme \citep{dif77, goldwasser1988digital, bonehsignatures} allows generation of signatures that can be publically verified but are computationally hard to forge.
\begin{definition}[Strongly Unforgeable Signature Scheme]\label{def:signatures}
A strongly unforgeable signature scheme consists of a tuple of probabilistic polynomial-time algorithms \((\Gen, \mathsf{Sign}, \mathsf{Verify})\) with the following properties:
\begin{itemize}[topsep=0pt, itemsep=0pt, parsep=0pt]
    \item \textbf{Key Generation.} \((\mathsf{vk}, \sk) \gets \Gen(1^\lambda)\) takes as input the security parameter \(\lambda\) and outputs a verification key \(\mathsf{vk}\) and a signing key \(\sk\).
    \item \textbf{Signing.} \(\sigma \gets \mathsf{Sign}(\sk, m)\) takes as input the signing key \(\sk\) and a message \(m\), and outputs a signature \(\sigma\).
    \item \textbf{Verification.} \(\mathsf{Verify}(\mathsf{vk}, m, \sigma) \in \{0,1\}\) takes as input the verification key \(\mathsf{vk}\), a message \(m\), and a signature \(\sigma\), and outputs 1 if \(\sigma\) is a valid signature on \(m\) under \(\mathsf{vk}\), and 0 otherwise.
\end{itemize}
The scheme must satisfy the following properties:
\begin{itemize}[topsep=0pt, itemsep=0pt, parsep=0pt]
    \item \textbf{Correctness.} For all \(\lambda\) and all messages \(m\),
    \[
    \Pr\left[ \mathsf{Verify}(\mathsf{vk}, m, \sigma) = 1 \GivenExperiment \StateExperiment{(\mathsf{vk}, \sk) \gets \Gen(1^\lambda) \\ \sigma \gets \mathsf{Sign}(\sk, m)} \right] = 1.
    \]
    \item \textbf{Strong Unforgeability.} For every polynomial-size adversary \(\mathsf{Adv}\), there exists a negligible function \(\negl\) such that
    \[
    \Pr \left[ \mathsf{Verify}(\mathsf{vk}, m^*, \sigma^*) = 1 \text{ and } (m^*, \sigma^*) \notin \mathcal{Q} \GivenExperiment \StateExperiment{(\mathsf{vk}, \sk) \gets \Gen(1^\lambda) \\ (m^*, \sigma^*) \gets \mathsf{Adv}^{\mathsf{Sign}(\sk, \cdot)}(\mathsf{vk})} \right] \leq \negl(\lambda),
    \]
    where \(\mathsf{Adv}\) asks the signing oracle for signatures on polynomially many messages of its choice and \(\mathcal{Q}\) is the set of the message-signature pairs it receives.
\end{itemize}
\end{definition}

\subsection{Succinct Non-Interactive Arguments of Knowledge}\label{app:snarks}
First we define universal relations \(\mathsf{R}_U\) \citep{barak2002universal}. 
\begin{definition}[Universal Relation]\label{def:universal-relation}
A universal relation \(\mathsf{R}_U\) consists of all instance-witness pairs \((y, w)\) where \(y = (M, x, t)\) and \(|y|, |w| \leq t\), such that \(M\) is a random-access machine that accepts \((x, w)\) within at most \(t\) steps. 
\end{definition}

A fully succinct non-interactive argument of knowledge (SNARK) \citep{bitansky2012extractable, bitansky2013recursive, Alefullysuccinctsnarks} lets a prover convince a verifier with a short proof that it knows a witness for an instance of \(\mathsf{R}_U\).

\begin{definition}[SNARK]\label{def:snark}
A SNARK for a relation \(\mathsf{R} \subseteq \mathsf{R}_U\) consists of a triple of algorithms \((\gensnark, \provsnark, \versnark)\) with the following properties:
\begin{itemize}[topsep=0pt, itemsep=0pt, parsep=0pt]
    \item \textbf{Parameter Generation.} \(\mathsf{pp} \gets \gensnark(1^\lambda, T)\) takes as input the security parameter \(\lambda\) and a time bound \(T\), and outputs public parameters \(\mathsf{pp}\).
    \item \textbf{Proving.} \(\pi \gets \provsnark(\mathsf{pp}, y, w)\) takes as input the public parameters \(\mathsf{pp}\), an instance \(y = (M, x, t)\) with \(t \leq T\) and a witness \(w\), and outputs a proof \(\pi\).
    \item \textbf{Verification.} \(\versnark(\mathsf{pp}, y, \pi) \in \{0,1\}\) takes as input the public parameters \(\mathsf{pp}\), an instance \(y\) and a proof \(\pi\), and outputs \(1\) if it accepts \(\pi\) and \(0\) otherwise.
\end{itemize}
The scheme must satisfy the following properties:
\begin{itemize}[topsep=0pt, itemsep=0pt, parsep=0pt]
    \item \textbf{Completeness.} For every large enough security parameter \(\lambda \in \N\), every time bound \(T \in \N\), and every instance-witness pair \((y, w) \in \mathsf{R}\) with \(y = (M, x, t)\) and \(t \leq T\),
    \[
    \Pr\left[\versnark(\mathsf{pp}, y, \pi) = 1 \GivenExperiment \StateExperiment{\mathsf{pp} \gets \gensnark(1^\lambda, T) ,\; \pi \gets \provsnark(\mathsf{pp}, y, w)}\right] = 1.
    \]
    \item \textbf{Adaptive Proof of Knowledge.} For every deterministic polynomial-size prover \(\provsnark^*\), there exists a polynomial-size extractor \(\mathcal{E}^{\provsnark^*}\) such that for every large enough \(\lambda \in \N\), all auxiliary inputs \(z \in \{0,1\}^{\poly(\lambda)}\), and all time bounds \(T \in \N\),
    \[
    \Pr\left[\versnark(\mathsf{pp}, y, \pi) = 1 \text{ and } (y, w) \notin \mathsf{R} \GivenExperiment \StateExperiment{\mathsf{pp} \gets \gensnark(1^\lambda, T) \\ (y, \pi) \gets \provsnark^*(\mathsf{pp}, z) \\ w \gets \mathcal{E}^{\provsnark^*}(\mathsf{pp}, z)}\right] \leq \negl(\lambda).
    \]
    A randomized prover is covered by including its coins in \(z\).
    \item \textbf{Efficiency.} There exists a polynomial \(p\) such that, for every large enough security parameter \(\lambda \in \N\), every time bound \(T \in \N\), and every instance \(y = (M, x, t)\) with \(t \leq T\), the following bounds hold:
    \begin{itemize}[topsep=0pt, itemsep=0pt, parsep=0pt]
        \item Generator running time is \(p(\lambda + \log T)\).
        \item Prover running time is \(p(\lambda + |M| + |x| + t + \log T)\).
        \item Verifier running time is \(p(\lambda + |M| + |x| + \log T)\).
        \item Proof size is \(p(\lambda + \log T)\).
    \end{itemize}
\end{itemize}
\end{definition}

Adaptive proof of knowledge is stated above for all auxiliary inputs \(z\). Extractability with respect to arbitrary auxiliary input is implausible under indistinguishability obfuscation \citep{bitansky2014extractable, boyle2015limits}. Lemma~\ref{lem:pool} invokes the extractor only for the benign auxiliary-input distribution consisting of a verification key, fresh signatures under an independently generated signing key and uniform coins, and the assumption can be restricted to that distribution.

A zero-knowledge SNARK \citep{benSasonzkNSARK} satisfies the zero-knowledge property \citep{ip_goldwasser} in its non-interactive form \citep{blum1988noninteractive}. That is, an honest prover can generate valid proofs without leaking any information beyond the truth of the statement, particularly without revealing any knowledge of the witness used to generate the proof. For zero-knowledge SNARKs, the public parameters \(\mathsf{pp}\) form a common reference string that is trusted by both the verifier and the prover.

\begin{definition}[Zero-Knowledge SNARK]\label{def:zk-snark}
A triple of algorithms \((\gensnark, \provsnark, \versnark)\) is a zero-knowledge SNARK for a relation \(\mathsf{R} \subseteq \mathsf{R}_U\) if it is a SNARK for \(\mathsf{R}\) and satisfies the following property.
There exists a stateful interactive polynomial-size simulator \(\mathsf{Sim}\) such that for every stateful interactive polynomial-size distinguisher \(\mathsf{Adv}\) there exists a negligible function \(\negl\) such that for every large enough security parameter \(\lambda \in \N\) and every time bound \(T \in \N\),
    \begin{multline*}
    \left|\Pr\left[t \leq T,\ (y,w) \in \mathsf{R},\ \mathsf{Adv}(\pi) = 1 \GivenExperiment \StateExperiment{\mathsf{pp} \gets \gensnark(1^\lambda, T) \\ (y, w) \gets \mathsf{Adv}(\mathsf{pp}) \\ \pi \gets \provsnark(\mathsf{pp}, y, w)}\right]\right. \\
    \left.-
    \Pr\left[t \leq T,\ (y,w) \in \mathsf{R},\ \mathsf{Adv}(\pi) = 1 \GivenExperiment \StateExperiment{(\mathsf{pp},\mathsf{trap}) \gets \mathsf{Sim}(1^\lambda, T) \\ (y, w) \gets \mathsf{Adv}(\mathsf{pp}) \\ \pi \gets \mathsf{Sim}(\mathsf{pp}, y, \mathsf{trap})}\right]\right| \leq \negl(\lambda).
    \end{multline*}
\end{definition}

\subsection{Incrementally Verifiable Computation}\label{app:ivc}
Incrementally verifiable computation (IVC) \citep{valiant2008incrementally} lets a party perform a long deterministic computation step by step.  After every step, there is a short proof that its current configuration is correct. Moreover, the proof can be updated to a proof of the next step.

\begin{definition}[Incrementally Verifiable Computation]\label{def:ivc}
An IVC scheme for a deterministic Turing machine \(M\), with language \(\mathsf{L}_M := \{(\chi,t,c) : c = M^{(t)}(\chi)\}\), where \(M^{(t)}(\chi)\) is the configuration of \(M\) on input \(\chi\) after \(t\) steps, and a halted machine stays in its final configuration, which is accepting if and only if \(M\) outputs \(1\), consists of a tuple of algorithms \((\mathsf{G}, \mathsf{P}, \mathsf{V}, \mathsf{Update})\) with the following syntax:
\begin{itemize}[topsep=0pt, itemsep=0pt, parsep=0pt]
    \item \textbf{Parameter Generation.} \(\mathsf{pp} \gets \mathsf{G}(1^\lambda)\) takes as input the security parameter \(\lambda\) and outputs public parameters \(\mathsf{pp}\).
    \item \textbf{Proving.} \(\pi \gets \mathsf{P}(\mathsf{pp}, y)\), takes as input public parameter \(\mathsf{pp} \) and \(y = (\chi, t, c) \in \mathsf{L}_M\) and outputs a proof \(\pi\).
    \item \textbf{Verification.} \(b = \mathsf{V}(\mathsf{pp}, y, \pi)\) outputs \(b \in \{0,1\}\) indicating acceptance or rejection of the proof \(\pi\) for instance \(y\).
    \item \textbf{Update.} \(\pi_t \gets \mathsf{Update}(\mathsf{pp}, (\chi, t-1, c_{t-1}), \pi_{t-1})\) takes as input the public parameters, a previous statement \((\chi, t-1, c_{t-1}) \in \mathsf{L}_M\), and the proof \(\pi_{t-1}\), and outputs a new proof \(\pi_t\) for the statement \((\chi, t, c_t)\).
\end{itemize}
The scheme must satisfy the following properties:
\begin{itemize}[topsep=0pt, itemsep=0pt, parsep=0pt]
    \item \textbf{Completeness.} For every \(y = (\chi, t, c) \in \mathsf{L}_M\),
    \[
    \Pr\left[\mathsf{V}(\mathsf{pp}, y, \pi) = 1 \GivenExperiment \StateExperiment{\mathsf{pp} \gets \mathsf{G}(1^\lambda) ,\; \pi \gets \mathsf{P}(\mathsf{pp}, y)}\right] = 1.
    \]
    \item \textbf{Efficiency.} \(\mathsf{G}\) runs in time \(\poly(\lambda)\) and \(\mathsf{P}\) in time \(\poly(\lambda, |y|, t)\). The proof length is \(|\pi| = \poly(\lambda, \log t)\), the verifier's running time is \(|y| \cdot \poly(\lambda, |\pi|)\), and \(\mathsf{Update}\) runs in time \(\poly(\lambda, |\chi| + |c_{t-1}|, \log t)\).
    \item \textbf{Soundness.} For every polynomial-size cheating prover \(\mathsf{P}^*\) and every polynomial \(t = t(\lambda)\), there exists a negligible function \(\negl\) such that
    \[
    \Pr\left[y = (\chi, t, c) \notin \mathsf{L}_M \text{ and } \mathsf{V}(\mathsf{pp}, y, \pi) = 1 \GivenExperiment \StateExperiment{\mathsf{pp} \gets \mathsf{G}(1^\lambda) ,\; (y, \pi) \gets \mathsf{P}^*(\mathsf{pp})}\right] \leq \negl(\lambda).
    \]
    \item \textbf{Incrementality.} For any valid sequence of configurations \(c_0, c_1, \dots, c_t\) with \(c_0\) the initial configuration of \(M(\chi)\), and \(\pi_0 = \bot\), the proofs \(\pi_1, \dots, \pi_t\) can be computed iteratively using
    \[
    \pi_i \gets \mathsf{Update}(\mathsf{pp}, (\chi, i-1, c_{i-1}), \pi_{i-1}) \quad \text{for each } i \in [t].
    \]
    Moreover, \(\mathsf{Update}\) extends every accepting proof, not only proofs it produced itself: for every \((\chi, t-1, c_{t-1}) \in \mathsf{L}_M\) and every \(\pi_{t-1}\) with \(\mathsf{V}(\mathsf{pp}, (\chi, t-1, c_{t-1}), \pi_{t-1}) = 1\), the proof \(\mathsf{Update}(\mathsf{pp}, (\chi, t-1, c_{t-1}), \pi_{t-1})\) is accepted by \(\mathsf{V}\) for \((\chi, t, c_t)\) with probability \(1\).
\end{itemize}
\end{definition}
The last property is what lets a party continue a computation from a certified configuration that it did not produce, as the inference phase in Section~\ref{sec:separation-time} does with the input it receives. It holds for the schemes obtained by recursive composition of SNARKs \citep{valiant2008incrementally, bitansky2013recursive}, where the proof for step \(t\) is an argument of knowledge of an accepting proof for step \(t-1\) and of the transition, so that any accepting predecessor proof serves as the witness. It is not guaranteed by every IVC scheme. In particular, \citet{ivc} note that their construction from incremental probabilistically checkable proofs does not satisfy it.

\subsection{Non-Parallelizing Languages with Average-Case Hardness}\label{app:npl}
Average-case non-parallelizing languages (NPLs) \citep{tlp} with difficulty parameter \(\tau\) are decidable in time \(\tau\), but polynomial-size circuits of depth at most \(\tau^\eta\), for a constant \(\eta < 1\), decide them on sampled instances with probability only negligibly higher than one half. We require an NPL for every difficulty parameter \(\tau\) and input length, together with a uniform way to decide these languages and to sample hard instances for every \(\tau\).

\begin{definition}[Average-Case Non-Parallelizing Language Ensemble]\label{def:npl}
An average-case NPL ensemble with gap \(\eta < 1\) for a collection of languages \(\{\mathsf{L}_{\lambda, \tau}\}_{\lambda, \tau \in \N}\) with \(\mathsf{L}_{\lambda, \tau} \subseteq \{0,1\}^\lambda\) is a pair of algorithms \(\mathsf{NPL} = (\mathsf{Samp}, M)\) with the following properties:
\begin{itemize}[topsep=0pt, itemsep=0pt, parsep=0pt]
    \item \textbf{Sampling.} \(\chi \gets \mathsf{Samp}(1^\lambda, \tau)\) takes as input a length \(\lambda\) and a difficulty parameter \(\tau\), runs in time \(\poly(\lambda, \tau)\), and outputs an instance \(\chi \in \{0,1\}^\lambda\).
    \item \textbf{Decision.} \(b = M(\tau, \chi)\) takes as input a difficulty parameter \(\tau\) and an instance \(\chi \in \{0,1\}^\lambda\), and outputs a bit \(b \in \{0,1\}\).
\end{itemize}
The algorithms must satisfy the following properties:
\begin{itemize}[topsep=0pt, itemsep=0pt, parsep=0pt]
    \item \textbf{Sequential Decidability.} For every \(\lambda, \tau \in \N\) and every \(\chi \in \{0,1\}^\lambda\), \(M(\tau, \chi)\) runs in time \(\tau\) and space \(\poly(\lambda)\), and outputs \(1\) if and only if \(\chi \in \mathsf{L}_{\lambda, \tau}\). In particular, every configuration of \(M(\tau, \chi)\) has length \(\poly(\lambda)\), independently of \(\tau\), and one step of \(M\) is computed by a circuit of size \(\poly(\lambda)\).
    \item \textbf{Average-Case Non-Parallelizing.} For every family \(\mathsf{Adv} = \{\mathsf{Adv}_\lambda\}_{\lambda \in \N}\) of polynomial-size circuits, there exists a negligible function \(\negl\) such that for every \(\lambda \in \N\) and every \(\tau \geq \lambda\), if \(\mathsf{Adv}_\lambda\) has depth at most \(\tau^\eta\), then
    \[
    \Pr\left[\mathsf{Adv}_\lambda(\chi) = M(\tau, \chi) \GivenExperiment \chi \gets \mathsf{Samp}(1^\lambda, \tau)\right] \leq \frac{1}{2} + \negl(\lambda).
    \]
\end{itemize}
\end{definition}
Since the depth of a circuit is at most its size, hardness against shallow circuits implies hardness against small ones. In particular, the second property applies to every family of polynomial-size circuits with \(|\mathsf{Adv}_\lambda| \leq \tau^\eta\). This is how we use it, since our parties are measured by size.